\documentclass[conference]{IEEEtran}
\usepackage{times}

\usepackage[numbers]{natbib}
\usepackage{multicol}

\usepackage{algorithm}
\usepackage{algpseudocode}
\usepackage{amsmath}
\usepackage{amssymb}
\usepackage{amsthm}
\usepackage{booktabs}
\usepackage{color}
\usepackage{graphicx}
\usepackage{listings}
\usepackage{mathrsfs}
\usepackage[labelformat=simple]{subcaption}

\usepackage{svg}
\usepackage{url}
\usepackage{wrapfig}
\usepackage{mathtools}
\usepackage{xspace}
\usepackage{booktabs}
\usepackage{tabularx}
\newcolumntype{Y}{>{\centering\arraybackslash}X}

\DeclarePairedDelimiterXPP\Prob[1]{{P}}(){}{
   
   #1}

\usepackage{multicol}
\usepackage{bm}
\usepackage[bookmarks=true,colorlinks=true,citecolor=blue,linkcolor=blue]{hyperref}
\usepackage{flushend}

\definecolor{deepblue}{rgb}{0,0,0.5}
\definecolor{deepred}{rgb}{0.6,0,0}
\definecolor{deepgreen}{rgb}{0,0.5,0}

\definecolor{es-blue}{rgb}{0,0.4,0.8}

\newcommand{\src}{\textrm{src}}
\newcommand{\tgt}{\textrm{tgt}}
\newcommand{\sw}{\textrm{sw}}

\newcommand{\fwr}{\gamma}

\usepackage[textsize=tiny]{todonotes} %

\usepackage{comment}

\usepackage{authblk}

\begin{document}

\title{To the Noise and Back:\\Diffusion for Shared Autonomy}

\author[1\thanks{Corresponding author: takuma@ttic.edu}]{Takuma Yoneda}
\author[2]{\,Luzhe Sun}
\author[3,4]{Ge Yang}
\author[5]{Bradly Stadie}
\author[1]{Matthew R.\ Walter}
\affil[1]{Toyota Technological Institute at Chicago (TTIC)}
\affil[2]{Department of Computer Science, University of Chicago}
\affil[3]{Institute of Artificial Intelligence and Fundamental Interactions (IAIFI)}
\affil[4]{Computer Science and Artificial Intelligence Laboratory (CSAIL), MIT}
\affil[5]{Department of Statistics and Data Science, Northwestern University}

\maketitle

\begin{abstract}
Shared autonomy is an operational concept in which a user and an autonomous agent collaboratively control a robotic system. It provides a number of advantages over the extremes of full-teleoperation and full-autonomy in many settings. Traditional approaches to shared autonomy rely on knowledge of the environment dynamics, a discrete space of user goals that is known a priori, or knowledge of the user's policy---assumptions that are unrealistic in many domains.
Recent works relax some of these assumptions by formulating shared autonomy with model-free deep reinforcement learning (RL).
In particular, they no longer need knowledge of the goal space (e.g., that the goals are discrete or constrained) or environment dynamics. However, they need knowledge of a task-specific reward function to train the policy. Unfortunately, such reward specification can be a difficult and brittle process. On top of that, the formulations inherently rely on human-in-the-loop training, and that necessitates them to prepare a policy that mimics users' behavior.
In this paper, we present a new approach to shared autonomy that employs a modulation of the forward and reverse diffusion process of diffusion models. Our approach does not assume known environment dynamics or the space of user goals, and in contrast to previous work, it does not require any reward feedback, nor does it require access to the user's policy during training. Instead, our framework learns a distribution over a 
space of desired behaviors. It then employs a diffusion model to translate the user's actions to a sample from this distribution. Crucially, we show that it is possible to carry out this process in a manner that preserves the user's control authority. We evaluate our framework on a series of challenging continuous control tasks, and analyze its ability to effectively correct user actions while maintaining their autonomy.

\end{abstract}

\IEEEpeerreviewmaketitle

\section{Introduction}

Contemporary robots primarily operate in one of two different ways---full teleoperation or full autonomy. Teleoperation is common in unstructured environments (e.g., underwater), where the proficiency with which robots are able to understand their surroundings is insufficient for fully autonomous robots to operate reliably. However, direct teleoperation requires users to  interpret the robot's environment observations while simultaneously controlling its low-level actions, a responsibility that is particularly challenging for highly dynamic tasks. This operational gap motivates a setting in which a human and an autonomous agent collaborate and share control of the robot.

\begin{figure}
    \centering
    \includegraphics[width=0.49\linewidth]{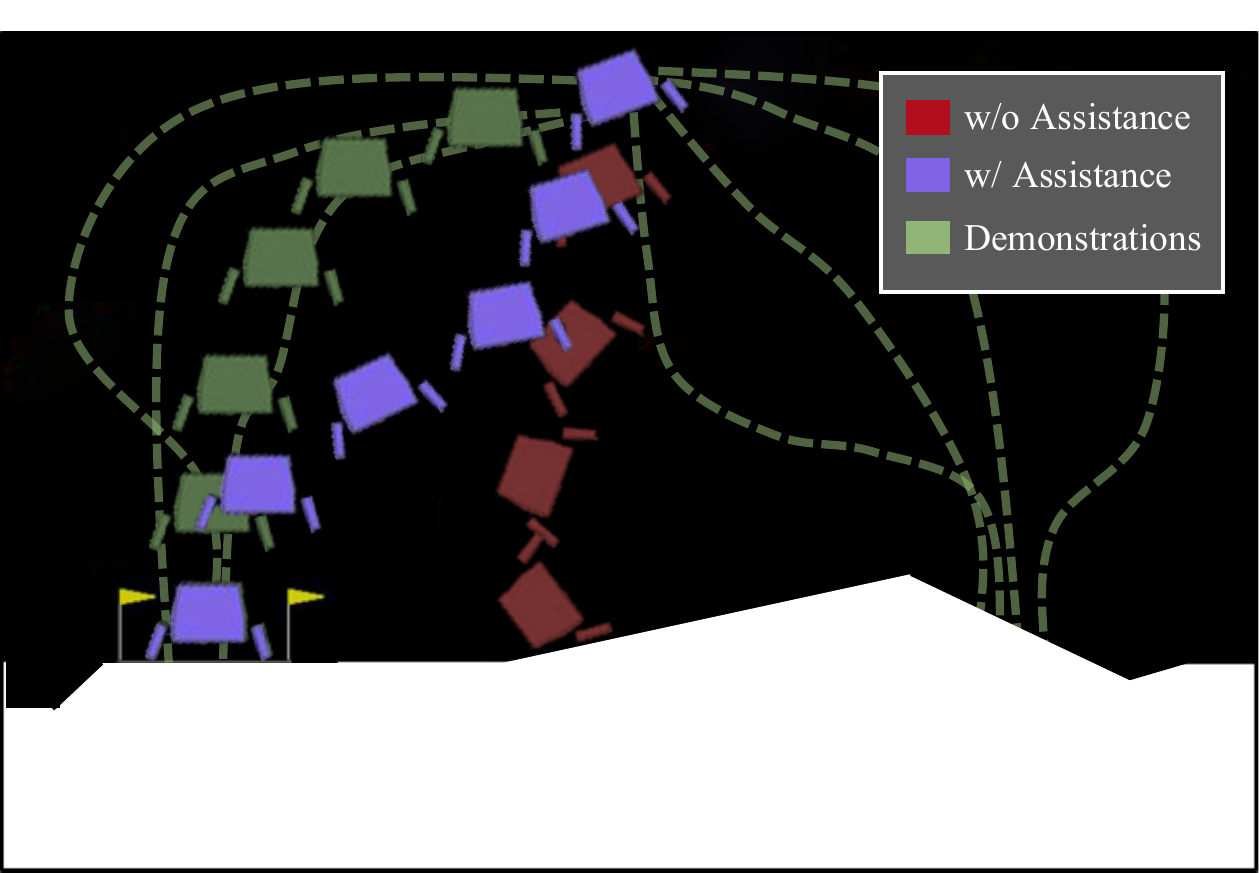}\label{fig:illustration1}%
    \includegraphics[width=0.49\linewidth]{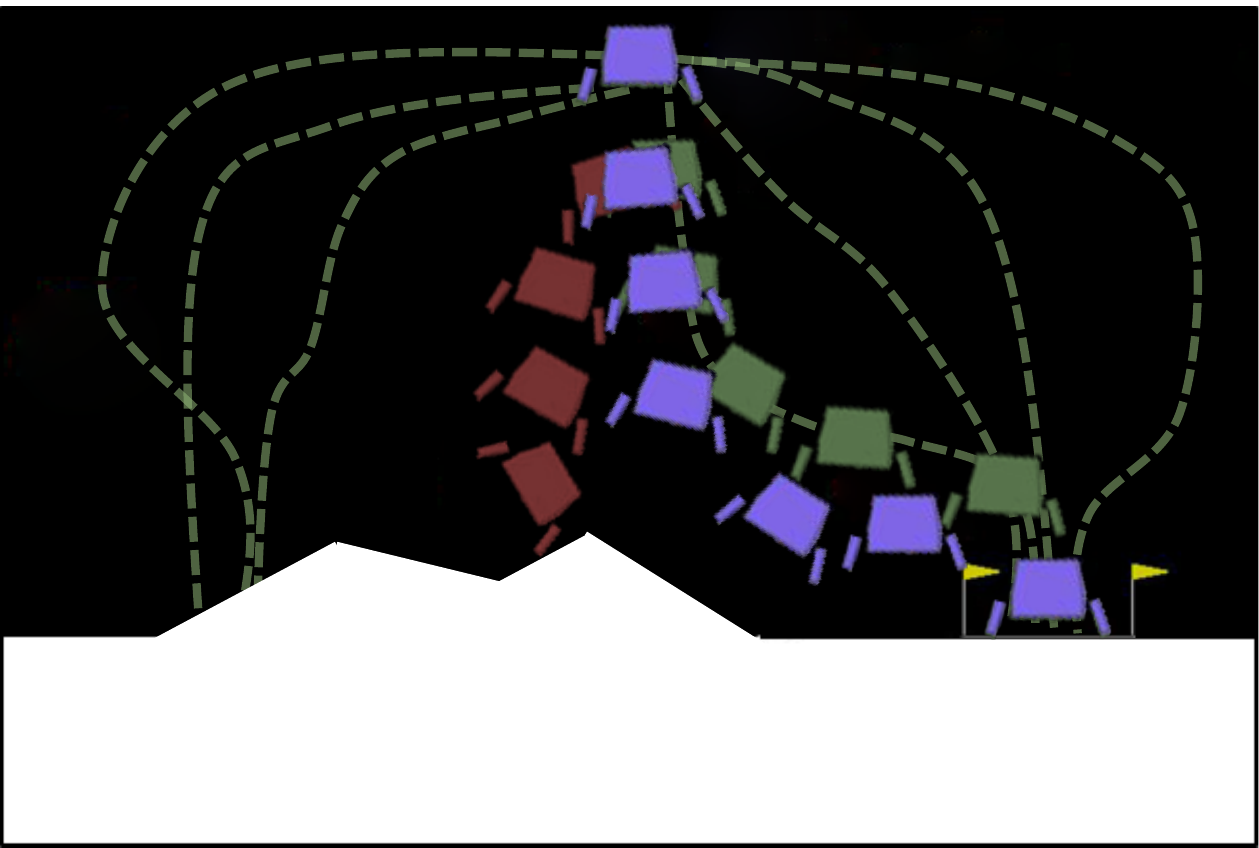}\label{fig:illustration2}
    \caption{Our framework utilizes a diffusion model to adapt a user's action (red) to those from a demonstration distribution (green) in a manner (blue) that balances a user's desire to maintain control authority with the benefits (e.g., safety) of conforming to the desired (demonstration) distribution. Without knowledge of the user's specific goal (e.g., the landing location), the demonstration distribution reflects different goals that the demonstration trajectories previously reached.}\label{fig:model_intuition}
\end{figure}
Shared autonomy~\cite{abbink18} is a framework in which a human user (also referred to as the \textit{pilot}) performs a task with an assistance of an autonomous agent (also referred to as the \textit{copilot})~\citep{goertz63, rosenberg93, Aigner97, dragan12, dragan13, gombolay14, billings2021towards}. 
The role of the agent is to complement the control authority of the user, whether to improve the robot's  performance on the current task or to encourage/ensure safe behavior. An important consideration when providing assistance via shared autonomy is the degree to which the agent balances the user's preference for maintaining control authority (i.e., the \emph{fidelity} of the assisted behavior relative to the user's actions), and the potential benefits of endowing more control to the agent (i.e., the \emph{conformity} of the assisted behavior to that of an autonomous agent).

A core difficulty of shared autonomy lies in the fact that the user's goal (intent) is typically not known. Many approaches to shared autonomy assume that there is a fixed, discrete set of candidate goals and seek to infer the user's specific goal at test-time based on observations, including the user's control input~\cite{muelling17, javdani15, perez15, hauser13, dragan13}. Such assumptions may be reasonable in structured environments (e.g., in the context of a manipulation task when there is a small number of graspable objects sitting on a table). However, they can be limiting in unstructured environments that lack well-defined goals or that have a very large set of potential goals.

Bootstrapped by function approximation with neural-networks, recent deep reinforcement learning (RL) algorithms seek to learn assistive policies without assumptions on the
knowledge of the goal space, or the assumption that the environment dynamics are known.
\citet{sha-via-deeprl} propose a deep RL approach to shared autonomy for domains with discrete actions, nominally relying on reward feedback from the user as an alternative to assuming that the goal space is known. In an effort to balance the user's control authority with task performance, the assistant chooses the action most similar to that of the user while also satisfying a state-action value constraint.
\citet{rsa} treat the \text{copilot} as providing a residual that is added to the user's actions to correct for unsafe behavior. They train their model to minimize the norm of the residual, subject to a goal-agnostic reward constraint that represents safe behavior.

These methods treat the pilot as a part of the environment, using an augmented state that includes the user's action. %
Framing the problem in this way has a clear and significant advantage---it enables the direct utilization of the modern suite of tools for deep RL. However, these methods have two notable limitations. First, they  nominally require human-in-the-loop interaction during training in order to generate user actions while learning the assistant's policy. Since the sample complexity of deep RL makes this interaction intractable, these methods replace the human with a \emph{surrogate policy}. If this surrogate is misspecified or invalid, this approach can lead to copilots that are incompatible with actual human pilots~\citep{rsa}. Second, these methods require access to task-specific reward during training, which may be difficult to obtain in practice.

In light of these limitations, we propose a model-free approach to shared
autonomy that interpolates between the user's action and an action sampled from
a generative model that provides a distribution over desired behavior
(Fig.~\ref{fig:model_intuition}).\footnote{For video and code, see \url{https://diffusion-for-shared-autonomy.github.io}.} Our approach has the distinct advantage that
it does not require knowledge of or access to the user's policy or any reward
feedback during training, thus eliminating the need for reward engineering.
Instead, our training process, which involves learning the generative model, only requires access to
trajectories that are representative of desired behavior.

The generative model that underlies our approach is a diffusion model~\citep{vincent11,thermo,song19,ddpm}, which has proven highly effective for complex generation tasks including image synthesis~\citep{dhariwal2021diffusion, saharia2022photorealistic}. Diffusion models consist of two key processes: the \textit{forward process} and the \textit{reverse process}.
The forward process iteratively adds Gaussian noise to the input with an increasing noise scale, while the reverse process is trained to iteratively denoise a noisy input in order to arrive at the target distribution. 
As part of this denoising process, the model produces the gradient direction in which the likelihood of its input increases under the target distribution.
Once the model is trained, generating a sample from the (unknown) target distribution involves running the reverse process on a sample drawn from a zero-mean isotropic Gaussian distribution.

\begin{figure}[!t]
    \centering
    \includegraphics[width=\linewidth]{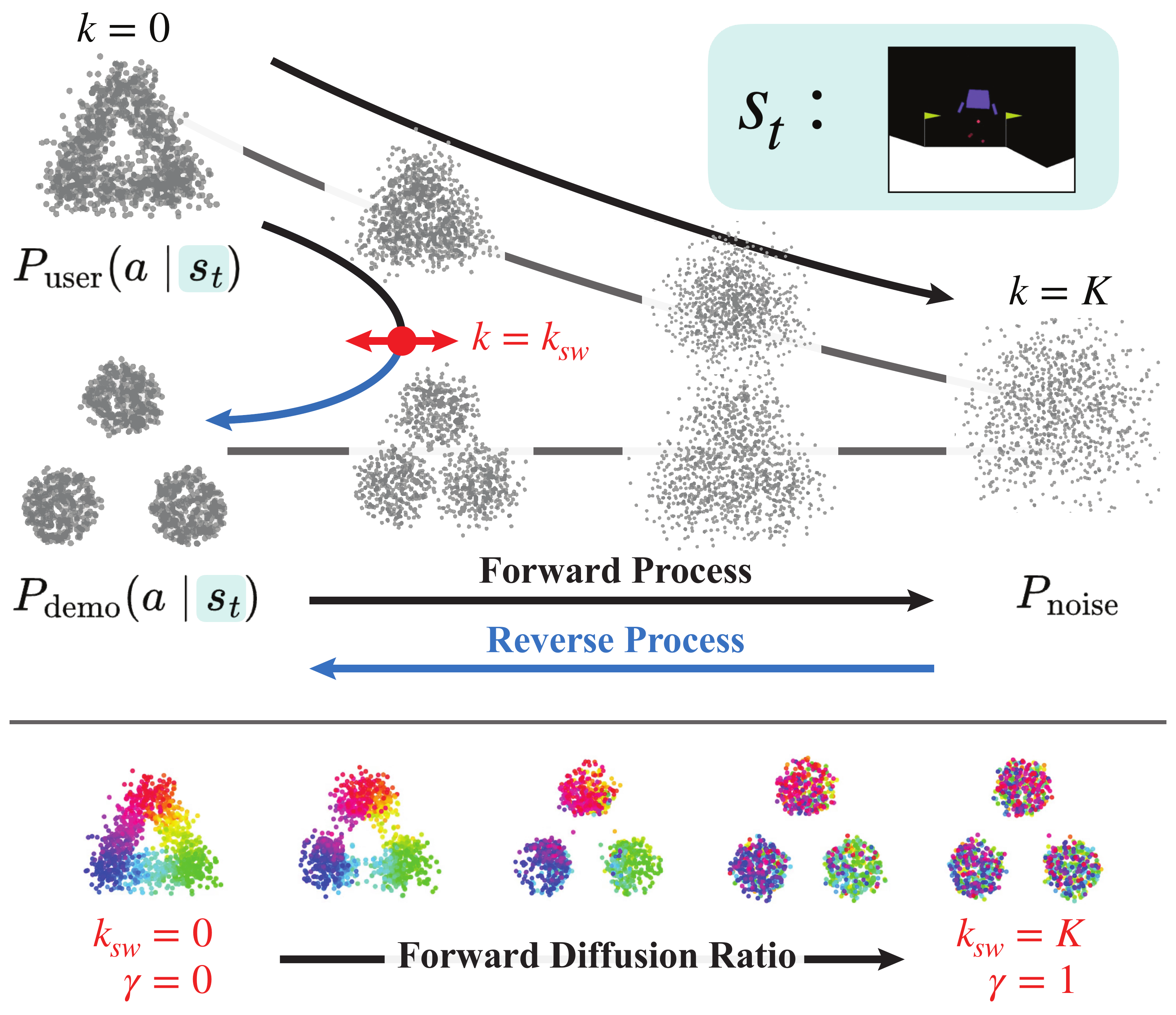}
    \caption{(Top) A visualization of diffusion processes of action distributions at state $s_t$. The black arrow at the top shows the forward diffusion of a source user distribution $P_\textrm{user}$ and the blue arrow below shows the reverse diffusion to the target demonstration distribution $P_\textrm{demo}$. We switch these two processes in the intermediate step $k=k_\textrm{sw}$ to achieve partial forward and reverse diffusion shown in the black and blue arrow on the left. (Bottom) The result of forward and reverse diffusion for different switching times $k_\textrm{sw}$, where standard reverse diffusion process corresponds to $k_\textrm{sw} = K$.}
    \label{fig:diffusion-visualization}
\end{figure}

As we will see in the following sections, a direct use of diffusion models for shared autonomy ends up in generating an action that ignores user's intent (i.e., low \emph{fidelity} to user intent), even though the action would be consistent with the desired behaviors (i.e., high \emph{conformity} to the target behaviors). To address this, we propose a new algorithm that 
controls the effect of the forward and reverse process through a \emph{forward diffusion ratio} $\fwr$ %
that regulates the balance between the fidelity and the conformity of the generated actions. The forward diffusion ratio provides a formal bound on the extent to which the copilot's action deviates from that of the user.

We evaluate our shared autonomy algorithm using a series of continuous control
tasks. In each case, we demonstrate that our algorithm significantly improves
the performance of a variety of different pilots, and we analyze the effects of
a range of different forward diffusion ratios.

\section{Method}

An integral part of our shared autonomy approach is the utilization of diffusion models as a generative model
that serves to correct the actions of the user. As such, we begin this section with an in-depth review of diffusion models.

\subsection{Background on diffusion models}
\label{subsec:background-diff}

We first present the mathematical formulation of diffusion models, with a particular emphasis on the denoising diffusion probabilistic model (DDPM)~\cite{ddpm}, which we employ. 

A probabilistic diffusion model~\citep{thermo} is a type of generative model that is characterized by two processes (Fig.~\ref{fig:diffusion-visualization}): a \emph{forward diffusion process} and a \emph{reverse diffusion process}. The forward diffusion process is an iterative first-order Markov chain that adds noise to a sample from a data distribution $\bm{x}_0 \sim q(\bm{x}_0)$. Assuming a predefined sequential noise schedule of $K$ steps, $\beta_1, \ldots, \beta_K$ and $\alpha_k \coloneqq 1 - \beta_k$,
a single step in the forward process operates as
\begin{equation}
    \bm{x}_k = \sqrt{\alpha_k} \bm{x}_{k-1} + \sqrt{1 - \alpha_k} \bm{\epsilon}, ~ \bm{\epsilon} \sim \mathcal{N}(\bm{0}, \bm{I}).\label{eqn:forward-process-single-step}
\end{equation}

Applying this recursively, multiple steps of forward process can be written in a closed form as
\begin{equation}
    \bm{x}_k = \sqrt{\bar{\alpha}_k} \bm{x}_0 + \sqrt{ 1 - \bar{\alpha}_k} \bm{\bm{\epsilon}}, ~ \bm{\epsilon} \sim \mathcal{N}(\bm{0}, \bm{I}),
\end{equation}
where $\bar{\alpha}_k \coloneqq \Pi_{s=1}^k \alpha_s$.
Equivalently,
\begin{subequations}
    \begin{align}
        \label{eqn:fwd-single-step}
        q(\bm{x}_k \mid \bm{x}_{k-1}) &= \mathcal{N}(\bm{x}_k; \sqrt{\alpha_k} \bm{x}_0, (1 - \alpha_k) \bm{I}) \\
        \label{eqn:fwd}
        q(\bm{x}_k \mid \bm{x}_0) &= \mathcal{N}(\bm{x}_k; \sqrt{\bar{\alpha}_k} \bm{x}_0, (1 - \bar{\alpha}_k) \bm{I})
    \end{align}
\end{subequations}

Meanwhile, the \emph{reverse diffusion process} is a Markov chain that iteratively denoises a noisy input, starting from $\bm{x}_K \sim N(\bm{0}, \bm{I})$. 
The conditional distribution over the output at each step $q(\bm{x}_{k-1}|\bm{x}_k)$ is Gaussian. However, computing these intermediate steps depends upon the entire data distribution and is generally not tractable in closed-form. Thus, we train a model that approximates it as $p_\theta(\bm{x}_{k-1} \mid \bm{x}_k) = \mathcal{N}(\bm{x}_{k-1}; \bm{\mu}_\theta(\bm{x}_k, k), \sigma_k^2 \bm{I})$ to run the reverse process.
The probability $q(\bm{x}_{k-1} \mid \bm{x}_k)$ that we want to model becomes tractable when conditioned on $\bm{x}_0$. 
By Bayes' rule, we have
\begin{subequations}
    \begin{align}
    q(\bm{x}_{k-1} \mid \bm{x}_k, \bm{x}_0) &= q(\bm{x}_k \mid \bm{x}_{k-1}, \bm{x}_0) \frac{q(\bm{x}_{k-1} \mid \bm{x}_0)}{q(\bm{x}_k \mid \bm{x}_0)} \\
    \label{eqn:cond-rev-factored}
    &= q(\bm{x}_k \mid \bm{x}_{k-1}) \frac{q(\bm{x}_{k-1} \mid \bm{x}_0)}{q(\bm{x}_k \mid \bm{x}_0)} \\ 
    &= \mathcal{N}(\bm{x}_{k-1}; \tilde{\bm{\mu}}(\bm{x}_k, \bm{x}_0), \tilde{\beta_k}\bm{I})
    \end{align}
\end{subequations}
Crucially, Equation~\ref{eqn:cond-rev-factored} only contains known terms from the Gaussian distributions in Equations~\ref{eqn:fwd-single-step} and \ref{eqn:fwd}. 
In the last expression, we have dropped the exact forms of mean and covariance for notational brevity.

Intuitively, we can train the model $p_\theta(\bm{x}_{k-1} \mid \bm{x}_k)$ by minimizing its divergence from $q(\bm{x}_{k-1} \mid \bm{x}_k, \bm{x}_0)$ over samples from the data distribution $q(\bm{x}_0)$.
Utilizing the evidence lower bound of $\mathbb{E}_{q}[-\log p_\theta (\bm{x}_0)]$, we can formulate the loss as:
\begin{equation}
    \begin{split}
        \mathcal{L} &\coloneqq \mathbb{E}_{q}\left[\sum_{k=2}^{K-1} L_k -\log p_\theta (\bm{x}_0 \mid \bm{x}_1) + C \right],\\
        L_k &= D_\text{KL}\bigl(q(\bm{x}_{k-1} \mid \bm{x}_k, \bm{x}_0) \Vert p_\theta(\bm{x}_{k-1} \mid \bm{x}_k)\bigr),  
    \end{split}
\end{equation}
where $D_\text{KL}$ is the Kullback-Leibler (KL) divergence, $C$
is a constant that corresponds to fixed parameters of the forward diffusion process~\cite{ddpm}. The term $L_k$ is the KL divergence between two Gaussian distributions, which can be computed in closed form as the squared error between two means
\begin{equation}
    \label{eqn:factored-loss}
    L_k = \mathbb{E}_{\bm{x}_k \sim q} \left[ \frac{1}{2\sigma^2_k} \lVert \tilde{\bm{\mu}}(\bm{x}_k, \bm{x}_0) - \bm{\mu}_\theta(\bm{x}_k, k) \rVert_2^2 \right] + C^\prime,
\end{equation}
where $C^\prime$ is a constant independent of $\theta$. Now we note the fact that predicting $\hat{\bm{x}}_{k-1} = \bm{\mu}_\theta(\bm{x}_k, k)$ given $\bm{x}_k$ is equivalent to predicting the noise $\bm{\epsilon}$ that was added to $\bm{x}_{k-1}$ (see Equation~\ref{eqn:forward-process-single-step}). We can thus reformulate the problem as one of minimizing the error in the noise prediction. By parameterizing $\bm{\mu}_\theta(\bm{x}_k, k)$ as a function of predicted noise $\bm{\epsilon}_\theta(\bm{x}_k, k)$, we can simplify Equation~\ref{eqn:factored-loss} to
\begin{equation}
\mathbb{E}_{q, \bm{\epsilon} \sim \mathcal{N}(\bm{0}, \bm{I})} \left[ c_k(\beta_{1:K}, \sigma_k) \lVert \bm{\epsilon} - \bm{\epsilon}_\theta(\bm{x}_k, k) \rVert_2^2 \right],
\end{equation}
where $c_k(\beta_{1:K}, \sigma_k)$ is a constant computed from the beta schedule for each timestep $k$.
\citet{ddpm} further simplify this equation to arrive at the final loss that, with a slight abuse of notation, becomes
\begin{subequations} \label{eqn:ddpm-loss}
    \begin{align}
        \mathcal{L}_\text{simple} &\coloneqq \mathbb{E}_{k, \bm{x}_0, \bm{\epsilon} \sim \mathcal{N}(\bm{0}, \bm{I})} \left[\lVert \bm{\epsilon} - \bm{\epsilon}_\theta(\bm{x}_k(\bm{x}_0, \bm{\epsilon}), k)\rVert_2^2\right] \\
        \bm{x}_k(\bm{x}_0, \bm{\epsilon}) &\coloneqq \sqrt{\bar{\alpha}_k} \bm{x}_0 + \sqrt{1 - \bar{\alpha}_k}\bm{\epsilon},
    \end{align}
\end{subequations}
where $k$ is sampled uniformly as $k \in [1, K]$.
After training, we can generate a sample $\bm{x}_0$ by running the reverse diffusion process recursively from $\bm{x}_K \sim \mathcal{N}(\bm{0}, \bm{I})$
\begin{equation}
    \bm{x}_{k-1} = \mu_{\theta^*}(\bm{x}_k, k) + \sigma_k \bm{z},%
\end{equation}
where $z \sim \mathcal{N}(\bm{0}, \bm{I})$ for $k > 1$, else $\bm{z} = \bm{0}$.

\subsection{Diffusion model guidance for distribution transformation}
\label{subsec:fwd-and-rev}

Consider generally the problem of transforming one distribution $P_\src$ onto another distribution $P_\tgt$. This problem is challenging, because finding such a correspondence between distributions has an inherent dependence on the intrinsic geometrical match between the two distributions. However, there is an enticing shortcut that presents itself in the form of diffusion processes. In particular, we note that the forward diffusion process should allow us to map $P_\src$ into a noise distribution $P_\text{noise}$ (i.e., zero-mean isotropic Gaussian noise). Drawing samples from this noise distribution, we can execute the reverse diffusion process to guide the sample to being drawn from $P_\tgt$, assuming of course that the reverse process was trained on data samples from $P_\tgt$.

Although this diffusion process will ultimately take points $\bm{x}_\src$ from $P_\src$ and map to samples drawn from $P_\tgt$, resulting in $\hat{\bm{x}}_\tgt$, it will do so in a destructive manner. In particular, as the forward diffusion process is applied to $\bm{x}_\src$, the distribution of these points becomes progressively more random (by design). In the end, its distribution reaches an isotropic Gaussian distribution, completely corrupting the information in the original sample. At this point, there is no relationship between the source points $\bm{x}_\src$ and the derived points $\hat{\bm{x}}_\tgt$. We may as well have simply started from Gaussian noise and run reverse diffusion onto $P_\tgt$.

We would like to learn a transformation $\mathcal{F}: \bm{x}_\src \to \hat{\bm{x}}_\tgt$ that preserves information. Such a transformation between distributions can be achieved by running the forward diffusion process on $P_\src(x)$ \textit{partway through}, up to $k=k_\textrm{sw}$, followed by the reverse diffusion process for the same number of steps ($k_\textrm{sw}$), in order to transform these partially diffused points onto $P_\tgt$.
Crucially, this process trades off the amount of information present in $\bm{x}_\src$ that is preserved as a result of forward diffusion, and the consistency (in terms of likelihood) of the sample generated via reverse diffusion with respect to the target distribution $P_\tgt$ (i.e., the distribution over desired behaviors).

Very recently, \citet{sdedit} proposed a similar partial diffusion process for the task of generating realistic images based upon a user-provided sketch. The authors derive a bound on the distance between $\hat{\bm{x}}_\tgt$ and $\bm{x}_\src$ as follows.\footnote{Their bound is for a variance exploding (VE) formulation of diffusion, whereas we employ DDPM that uses a variance preserving (VP) formulation. Despite the difference, they share the same mathematical intuition.}
Assuming that 
$\lVert \bm{\epsilon}_\theta(\bm{x}, k) \rVert \leq \mathcal{K}$
for all $\bm{x}\in X$ and $k \in [0, K]$, then for all $\delta \in (0, 1)$ with probability at least $(1-\delta)$,
\begin{equation}\label{eqn:displacement-bound}
    \lVert \bm{x}_\src - \hat{\bm{x}}_\tgt \rVert^2_2 \leq \sigma_{k_\text{sw}}^2 (\mathcal{K} \sigma_{k_\text{sw}}^2 + d + 2 \sqrt{-d \cdot \log \delta} - 2 \log \delta),
\end{equation}
where $d$ is the dimensionality of $\bm{x}$. %
From this bound, we can see that the distance between $\bm{x}_\src$ and $\hat{\bm{x}}_\tgt$ increases with $k_\textrm{sw}$, since $\sigma_{k_\textrm{sw}}$ increases with $k_\textrm{sw}$ while other terms are not affected. This expression allows us to bound the difference between the action generated by the assistant $\hat{\bm{x}}_\tgt$ and that of the user $\bm{x}_\src$.

Figure~\ref{fig:diffusion-visualization} visualizes this property in two dimensions. 
Here, temporarily ignoring state condition, samples from $P_\textrm{user}$ (i.e., source distribution $P_\src$) form the shape of a triangle, and 
samples from $P_\textrm{demo}$ (i.e., target distribution $P_\tgt$) follow a distribution with three modes, each centered on one of the vertices of a triangle. The black and blue arrow represents the idea of executing partial diffusion---running the forward (black) and reverse (blue) processes for $k_\textrm{sw}$ steps.
At the bottom of Figure~\ref{fig:diffusion-visualization}, we visualize the result of partial forward and reverse diffusion for different values of $k_\textrm{sw}$. Here, we color points based on their original spatial location in $P_\src$ (e.g., green points in the lower-right, red at the top, and blue at the lower-left), and track their location over different degrees of forward and reverse diffusion. Based on the visualization of the resulting distributions, we see that the distance between the initial points $\bm{x}_\src$ and $\hat{\bm{x}}_\tgt$ increases with $k_\textrm{sw}$, consistent with Equation~\ref{eqn:displacement-bound}. More generally, we find that:

\noindent\emph{\textbf{When $k_\textrm{sw}$ is small}}
\begin{itemize}
    \item The original information is well-preserved (small displacements; high \emph{fidelity})
    \item The obtained distribution is far from $P_\tgt$ (low \emph{conformity})
\end{itemize}

\noindent\emph{\textbf{When $k_\textrm{sw}$ is large}}
\begin{itemize}
    \item The original information is corrupted (large displacements; low \emph{fidelity})
    \item The obtained distribution is close to $P_\tgt$ (high \emph{conformity})
\end{itemize}

To discuss the effect of $k_\textrm{sw}$ independent of the number of diffusion steps $K$, we herein define the \textit{Forward Diffusion Ratio} as $\fwr \coloneqq k_\textrm{sw} / K$, and will refer to this throughout the paper.
\begin{algorithm}[t]
    \caption{Shared autonomy as partial diffusion}\label{alg:gen-fwd-rev}
    \small
    \textbf{Input:} Observation $\bm{s}_t$ and pilot's action $\bm{a}^h_t$ \\
    \textbf{Output:} Shared action $\bm{a}^s_t$\\
    \textbf{Require:} A pretrained state-conditioned denoising model $\mu_{\theta^{*}}(\bm{a}, k \mid \bm{s})$, forward diffusion ratio $\fwr$, diffusion timestep $K$
    \begin{algorithmic}[1]
        \small
        \State Compute the switching timestep $k_\sw \leftarrow \text{ToInteger} (\fwr K)$
        \State Sample a Gaussian noise $\bm{\epsilon} \sim \mathcal{N}(\bm{0}, \bm{I})$
        \State $\tilde{\bm{a}}^h_{t, k_\sw} \leftarrow \sqrt{\bar{\alpha}_{k_\sw}} \bm{a}^h_t + \sqrt{1 - \bar{\alpha}_{k_\sw}} \bm{\epsilon}$ \Comment{Forward process}
        \State $\bm{a}^s_{t, k_\sw} \leftarrow \tilde{\bm{a}}^h_{t, k_\sw}$
        \For{$k$ in $k_\sw,\dots, 2$}
        \State Sample a noise vector $\bm{z} \sim \mathcal{N}(\bm{0}, \bm{I})$
        \State $\bm{a}^s_{t, k-1} \leftarrow \bm{\mu}_{\theta^{*}}(\bm{a}^s_{t, k}, k \mid \bm{s}_t) + \sigma_k \mathbf{z}$  \Comment{Reverse process}
        \EndFor
        \State $\bm{a}^s_{t, 0} \leftarrow \bm{\mu}_\theta(\bm{a}^s_{t, 1}, 1 \mid \bm{s}_t)$
        \State return $\bm{a}^s_{t, 0}$
    \end{algorithmic}
\end{algorithm}

A similar idea in diffusion models has been adopted in image generation,
manipulation, and 3D geometry generation~\citep{sdedit, sjc, dreamfusion,
ddnm}. %

\subsection{Distribution transformation for shared autonomy}
\label{subsec:task-setting}

\begin{figure*}[!th]
    \centering
    \subfloat[2D Control]{\includegraphics[height=3.25cm]{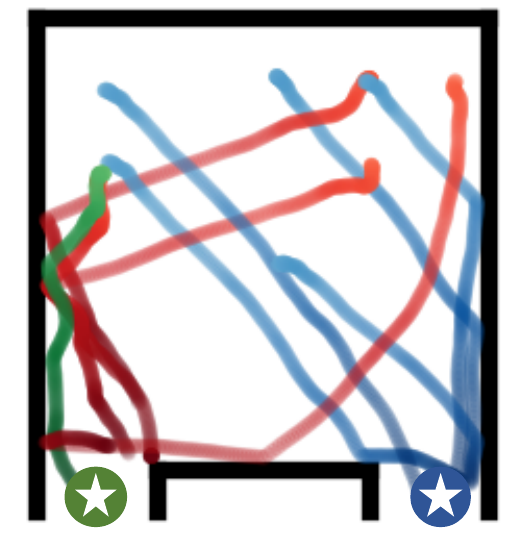}\label{fig:maze_env_render}} %
    \hfil
    \subfloat[Lunar Lander]{\includegraphics[height=3.25cm]{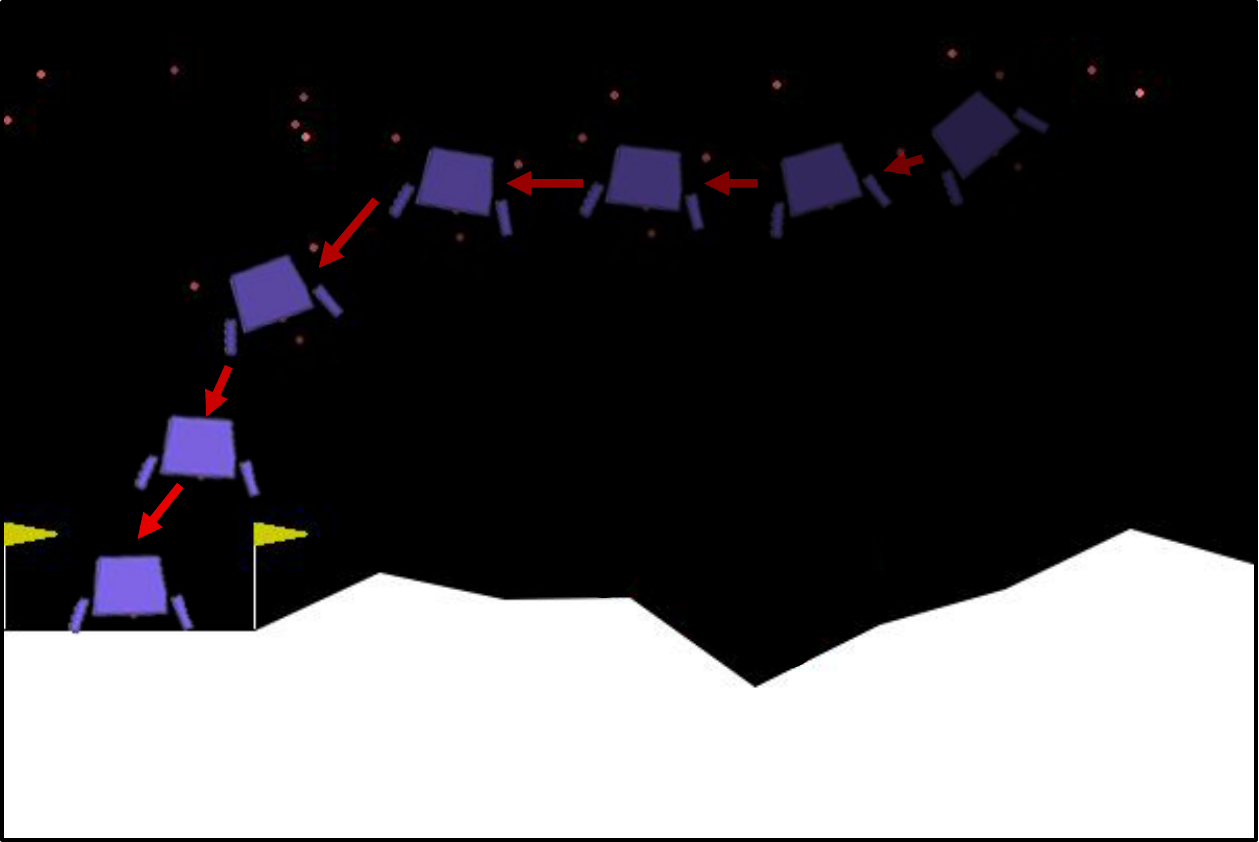}\label{fig:ll-lander_env_render}}
    \hfil
    \subfloat[Lunar Reacher]{\includegraphics[height=3.25cm]{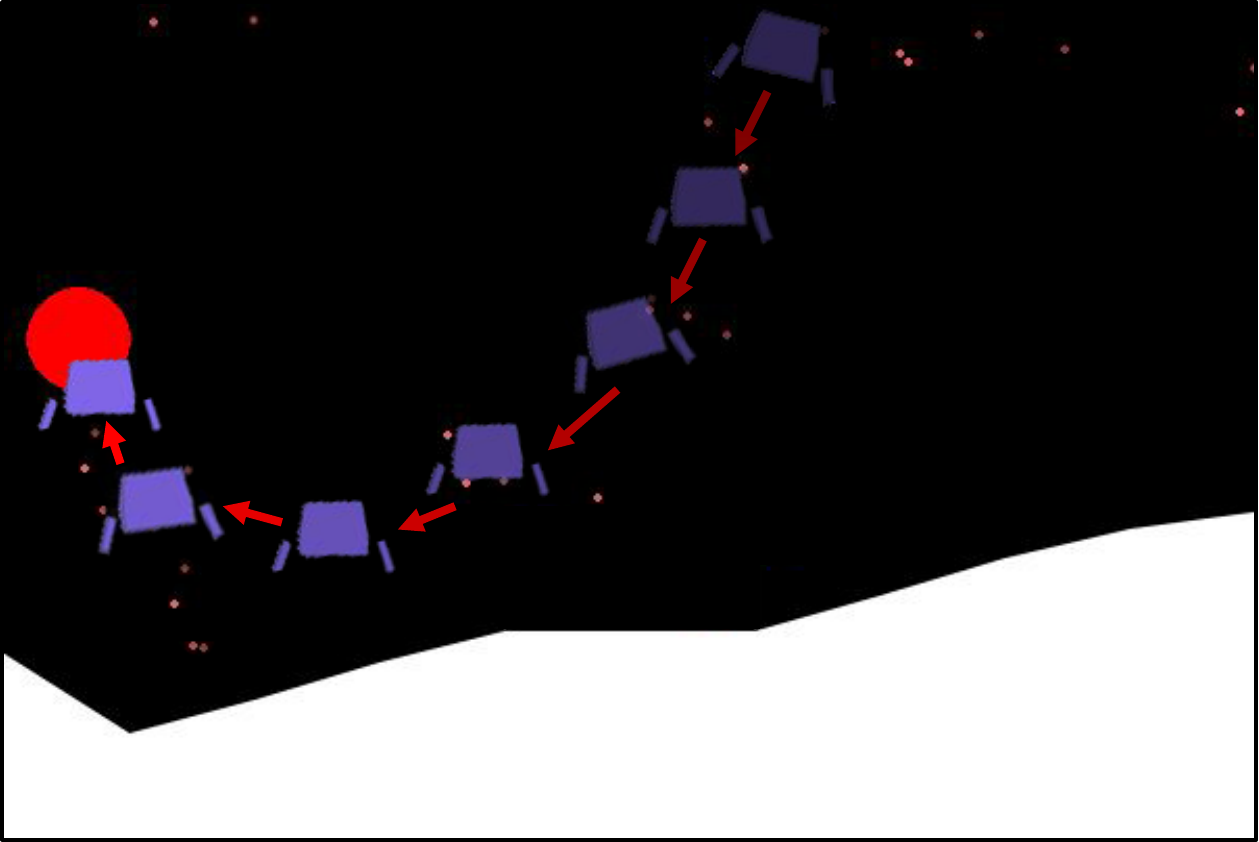}\label{fig:ll-reacher_env_render}}%
    \hfil
    \subfloat[Block Pushing]{\includegraphics[height=3.25cm]{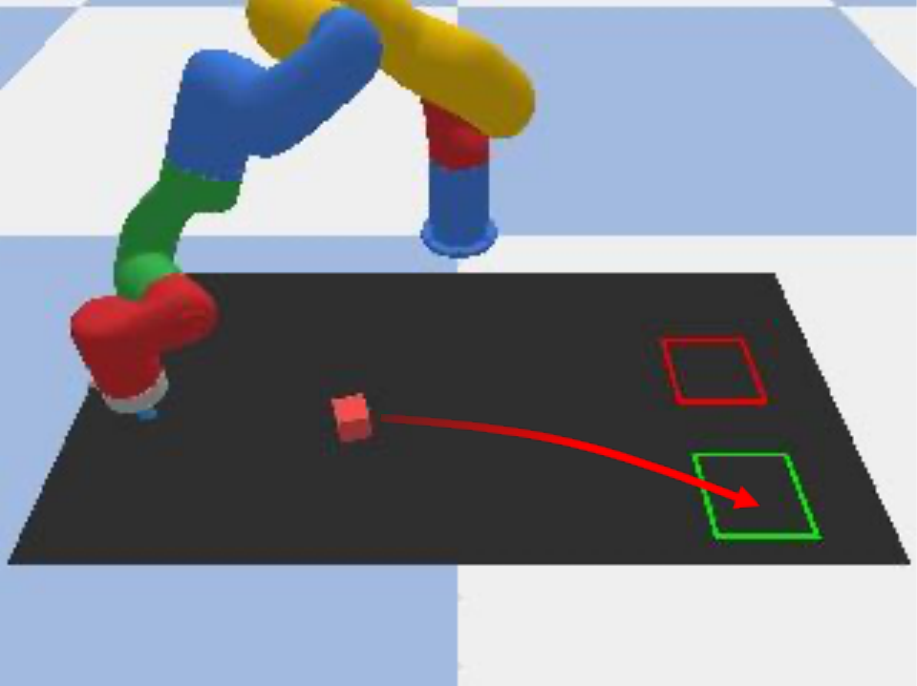}\label{fig:bp_env_render}}%
    \caption{We evaluate our algorithm in the context of four shared autonomy environments including a \subref{fig:maze_env_render} 2D Control task in which an agent navigates to one of two different goals,  \subref{fig:ll-lander_env_render} Lunar Lander that tasks a drone with landing at a designated location, \subref{fig:ll-reacher_env_render} a Lunar Reacher variant in which the objective is to reach a designated region in the environment, and \subref{fig:bp_env_render} Block Pushing, in which the objective is to use a robot arm to push an object into one of two different goal regions.}\label{fig:four_exp_env}
\end{figure*}
In a shared autonomy task, a copilot is asked to produce a shared action $a^s_t$, 
given the current state $s_t$ and the pilot's action $a_t^h$.
The goal is for the copilot to intervene in a manner that preserves the pilot's "intention" while correcting the action. 
 However, formulating the pilot's "intention" is challenging, especially when the space of goals is unknown or not well-defined.

Our algorithm assumes that the latent intent of the user has non-zero likelihood under the distribution over target behaviors that we learn to sample from using a diffusion model. In other words, if we have a finite set of demonstrations to train the diffusion model, the states or state-action pairs that a user would want to achieve need to exist in the demonstrations.
With our formulation to shared autonomy, the copilot manages the trade-off between
respecting pilot's action (i.e., \emph{fidelity} to the pilot) 
and executing an action that is likely under the learned behavior distribution (i.e., \emph{conformity} to the target behaviors). This is illustrated in Fig. \ref{fig:trade-off}, and it bears similarity to the trade-off discussed in Section~\ref{subsec:fwd-and-rev}
 based on Figure~\ref{fig:diffusion-visualization}.
 
Now, given a state $s_t$,
$P_\textrm{user}$ in Figure~\ref{fig:diffusion-visualization} is an (unknown) distribution over pilot actions 
and $P_\textrm{demo}$ is a distribution over target behaviors where each point represents a single action.
We can interpret that each cluster in $P_\textrm{demo}$
corresponds to a set of actions, each of which causes distinct transitions. Hence, each cluster can be thought of as a pilot's ``intention'', 
and the actions outside of the cluster as being undesirable. %
Considering the visualization of the different partially forward and reverse diffused distributions at the bottom of the figure, we see that as we increase $\fwr$ from $0.0$ to $1.0$, most of the pilot's actions begin to go inside of the set of ideal %
actions.
This captures the notion that conformity to the target behavior increases with $\fwr$.
On the other hand, once $\fwr$ exceeds $0.5$, 
the displacement of pilot actions (signified by the mixture of colors)
begin to enlarge, which is equivalent to the copilot producing an action that ignores the pilot's ``intention''. This captures the notion that the fidelity to the pilot decreases as we increase $\fwr$.

Algorithm \ref{alg:gen-fwd-rev} summarizes the procedure of applying partial forward and reverse diffusion to a pilot action $a^s_t$ and generating a shared action $a^s_t$. In Section~\ref{sec:experiments}, we investigate the fidelity-conformity trade-off by modulating $\fwr$ empirically for various environments with different pilots.

\section{Experiments}\label{sec:experiments}

\begin{table*}[ht]
    \centering
    \caption{Success and crash/out-of-bounds (OOB) rates on Lunar Lander and Lunar Reacher for different pilots with ($\gamma = 0.4$) and without assistance. Each entry corresponds to $10$ episodes across $30$ random seeds. Note that the Zero and Random pilots have no knowledge of the goal.}
    \label{tb:lunar-lander-reacher}
    \begin{tabularx}{1.0\linewidth}{lYYYYYYYY}%
        \toprule
        & \multicolumn{4}{c}{\bf Lunar Lander} & \multicolumn{4}{c}{\bf Lunar Reacher}\\
        \midrule
        & \multicolumn{2}{c}{Success Rate $\uparrow$} & \multicolumn{2}{c}{Crash/OOB Rate $\downarrow$} & \multicolumn{2}{c}{Success Rate $\uparrow$} & \multicolumn{2}{c}{Crash/OOB Rate $\downarrow$}\\
        Pilot & w/o Copilot & w/ Copilot & w/o Copilot & w/ Copilot & w/o Copilot & w/ Copilot & w/o Copilot & w/ Copilot\\
        \midrule
        Noisy & 
        $ 20.67 \pm 4.50 $ &  $ \bm{68.00 \pm 5.35} $ &
        $\hphantom{0} 28.33 \pm 2.62 $ &  $ \hphantom{0}\bm{7.67 \pm 2.87} $ &
        $ 14.33 \pm 2.49 $ &  $ \bm{45.33 \pm 3.30} $ &
        $ \hphantom{0}77.33 \pm 3.09 $ &  $ \bm{38.00 \pm 2.94} $\\
        Laggy & 
        $ 21.33 \pm 2.05 $ &  $ \bm{75.00 \pm 3.56} $ & 
        $\hphantom{0} 76.67 \pm 2.49 $ &  $ \hphantom{0} \bm{9.67 \pm 3.86} $ &
        $ 30.67 \pm 5.56 $ &  $ \bm{55.33 \pm 6.13} $ &
        $ \hphantom{0}69.33 \pm 5.56 $ &  $ \bm{31.33 \pm 2.87} $\\
        \midrule[0.1pt]
        Zero &
        $ \hphantom{0}0.00 \pm 0.00 $ &  $ 27.00 \pm 0.82 $ &
        $ 100.00 \pm 0.00 $ &  $ 19.00 \pm 2.94 $ &
        $ \hphantom{0}0.00 \pm 0.00 $ &  $ 19.67 \pm 2.62 $ &
        $ 100.00 \pm 0.00 $ &  $ 58.33 \pm 3.09 $\\
        Random & 
        $ \hphantom{0}0.00 \pm 0.00 $ &  $ 25.00 \pm 4.32 $ &
        $ 100.00 \pm 0.00 $ &  $ 19.33 \pm 5.25 $ &
        $ \hphantom{0}4.33 \pm 1.89 $ &  $ 22.33 \pm 2.87 $ &
        $ \hphantom{0}95.33 \pm 1.70 $ &  $ 59.00 \pm 0.82 $\\
        Expert &
        $ 77.67 \pm 2.62 $ &  $ 78.67 \pm 2.87 $ &
        $\hphantom{0} 12.33 \pm 0.94 $ &  $ \hphantom{0}8.00 \pm 1.63 $ &
        $ 49.33 \pm 4.78 $ &  $ 55.00 \pm 2.16 $ &
        $ \hphantom{0}44.00 \pm 3.56 $ &  $ 31.67 \pm 2.87 $\\   
        \bottomrule
    \end{tabularx}
\end{table*}

We evaluate our approach to shared autonomy by pairing our copilot with various pilots on a variety of simulated continuous control tasks (Fig.~\ref{fig:four_exp_env}): 2D Control, Lunar Lander, Lunar Reacher and Block Pushing. Each of these tasks provide the opportunity for the pilot to execute one of several different behaviors that is not known to the copilot. We design each domain to include a randomly sampled target state the pilot intends to reach (herein referred to as the \textit{goal}). 
For example, in Lunar Reacher, a goal location is sampled randomly above ground. Across all tasks, we reveal the goal only to the pilot, by including it as part of the state. The copilot never has access to the goal. This results in a scenario in which pilot's intention (i.e., the goal) is unknown to the copilot.

We seek to answer the following questions: (1) How much can our copilot assist a pilot? (2) Does our copilot generalize to different pilots? and (3) What is the effect of the forward diffusion ratio $\fwr$, and how can we interpret it?%

\subsection{Continuous control domains}

\begin{figure}[!th]
    \centering
    \subfloat[$\gamma=0.0$]{\includegraphics[height=3.0cm]{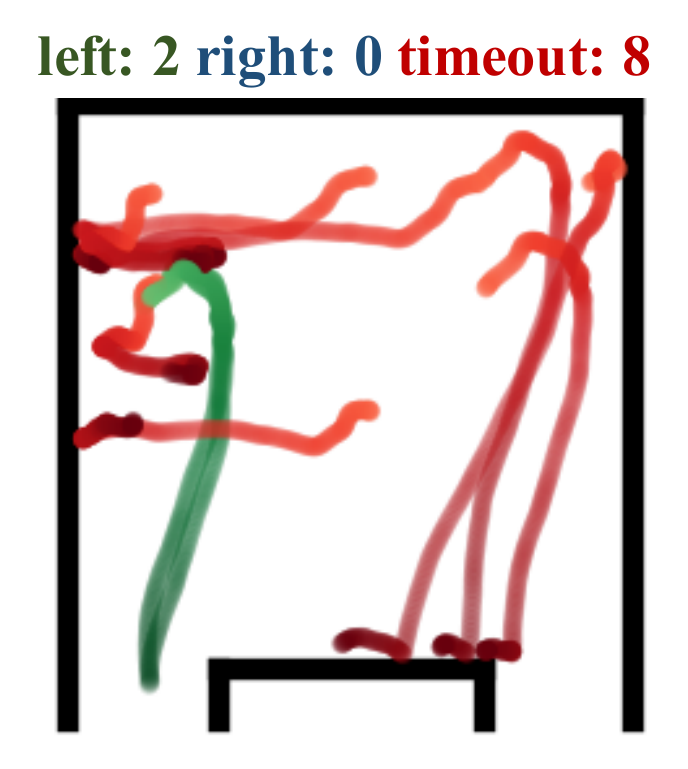}\label{fig:maze0}}
    \hfil
    \subfloat[$\gamma=0.1$]{\includegraphics[height=3.0cm]{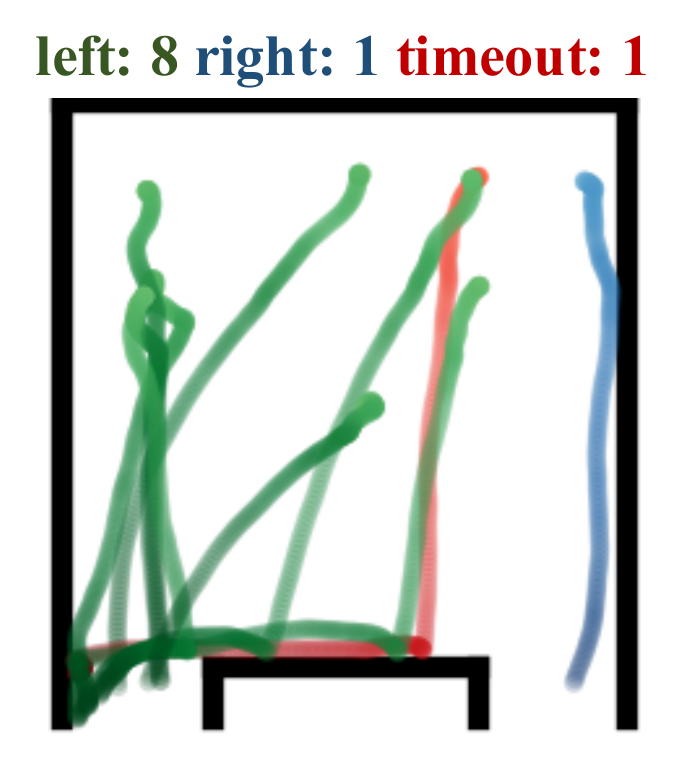}\label{fig:maze1}}
    \hfil
    \subfloat[$\gamma=0.2$]{\includegraphics[height=3.0cm]{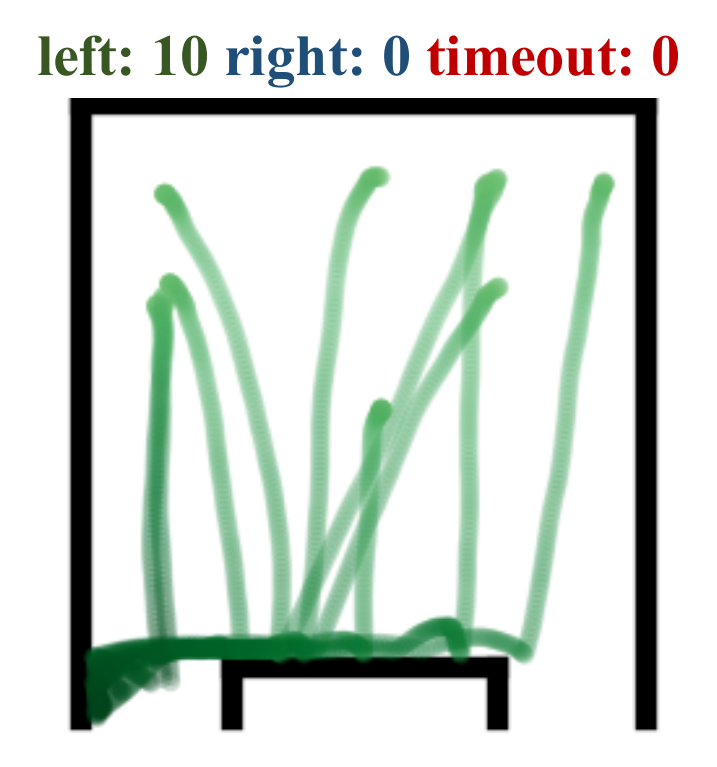}\label{fig:maze2}}\\
    \subfloat[$\gamma=0.4$]{\includegraphics[height=3.0cm]{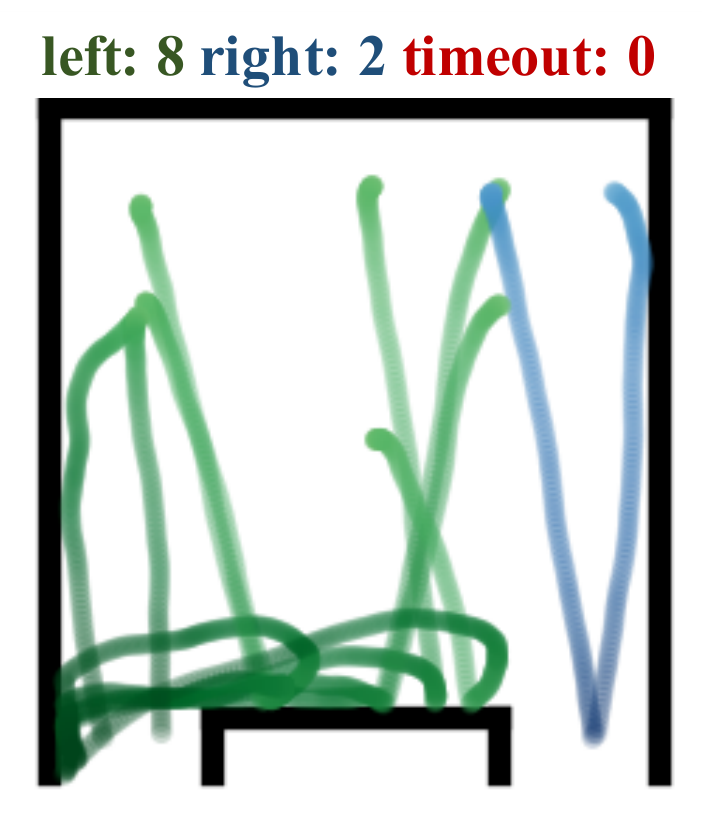}\label{fig:maze4}}
    \hfil
    \subfloat[$\gamma=0.6$]{\includegraphics[height=3.0cm]{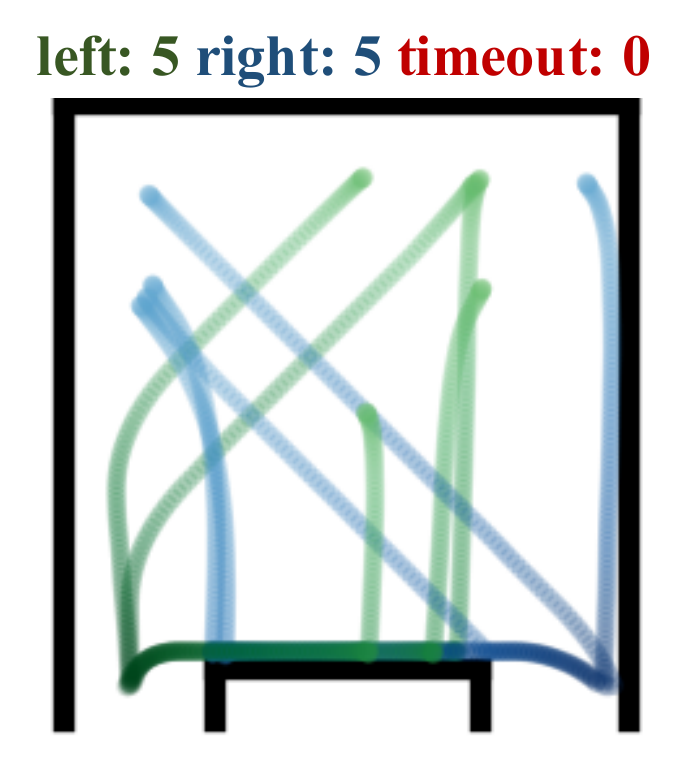}\label{fig:maze6}}
    \hfil
    \subfloat[$\gamma=0.8$]{\includegraphics[height=3.0cm]{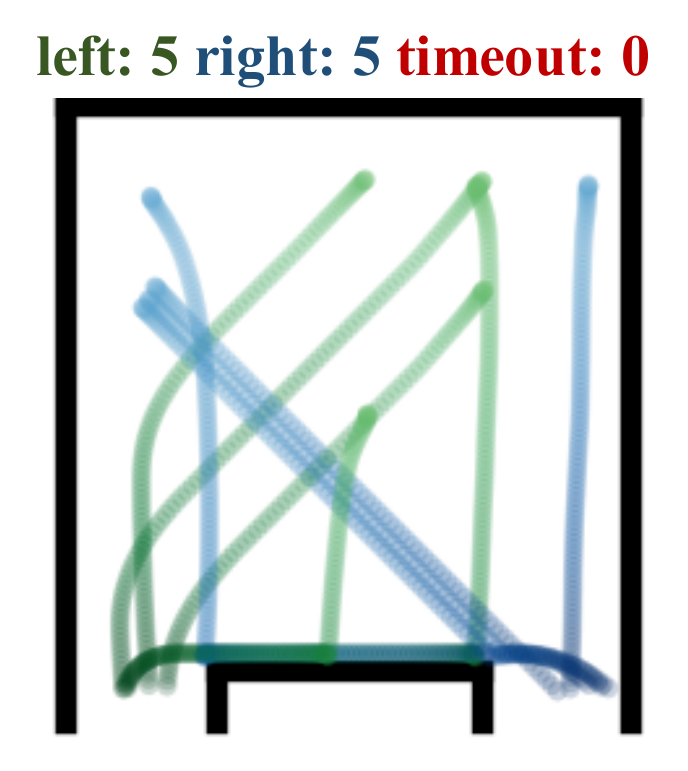}\label{fig:maze8}}
    \caption{A visualization of the resulting trajectories in the 2D Control environment for different settings for the forward diffusion ratio $\fwr$. The user's objective is to reach the left-hand goal. Without assistance \subref{fig:maze0} the user successfully reaches the goal two times, while the eight others timeout. As we increase $\fwr$, we see that \subref{fig:maze1}--\subref{fig:maze4} the user reaches the desired goal a vast majority of the time. As $\fwr$ gets closer to $1.0$, \subref{fig:maze6} \subref{fig:maze8} the assisted policy conforms to the expert policy, which avoids timeouts, but without knowledge of the user's goal distributes the trajectories evenly between the left and right goals.}%
    \label{fig:maze-vs-fwr}%
\end{figure}
\subsubsection{2D Control} 
We build a simple 2D continuous control domain based on the maze-2D environment in D4RL~\citep{fu2020d4rl} built on the MuJoCo simulator~\cite{mujoco}, where the goal is located either at the lower-left or lower-right corners of the large open space (Fig.~\ref{fig:maze_env_render}).
The agent is represented as a point mass, and the actions are 2D forces applied to itself. The state consists of the agent's location and velocity, as well as the goal location. Episodes are terminated if a timeout of $300$ steps is reached.

\subsubsection{Lunar Lander}
Lunar Lander (Fig.~\ref{fig:ll-lander_env_render}) is a continuous control environment adopted from Open AI Gym \cite{brockman16} that involves landing a spaceship on a landing pad. The actions consist of continuous left, right and upward forces that emulate thrusters on the spaceship.
The state contains the position, orientation, linear and rotational velocity of the spaceship, whether each leg touches the ground, and the landing pad location (provided only to the pilot).
An episode ends when the spaceship lands on the landing pad and becomes idle, crashes, flies out of bounds, or it reaches a timeout of $1000$ steps.
\begin{figure}[!t]
    \centering
    \subfloat[Noisy Pilot]{\includegraphics[width=0.48\linewidth]{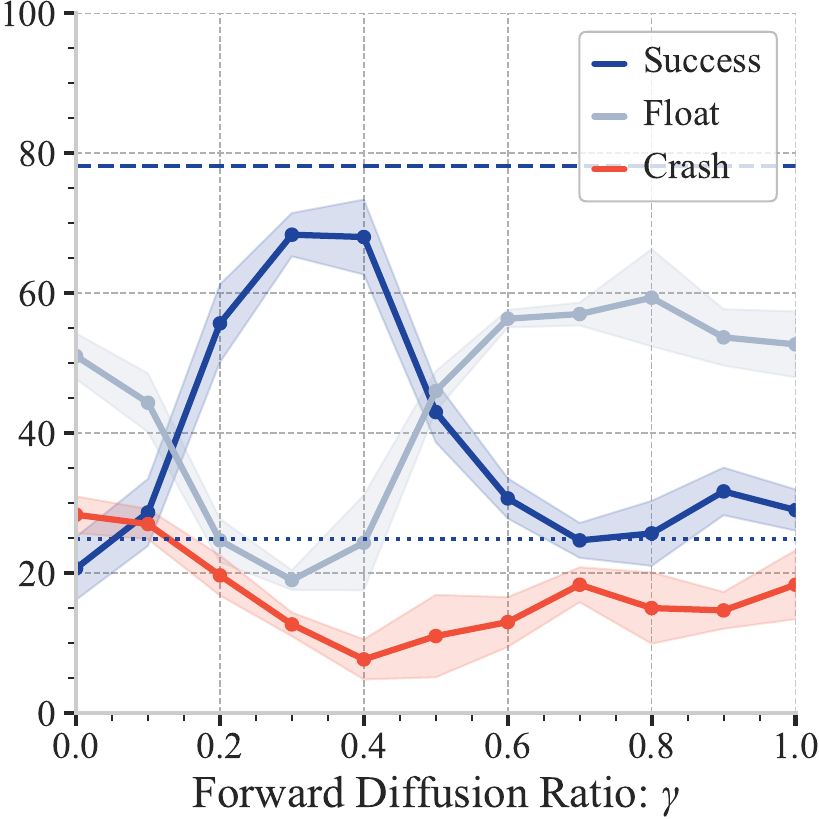}\label{fig:ll-land-assisted-noisy}}
    \hfil
    \subfloat[Laggy Pilot]{\includegraphics[width=0.48\linewidth]{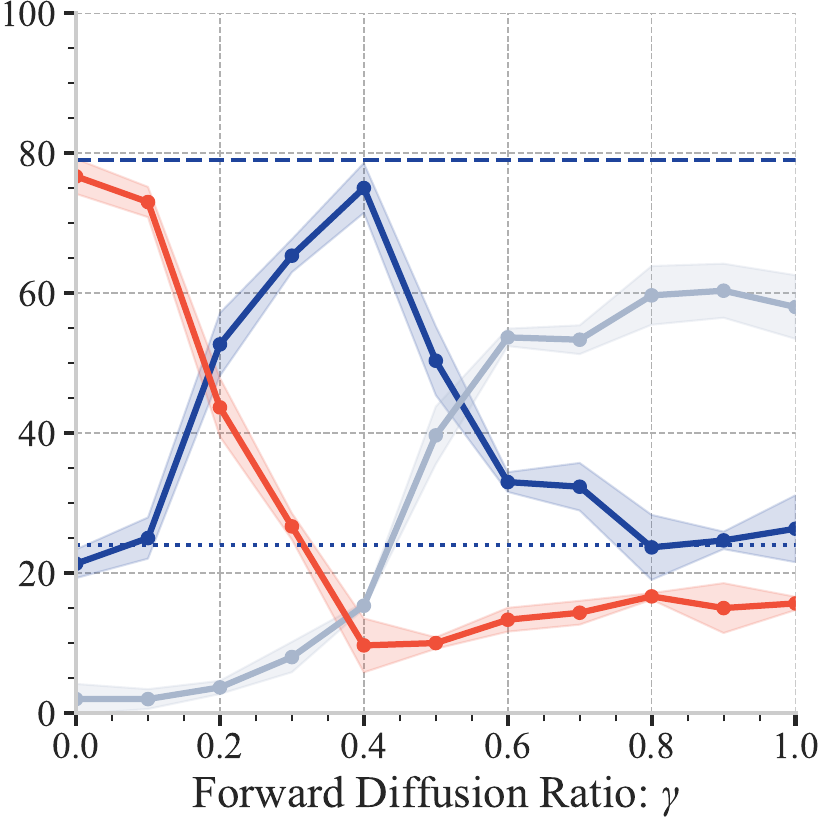}\label{fig:ll-land-assisted-laggy}}%
    \
    \subfloat[Noisy Pilot]{\includegraphics[width=0.48\linewidth]{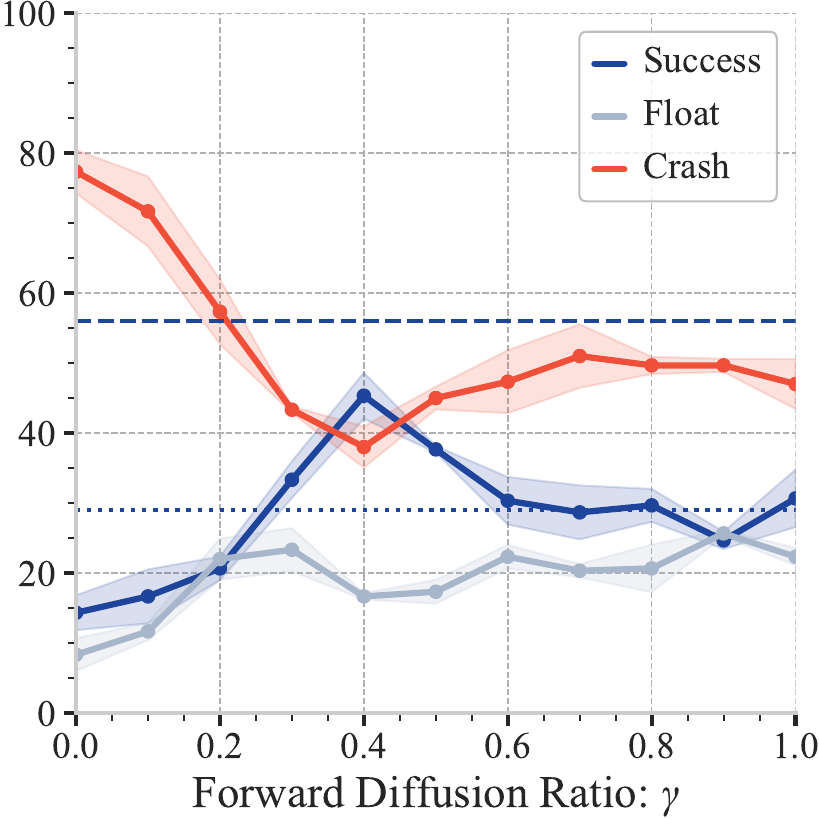}\label{fig:ll-reach-assisted-noisy}}
    \hfil
    \subfloat[Laggy Pilot]{\includegraphics[width=0.48\linewidth]{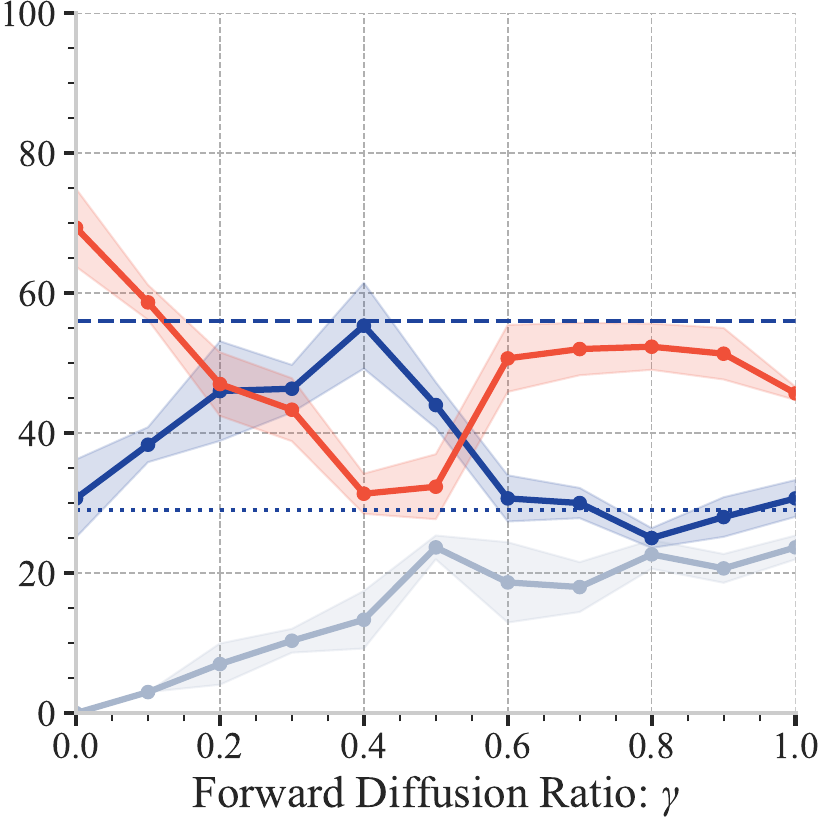}\label{fig:ll-reach-assisted-laggy}}%
    \caption{Success (higher is better), floating, and crash (lower is better) rates for Lunar Lander (top) and Lunar reacher (bottom) with \subref{fig:ll-land-assisted-noisy} \subref{fig:ll-reach-assisted-noisy} noisy and \subref{fig:ll-land-assisted-laggy} \subref{fig:ll-reach-assisted-laggy} laggy pilots. The dashed blue line denotes the success rate of an expert policy, while the dotted blue line denotes the success rate of our model with full-diffusion ($\fwr=1.0$).}\label{fig:ll-land-assisted}%
\end{figure}

\subsubsection{Lunar Reacher}
A variant of Lunar Lander adopted from previous work~\citep{rsa, brockman16}, where the goal is not to land, but to reach a random target location above the ground (Fig.~\ref{fig:ll-reacher_env_render}). The setting otherwise matches that of Lunar Lander.

\subsubsection{Block Pushing}
A variant of the Simulated Pushing environment~\cite{ibc}. The environment (Fig.~\ref{fig:bp_env_render}) consists of a simulated six-DoF robot xArm6 in PyBullet~\cite{coumans2021} equipped with a small cylindrical end effector. The task is to push an object into one of two target zones in the robot's workspace. The episode terminates when the target reaches one of the two locations or the number of steps exceeds a timeout of $100$. The position and orientation of the block and end effector are randomly initialized at the start of each episode, while the target locations are fixed.

\begin{figure}[!th]
    \centering
    \subfloat[Noisy Pilot]{\includegraphics[width=0.48\linewidth]{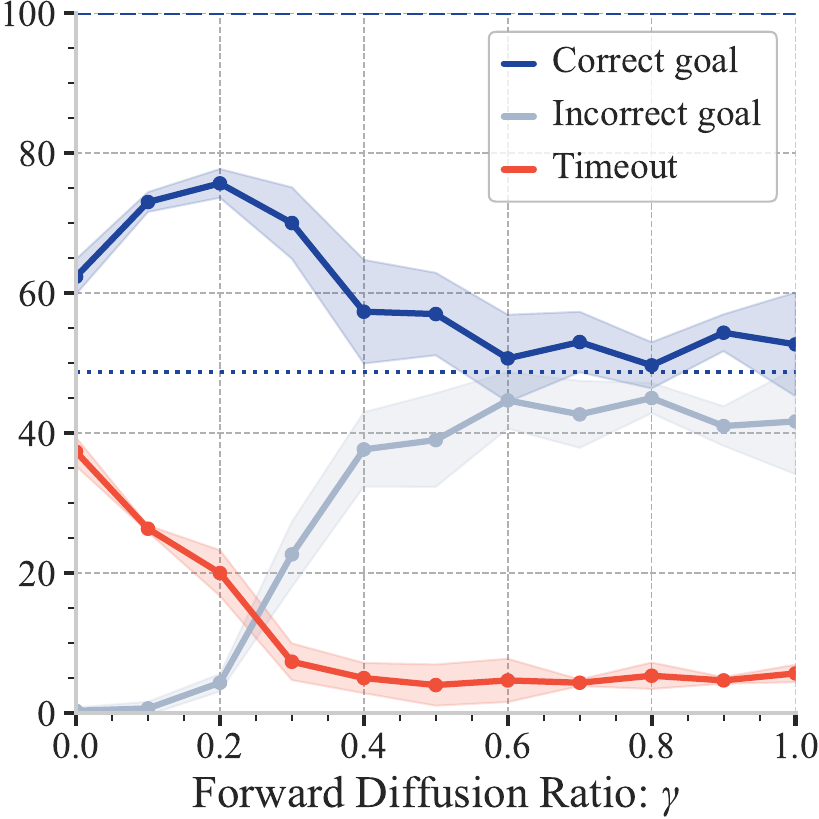}\label{fig:bp-assisted-noisy}}
    \hfil
    \subfloat[Laggy Pilot]{\includegraphics[width=0.48\linewidth]{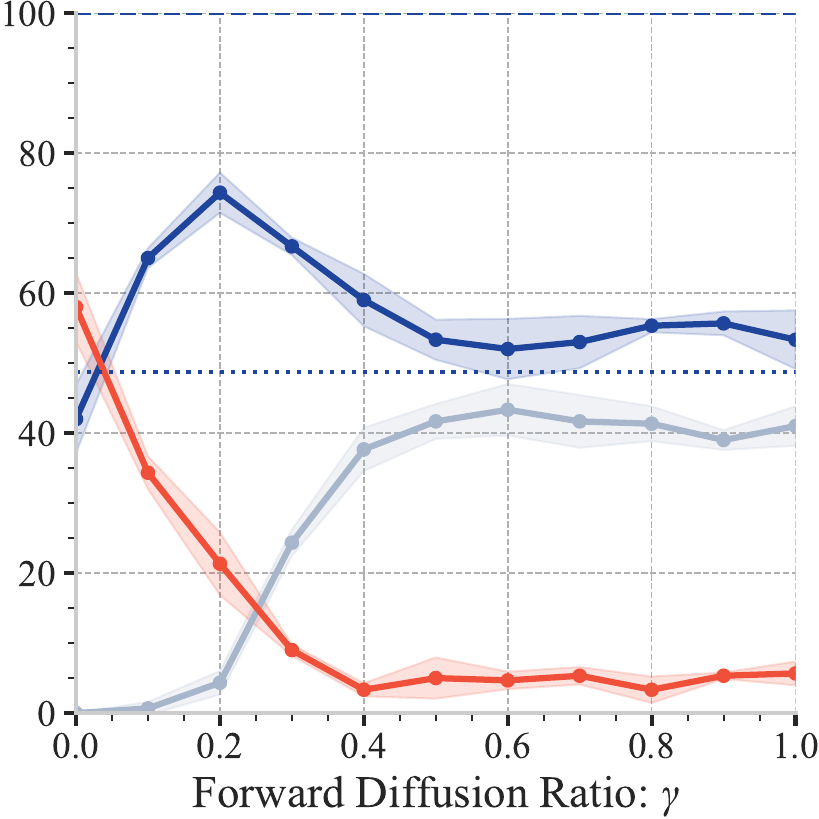}\label{fig:bp-assisted-laggy}}%
    \caption{The rates of block pushing task to terminate at correct goal, wrong goal or timeout with  \subref{fig:ll-reach-assisted-noisy} noisy and \subref{fig:ll-reach-assisted-laggy} laggy pilot. The dashed blue line denotes the rate at which an expert policy reaches the correct goal, while the dotted blue line denotes the same rate for our model with full-diffusion ($\fwr=1.0$).}\label{fig:bp-assisted}
\end{figure}

\subsection{Training}
To understand the pilot's intent, our copilot relies on a state-conditioned diffusion model $p_\theta(a_t \vert s_t)$.
This diffusion model is trained as follows. First, we collect expert demonstrations, each containing a sequence of goal-embedded state-action pairs for each domain, where the goal is randomly sampled at each episode.
We then use the resulting demonstrations to train a state-conditioned diffusion model using the DDPM loss (Eqn.~\ref{eqn:ddpm-loss}).
The details of the model architecture are described in Appendix~\ref{app:model-arch-and-hparams}.
As we hide a goal from our copilot, we remove goal locations from each observation prior to training.

The following experiments use a single diffusion model trained separately for each task. We note that the copilot's behavior changes only according to $\fwr$ given a diffusion model.

To collect demonstrations, we first train an expert policy with soft actor-critic~\cite{sac}
 for $3\text{M}$ timesteps in Lunar Lander and Lunar Reacher, and
$1\text{M}$ timesteps in Block Pushing.
We then roll out the policies in each environment to collect demonstrations of state-action pairs $D_\textrm{expert}$ for various goals.
Appendix~\ref{app:model-arch-and-hparams} details the model architecture and the hyperparameter settings.

\subsection{Surrogate pilots for evaluation}\label{subsec:pilot-and-copilot}
Our method does not require access to a pilot (surrogate or otherwise) when training the copilot. However, surrogate pilots are useful when we want to perform a large number of evaluations
in a reproducible manner.
Thus, we prepare two surrogate pilots consistent with previous work~\cite{sha-via-deeprl,rsa}: a \emph{Laggy} pilot and a \emph{Noisy} pilot, both of which are corrupted versions of a single expert.
At each time step, the Laggy pilot repeats its previous action with probability $p_\textrm{laggy}$, and otherwise executes an action drawn from the expert's policy. With probability $p_\textrm{noisy}$, the Noisy pilot samples an action from a uniform distribution over the action space, and otherwise executes an action sampled from the expert policy. We evaluate our shared autonomy algorithm for pilots across a broad range of parameters $p_\textrm{noisy}$ and $p_\textrm{laggy}$. Due to space constraints, we only include results for a representative subset of the pilot parameters for Lunar Lander ($p_\text{laggy} = 0.85,~p_\text{noisy} = 0.3$), Lunar Reacher ($p_\text{laggy} = 0.85,~p_\text{noisy} = 0.6$), and Block Pushing ($p_\text{laggy} = p_\text{noisy} = 0.6$), but include results for the full range of pilot parameters in Figures~\ref{fig:noisy_pilot_ll_landing_cmp}--\ref{fig:laggy_pilot_bp_cmp} of Appendix~\ref{app:effect-of-alpha-across-surrogates}.

\subsection{Various pilots with our copilot}
We evaluate our copilot with various surrogate pilots in all four environments.
Table \ref{tb:lunar-lander-reacher} shows success and crash/out-of-bound rates when various pilots are paired with our copilot in Lunar Lander and Lunar Reacher.
We see that adopting our copilot significantly improves the success and crash rates for all but the expert policy (as expected).
We note that each task involves reaching or landing on a randomly sampled target that is not known to the copilot. Consequently, one can not expect the copilot to have significant effect on the success rate of the Random or Zero pilot. Nevertheless, the results show that our copilot improves their success rates with performance similar to that of full diffusion ($\fwr = 1.0$) and significantly decreases the rate at which the pilots crash.

\begin{table}[!t]
    \centering
    \caption{Correct goal and timeout rates on the Block Pushing task for different pilots with ($\gamma = 0.2$) and without assistance. Each entry corresponds to $10$ episodes across $30$ random seeds. Note that the Zero and Random pilots have no knowledge of the goal.}
    \label{tb:Block-Push}
    \setlength{\tabcolsep}{3pt}
    \begin{tabularx}{1.0\linewidth}{lYYYY}%
        \toprule
        & \multicolumn{2}{c}{Correct Goal Rate $\uparrow$} & \multicolumn{2}{c}{Timeout Rate $\downarrow$}\\
        Pilot & w/o Copilot & w/ Copilot & w/o Copilot & w/ Copilot\\
        \midrule
        Noisy &
        $ 62.33 \pm 2.49 $ &  $ \bm{75.67 \pm 2.05} $ &
        $ \hphantom{0}37.33 \pm 2.05 $ &  $ \bm{20.00 \pm 3.27} $\\
        Laggy &
        $ 42.00 \pm 4.90 $ &  $ \bm{74.33 \pm 2.87} $ &
        $ \hphantom{0}58.00 \pm 4.90 $ &  $ \bm{21.33 \pm 4.50} $ \\
        \midrule[0.1pt]
        Zero &
        $ \hphantom{0}0.00 \pm 0.00 $ &  $ 40.67 \pm 2.62 $ &
        $ 100.00 \pm 0.00 $ &  $ \hphantom{0}6.67 \pm 1.70 $\\
        Random &
        $ \hphantom{0}0.00 \pm 0.00 $ &  $ 16.33 \pm 3.09 $ & 
        $ 100.00 \pm 0.00 $ &  $ 68.00 \pm 1.63 $\\
        Expert &
        $ 99.00 \pm 0.82 $ &  $ 94.67 \pm 2.05 $ &
        $ \hphantom{00}1.00 \pm 0.82 $ &  $ \hphantom{0}5.33 \pm 2.05 $\\                
        \bottomrule
    \end{tabularx}
\end{table}
For the Block Pushing task, the Laggy and Noisy policy are based on an expert whose intent is always to push the block to the green zone, which we refer to as the correct goal. The set of state-action pairs on which we trained the diffusion model also include trajectories that reach the red zone. As such, without knowledge of the pilot's goal, the copilot may generate actions that push the block into the red zone (referred to as the incorrect goal), causing the episode to terminate. As we see in Table~\ref{tb:Block-Push}, our copilot improves the rate at which all but an expert pilot reach the correct goal, while decreasing the rate at which episodes timeout without reaching any goal. %

\subsection{The effects of the forward diffusion ratio $\fwr$}
The forward diffusion ratio $\fwr$ determines how many steps of forward diffusion process to apply on pilot's action $a^h_t$. 
This value changes the balance of interpolating between source and target distributions, which in turn controls the trade-off between fidelity and conformity of an action.
In this section, we demonstrate this trade-off with different $\fwr$ values in various environments.
We deployed our copilot with various surrogate pilots as discussed in Section \ref{subsec:pilot-and-copilot} for each $\fwr$ value.

Figure~\ref{fig:maze-vs-fwr} shows the effect of $\fwr$ for the  2D Control domain. Although the copilot is trained on demonstrations for both goals, the pilot's goal is fixed to the left. As $\fwr$ increases, we see that the number of trajectories that reach the correct goal increases. Qualitatively, we also see that the stochasticity of the trajectories lessens, and the paths towards the goal become smoother. However, when $\fwr$ is too high, the copilot ignores the pilot's actions. Without knowledge of the pilot's goal, the copilot ends up distributing trajectories evenly between the left and right goals, consistent with the set of trajectories on which the model was trained.

Figures~\ref{fig:ll-land-assisted} summarizes the results on Lunar Lander and Lunar Reacher, respectively, by categorizing the trajectories into success, crash and float. Here, float denotes a trajectory where the spaceship 
does not crash nor go out of bounds, but nevertheless fails to complete the task within the time limit.

In all cases, we observe that the success rate follows a distinct pattern. First, the success rate monotonically increases with $\fwr$ as the copilot improves task performance. After reaching a peak success rate at a critical value of $\fwr$, the success rate gradually decreases.
The drop in the success rate reflects the copilots infidelity to the user's intent. The pilot's original actions provide a signal of which goal or landing pad is correct. When $\fwr$ is too large, the copilot totally ignores the pilot's actions and instead chooses to land in a manner following the copilot's original training distribution.

In general, the crash rate monotonically decreases with $\fwr$, with the exception of Lunar Reacher. This demonstrates that the quality of the generated actions improves as $\fwr$ increases.
In Lunar Reacher, many of the demonstrations go straight to a target, which is often placed at the edge of the the environment. Consequently, if the agent overshoots and misses the target, it is very likely that it will go out of bounds, which is counted as a crash. This is likely why increasing $\fwr$ does not necessarily result in lower crash rates for Lunar Reacher.

\subsection{Real Human User Experiments}
\label{subsec:real-user-exp}
We conducted a set of human user experiments involving the Lunar Lander and Lunar Reacher continuous control domains. For the experiments, we recruited $17$ participants ($9$ who identified as male, and $8$ as female, with an average age of $25$).\footnote{None of the participants were co-authors or otherwise involved in this research.} We asked each user to interact with one of two different co-pilots within each episode, one corresponding to direct teleoperation (i.e., no assistance) and the other being our diffusion-based shared autonomy assistant. The identity of the co-pilot was not disclosed to the participant. At the beginning of each experiment, we allowed the user to practice for $10$ episodes with each co-pilot. In the subsequent testing phase, the user controlled the system for another $10$ episodes with each of the co-pilots. We conducted this experiment for both Lunar Lander and Lunar Reacher and emphasize that the location of the goal (the landing pad for Lunar Lander and goal region for Lunar Reacher) was chosen randomly from the continuous space of goals for each episode.

\begin{table}[!th]
    \centering
    \caption{Success, crash/out-of-bounds (OOB), and float rates on Lunar Lander and Lunar Reacher for human user experiments.}
    \label{tb:human-lunar-lander-reacher}
    \begin{tabularx}{1.0\linewidth}{lYYYYYY}%
        \toprule
        & \multicolumn{3}{c}{\bf Lunar Lander} & \multicolumn{3}{c}{\bf Lunar Reacher}\\
        \midrule
        & Success Rate $\uparrow$ & Crash Rate $\downarrow$ & Float Rate & Success Rate $\uparrow$ & Crash Rate $\downarrow$ & Float Rate\\
        \midrule
        w/o Copilot & $\hphantom{0}1.76$ & $98.24$ & $0.00$ & $17.06$ & $82.94$ & $0.00$\\
        w/ Copilot & $32.35$ & $61.18$ & $6.47$ & $34.71$ & $65.29$ & $0.00$\\ 
        \bottomrule
    \end{tabularx}
\end{table}
We consider both the quantitative and qualitative performance of our shared autonomy algorithm. Quantitatively, Table~\ref{tb:human-lunar-lander-reacher} compares the average performance of human pilots on Lunar Lander and Lunar Reacher in terms of success, crash/out-of-bounds, and float (i.e., neither crashing, going out-of-bounds, nor succeeding) with and without the assistance of our diffusion-based shared autonomy. We performed a Welch $t$-test for the success rate of each participant and find $p$-values of $p = 3.363 \times 10^{-7}$ for Lunar Lander and $p = 9.033 \times 10^{-3}$ for Lunar Reacher, based on which we can reject the null hypothesis that our diffusion-based co-pilot does not improve success rate.

Qualitatively, we asked participants to rate their agreement with five statements about whether they felt that each co-pilot behaved in a manner that was ``helpful'', ``consistent'', ``responsive'', ``collaborative'', and ``trustworthy'' using a five-point Likert scale. Figure~\ref{fig:human-lunar-lander-reacher-survey} in Appendix~\ref{app:human-user-experiments} visualizes the survey results for all four combinations of tasks and co-pilots. The results reveal that human users rated our shared autonomy co-pilot higher than the no-assistance co-pilot (i.e., teleoperation) with regards to all five qualities for both environments.

\subsection{Real-Robot Experiments}

\begin{figure}[!t]
    \centering
    \subfloat{\includegraphics[width=0.48\linewidth]{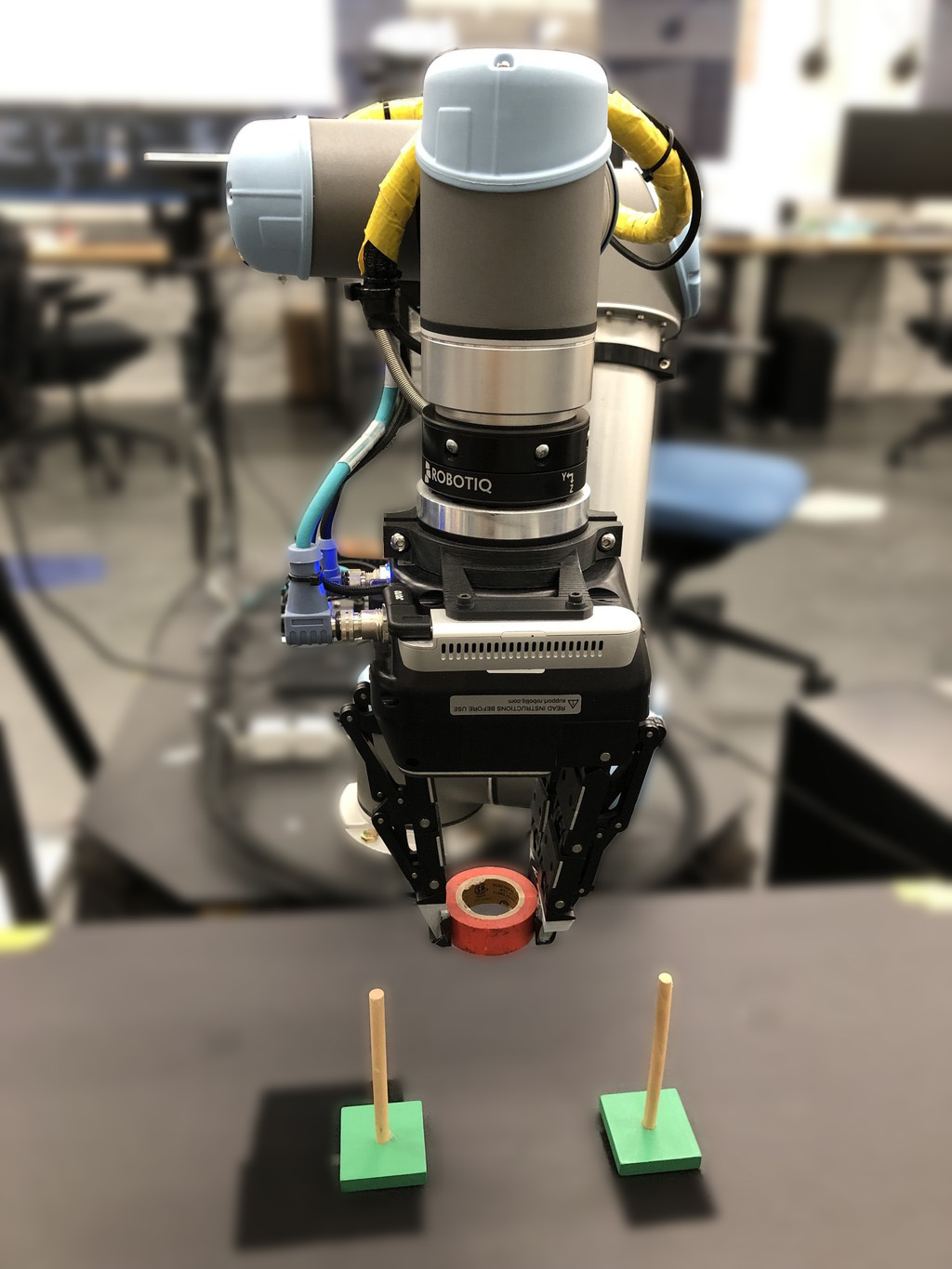}}\hfil
    \subfloat{\includegraphics[width=0.48\linewidth]{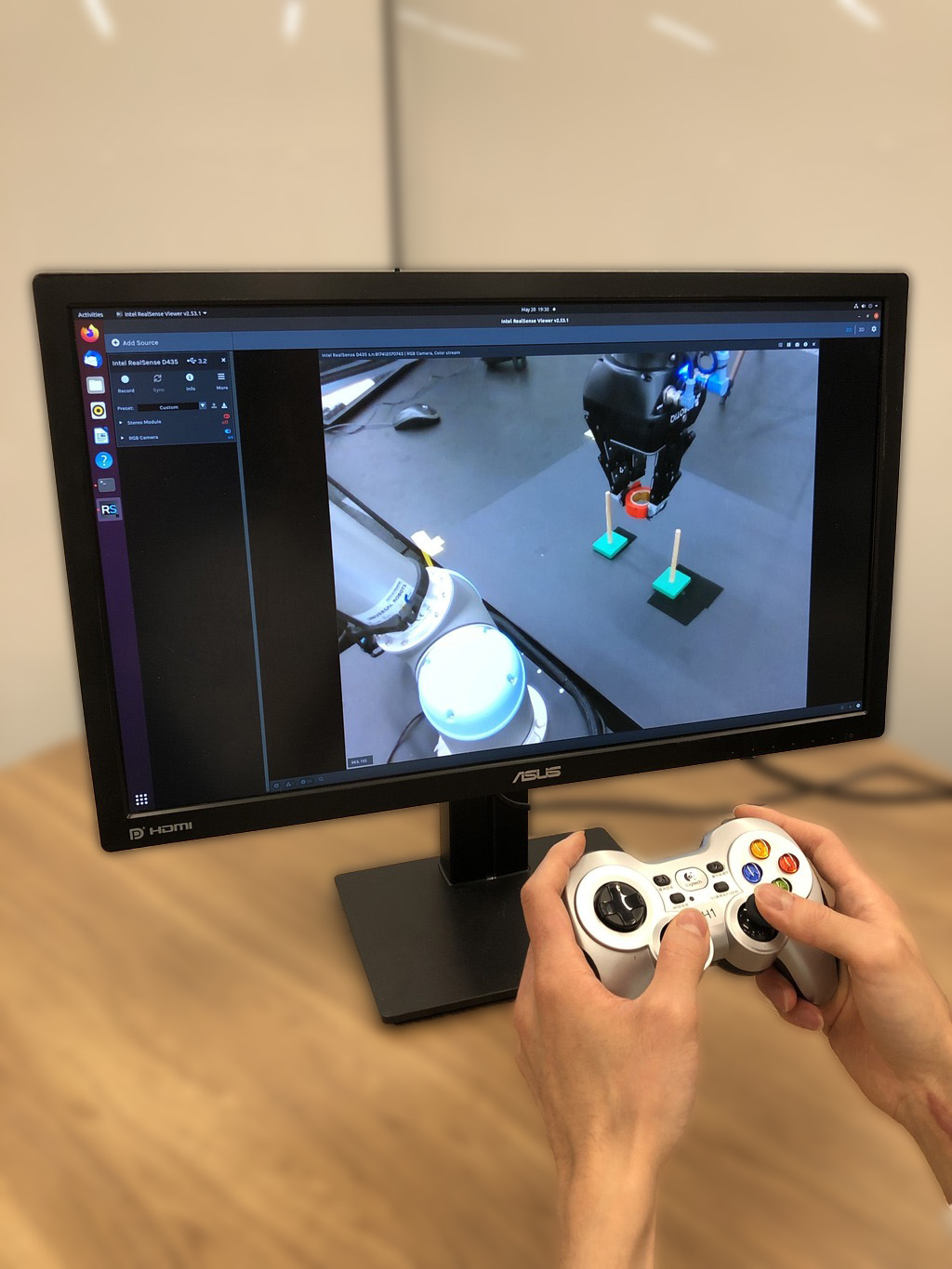}}
    \caption{The real-robot experiments involved (left) placing a disk on one of two posts. Human operators (right) controlled the 3D motion of the end-effector while observing images of the robot's workspace streamed in real-time.}\label{fig:real-robot-experiments}
\end{figure}

We consider an real-world task (Fig.~\ref{fig:real-robot-experiments}) in which a user commands a UR5 manipulator equipped with a pinch gripper to place a donut-shaped object (e.g., a disk with a hole in the center) held by the robot onto one of two posts in the robot's workspace (e.g., analogous to Towers of Hanoi, but with two poles and one disk). %
Emulating remote operation, the user observes a real-time (i.e., little-to-no latency) video feed of the robot and its workspace from a third-person view. The user controls the three-dimensional velocity of the end-effector in Cartesian space using a game controller with two joysticks (velocity in the $x-$ and $y-$directions using the right joystick, and the $z$-direction using the left joystick). The arm position is initialized to a random location.

As in the human user experiments discussed above, we presented the user with two unidentified co-pilots, one being our diffusion-based shared autonomy assistant and the other being direct teleoperation (i.e., no assistance). We allowed each user to practice with each co-pilot for three episodes. In the subsequent testing phase, participants controlled the arm using both interfaces for five episodes each. We made sure to use the same set of (random) initial arm positions for both test phases. We set the time limit to be $500$ steps (where each velocity command is considered a single step), and treat any episode that exceeds this limit as a failure.
After the testing phase, we asked each participant to rate their agreement with the same quantitative questions in the same manner as the Lunar Lander and Lunar Reacher experiments using a five-point Likert
scale.

We recruited $14$ participants ($10$ who
identified as male, $4$ who
identified as female, average age of $27.7$)\footnote{None of the participants were co-authors or otherwise involved in this research.} for human user experiments with the real robot.

\begin{table}[!t]
    \centering
    \caption{Success rate and episode length with and without copilot for the real-robot experiment with human users.}
    \label{tb:human-ur5}
    \begin{tabularx}{0.95\linewidth}{lYY}
        \toprule
        & success rate $\uparrow$ & episode length $\downarrow$\\
        \midrule
        w/o Copilot & $0.89$ & $209.23 \pm 83.71$\\
        w/ Copilot & $1.00$ & $143.66 \pm 68.42$\\
        \bottomrule
    \end{tabularx}
\end{table}

Table~\ref{tb:human-ur5} shows the success rate and episode length %
that users took to reach the goal. We found that providing users with the assistance of our diffusion-based shared autonomy algorithm both improved their success rate and decreased the time required to complete the task. The results of the qualitative evaluation reveal that users preferred commanding the manipulator with the assistance of our algorithm as detailed in Appendix~\ref{app:human-user-experiments}.

\section{Related Work} \label{sec:related-work}
Shared autonomy has appeared in many problem domains, including remote telepresense~\cite{telepresence01, goertz63, rosenberg93}, assistive robotic manipulation \cite{assistive-manip-01, muelling17, assistive-manip-03}, and assistive navigation~\cite{assistive-navi-01, assistive-navi-02}.
In shared autonomy, one of the most persistent challenges has been correctly identifying the pilot's intentions or goals. 

Early work sidesteps this challenge by assuming a priori knowledge of the pilot's goals ~\cite{users-goal-is-known-01, users-goal-is-known-02}.
Recent work has managed to relax this assumption by treating the pilot's goal as a latent random variable which can be inferred from environmental observations and pilot actions~\cite{infer-user-goal-01, infer-user-goal-02, dragan13, infer-user-goal-04, javdani15, users-goal-is-known-02, muelling14, perez15, infer-user-goal-09}.
In spite of this forward progress, these methods still assume knowledge about some combination of the transition dynamics, the pilot's goals, or pilot's policy, making them difficult to deploy in many unstructured scenarios.

\citet{sha-via-deeprl}, and subsequently \cite{optimizing-interventions-sha, rsa} introduce model-free deep reinforcement learning (RL) to the shared autonomy setting. Because these methods are model free, knowledge of environment dynamics is no longer required, allowing one to train a policy that is not limited to a specific model class. Several follow up works have adapted deep RL to a variety of shared autonomy problems \cite{ave, Reddy2022FirstCU, disagreement-sub-policies, asha}.

To the best of our knowledge, all previous work on shared autonomy with deep RL either explicitly or implicitly assumes a human-in-the-loop setting. In particular, the main training loop typically contains a step to query a pilot (i.e., human user) to obtain its action. This constraint often makes training inefficient or impractical. \citet{asha} addresses this issue by proposing a two-phase training scheme.
In the first phrase, an autonomous agent learns a task-conditioned policy that can be helpful for assisting a pilot. This is followed by the second phase, which incorporates sparse feedback from humans. Although such approach improves the efficiency of the training pipeline, it does not change the fact that these methods inherently rely on human feedback.

Diffusion models \cite{thermo} have recently been applied to many problems including image generation, image editing, text-conditioned image generation and video generation. In the field of robotics, \citet{janner2022diffuser} trained a diffusion model over trajectories, demonstrating that diffusion is capable of generating a diverse set of trajectories reaching a goal location. \citet{anonymous2023imitating} applies diffusion models to imitation learning. Meanwhile, \citet{diffusion-policy} show that the policy parameterized with diffusion can generate a multi-modal distributions over possible actions. As far as we are aware, ours is the first approach that uses diffusion models in a shared control scenario, where a copilot learns from expert demonstrations and provides guidance by denoising noisy pilot action.

The core idea of our approach is in the partial forward and reverse diffusion that enables us to generate an output that effectively interpolates between the original input and one from target distribution that we train our diffusion models on. Although we independently came up with this idea, a others have explored a similar approach for image editing~\cite{sdedit}. Related, many recent papers have consider running reverse diffusion from some intermediate step rather than pure Gaussian noise~\cite{Lyu2022AcceleratingDM, sjc, dreamfusion, ddnm}, as we do in this paper.

\section{Conclusion} \label{sec:conclusion}
In this paper, we presented a new approach to shared autonomy based on diffusion models.
Our approach only requires access to demonstrations that are representative of desired behavior, and does not assume access to or knowledge of the user's policy, reward feedback, or knowledge of the goal space or environment dynamics. Integral to our approach is its modulation of the forward and reverse diffusion processes in a manner that seeks to balance the user's desire to maintain control authority with the benefits of generating actions that are consistent with the distribution over desired behaviors.
We evaluated our copilot on various continuous control environments and demonstrated that our diffusion-based copilot generalizes across a variety of pilots, improving their performance, while preserving their intention. We further presented an analysis of the effects of different degrees of partial diffusion on task performance.
One limitation of our approach is that there is no component that explicitly addresses the likely mismatch in state distributions between $P_\text{pilot}(s)$ and $P_\text{target}(s)$. Intuitively, it is very likely that the target state visitation distribution (i.e., expert demonstrations) is different from that of the pilot. However, our empirical results suggest that this is not a critical limitation, possibly because executing corrected actions tends to encourage the agent to visit states that are close to those visited as part of the expert demonstrations. One means of addressing this is to design a goal-conditioned policy that can navigate itself to an in-distribution state at test time. We leave this for a future work.

\section*{Acknowledgments}

Ge Yang is supported by the National Science Foundation Institute for Artificial Intelligence and Fundamental Interactions (IAIFI, \url{https://iaifi.org/}) under the Cooperative Agreement PHY-2019786. This work was supported in part by ARO grant W911NF‐17‐1‐0188 and NSF HDR TRIPODS grant 2216899. We would like to thank the members of the TTIC robotics group for insightful discussions. We also appreciate the conversations we had with Haochen Wang. His in-depth knowledge of diffusion models was invaluable to us.

\bibliographystyle{plainnat}
\bibliography{references}

\flushcolsend

\clearpage
\appendices

\section{Model architecture and hyperparameters}
\label{app:model-arch-and-hparams}
We implement the state-conditioned denoising network $\bm{\epsilon}_\theta(\bm{a}_t, k \mid \bm{s}_t)$ that makes up the diffusion model as a 4-layer multi-layer perceptron (MLP) with latent dimension $h_\textrm{dim} = 128$ and softplus~\cite{softplus-01} activation after every layer except for the last one.
The network outputs a vector that has same dimension as in the input.
Each layer is conditioned on a diffusion timestep $k$ that looks up a corresponding embedding with dimension $h_\textrm{dim}=128$, and is subsequently fused with the output of the linear layer by an element-wise product.
To accommodate its conditioning on $\bm{s}_t$, our denoising network works on a concatenation of state-action pairs $(\bm{s}_t, \bm{a}_t)$, 
and during training, we add Gaussian noise only to the action component to produce a target output. The input $\bm{x}$ and the target $\bm{y}$ are then follow as
\begin{subequations}
    \begin{align}
        \bm{x} &= (\bm{s}_t, \bm{a}_t + \epsilon), ~\epsilon \sim \mathcal{N}(\bm{0}, \bm{I})\\
        \bm{y} &= (\bm{0}, \bm{\epsilon}).
    \end{align} 
\end{subequations}
We generate the set of demonstrations used to train our diffusion model by collecting $100$k transitions for each goal in the 2D Control environment using the scripted expert policy provided in the original implementation~\cite{fu2020d4rl}, $3$M transitions for each goal in the Block Pushing environment using the pretrained expert, and $10$M transitions for Lunar Lander and Lunar Reacher using the corresponding pretrained expert. When we collect expert demonstrations, we filter out episodes that do not succeed, effectively making the quality of the demonstrations better than the original expert. This explains why the assisted expert often performs better than the expert (e.g., Table~\ref{tb:lunar-lander-reacher}---Lunar Reacher).

We use a publicly available implementation of  DDPM~\citep{diffusion-models-notebook}.
For the sake of faster action generation, we set the number of diffusion steps to $K = 50$, and use settings of $\beta_\textrm{min} = 10^{-4}$ and $\beta_\textrm{max} = 0.26 $ with sigmoid scheduling.

An implementation of our method as source code together with video demonstrations are available at \url{https://diffusion-for-shared-autonomy.github.io/}.

\section{Effect of $\fwr$ for Different Surrogate Pilots} \label{app:effect-of-alpha-across-surrogates}

\begin{figure}[!t]
    \centering
    \includegraphics[width=0.7\linewidth]{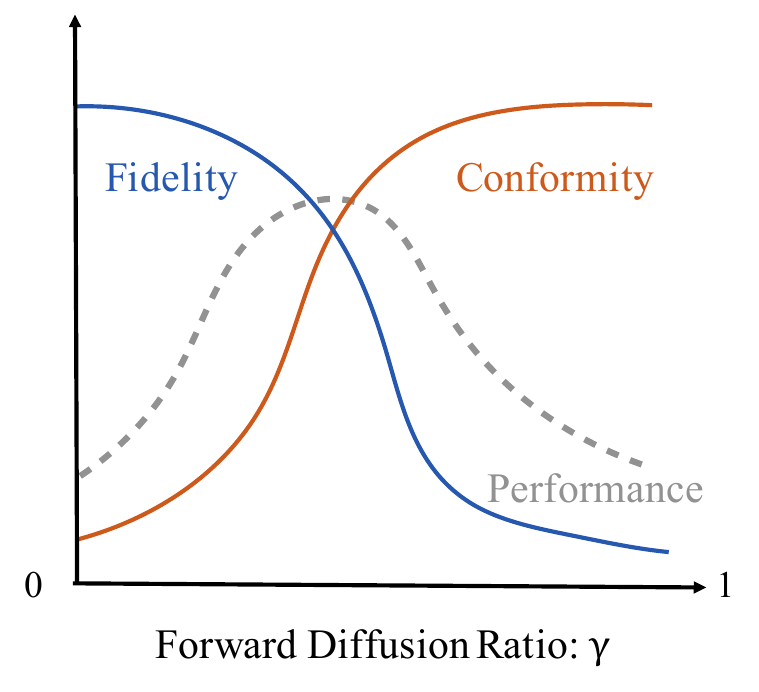}
    \caption{Our algorithm uses the \emph{forward diffusion ratio} $\fwr$ to regulate the extent of forward and reverse diffusion. When $\fwr$ is small, the assistant generates actions nearly identical to those of the user (i.e., high fidelity) but may
    be far from the desired behaviors
    (i.e., low conformity). Depending on the competency of the user, these actions may result in poor task performance (e.g., due to crashes). As with standard diffusion, when $\fwr$ is large, the assistant will generate actions that have high likelihood under the distribution over desired behavior (i.e., high conformity), with little-to-no regard for their similarity to the user's actions (i.e., low fidelity). While these actions will be safe, they will likely result in poor performance since the forward process has removed any information about user's intention. By controlling $\fwr$, our algorithm identifies a regime that trades off between fidelity and conformity to generate actions that preserve the user's intent and, in turn, the critical information to complete the task.}%
    \label{fig:trade-off}
\end{figure}
Our framework modulates the amount of forward and reverse diffusion as a means of trading off a user's desire to maintain control authority with the general benefits that come with emulating behavior that may be seen as more desirable. Figure~\ref{fig:trade-off} provides a visualization of how the forward diffusion ratio may influence the ``fidelity'' of the diffused state-conditioned action with respect to the user's input and the extent to which it ``conforms'' to the desirable behavior.

As discussed in Section~\ref{subsec:pilot-and-copilot}, we investigate the effectiveness of our shared autonomy algorithm when used in conjunction with  pilots that seek to emulate the deficiencies typical of human control. Following previous work~\cite{sha-via-deeprl,rsa}, we employ two surrogate  policies for evaluation, namely a \emph{Laggy} pilot and a \emph{Noisy} pilot, both of which are corrupted versions of a single expert.

At each time step, with probability $p_\textrm{noisy}$, the Noisy pilot samples an action from a uniform distribution over the action space, and otherwise executes an action sampled from the expert policy. Similarly, the Laggy pilot repeats its previous action at each timestep with probability $p_\textrm{laggy}$, and otherwise carries out an action drawn from the expert policy. We evaluate our shared autonomy algorithm for pilots with a broad range of different parameters $p_\text{noisy} \in \{0.2, 0.3, \ldots, 0.8, 0.9\}$ and $p_\text{laggy} \in \{0.2, 0.3, \ldots, 0.8, 0.9\}$ for Lunar Lander, Lunar Reacher, and Block Pushing.

\subsection{Lunar Lander and Lunar Reacher}
\begin{figure*}
    \centering
    \subfloat[$p_\text{noise}=0.2$]{\includegraphics[width=0.2\linewidth]{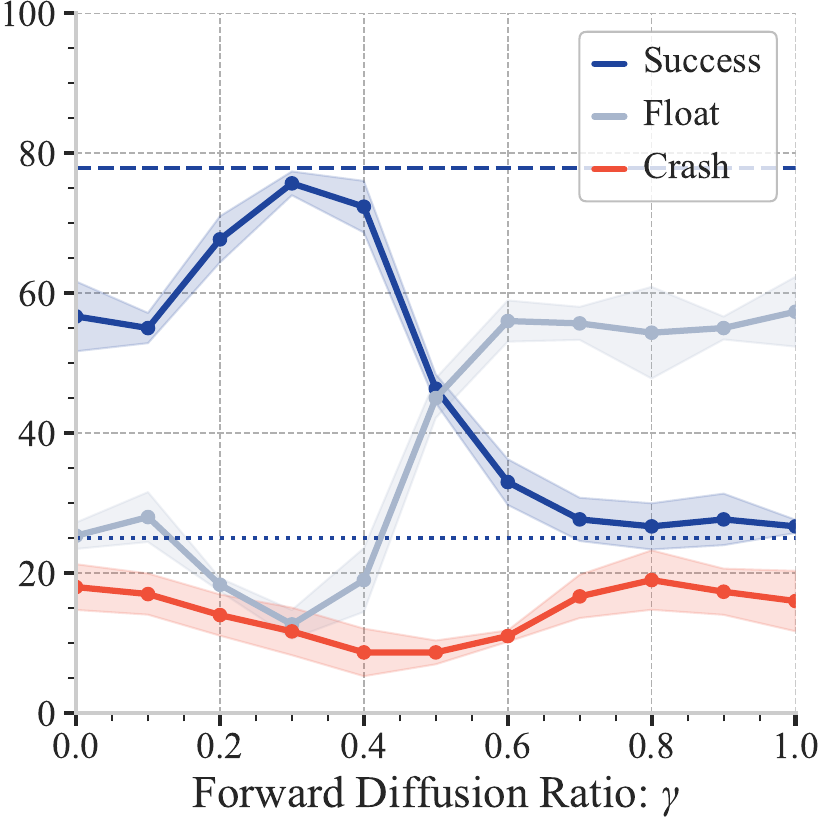}\label{fig:ll_noisy2}}
    \hfil
    \subfloat[$p_\text{noise}=0.3$]{\includegraphics[width=0.2\linewidth]{figures/noisy_laggy_prob_cmp/Landing/landing_rates_noisy_3.pdf}\label{fig:ll_noisy3}}
    \hfil
    \subfloat[$p_\text{noise}=0.4$]{\includegraphics[width=0.2\linewidth]{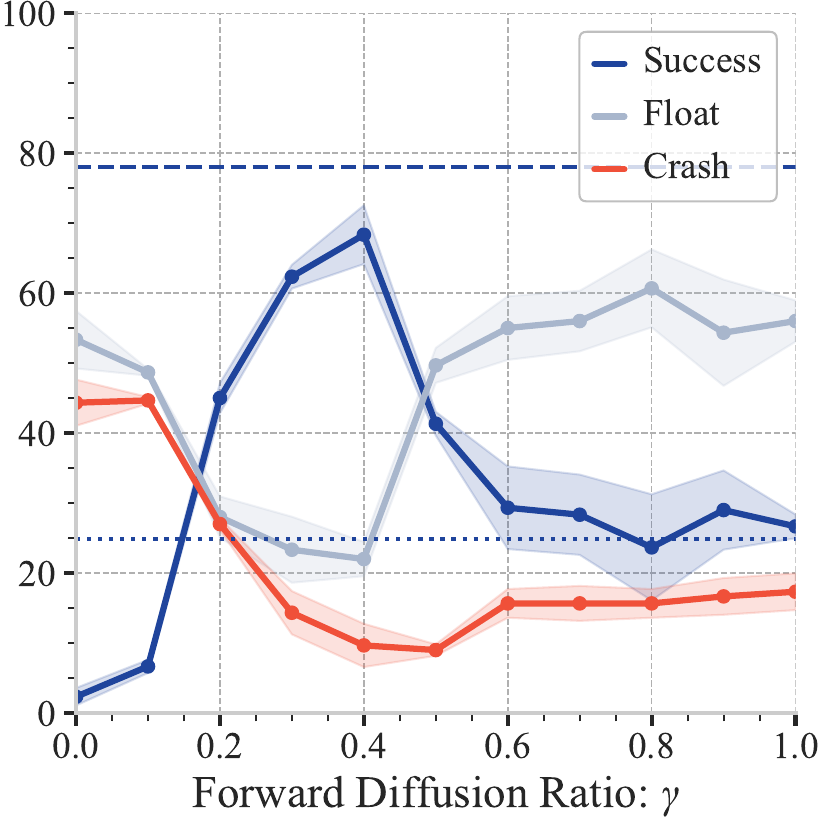}\label{fig:ll_noisy4}}
    \hfil
    \subfloat[$p_\text{noise}=0.5$]{\includegraphics[width=0.2\linewidth]{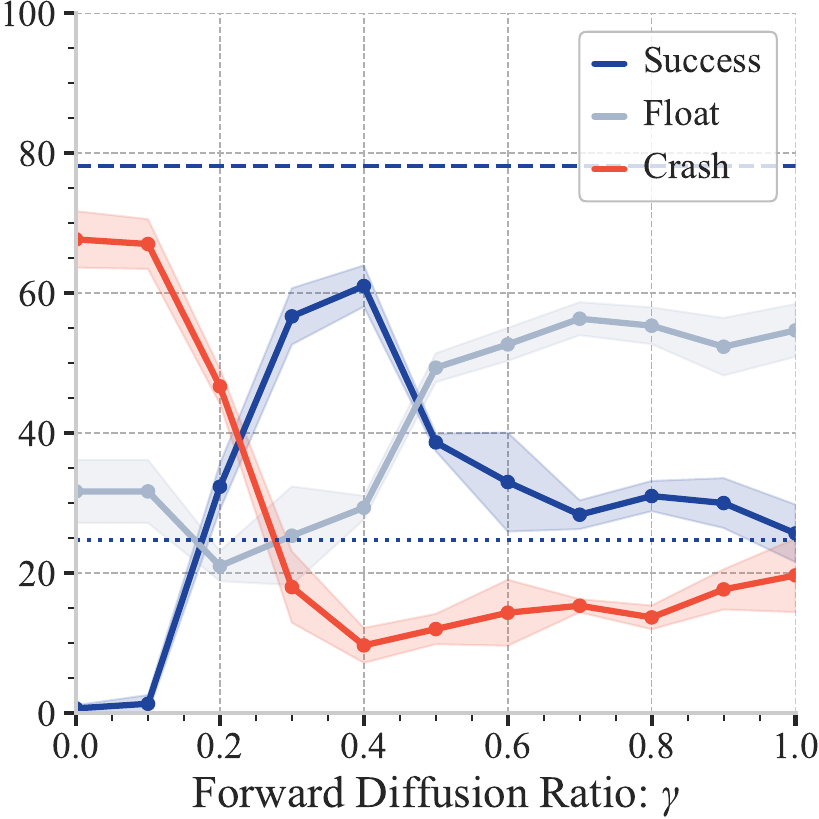}\label{fig:ll_noisy5}}\\
    
    \subfloat[$p_\text{noise}=0.6$]{\includegraphics[width=0.2\linewidth]{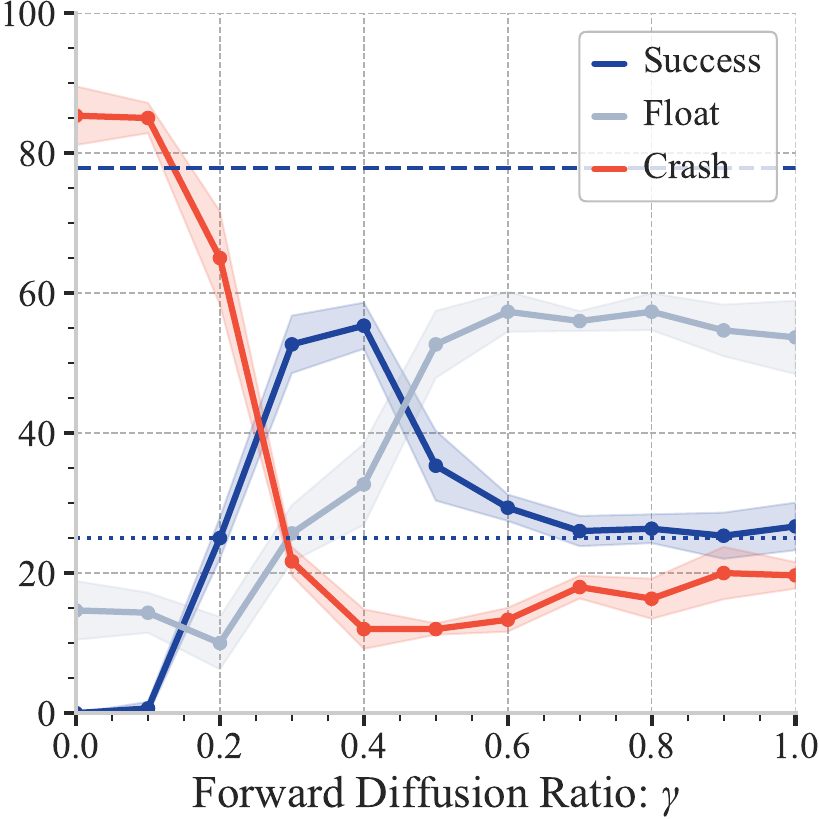}\label{fig:ll_noisy6}}
    \hfil
    \subfloat[$p_\text{noise}=0.7$]{\includegraphics[width=0.2\linewidth]{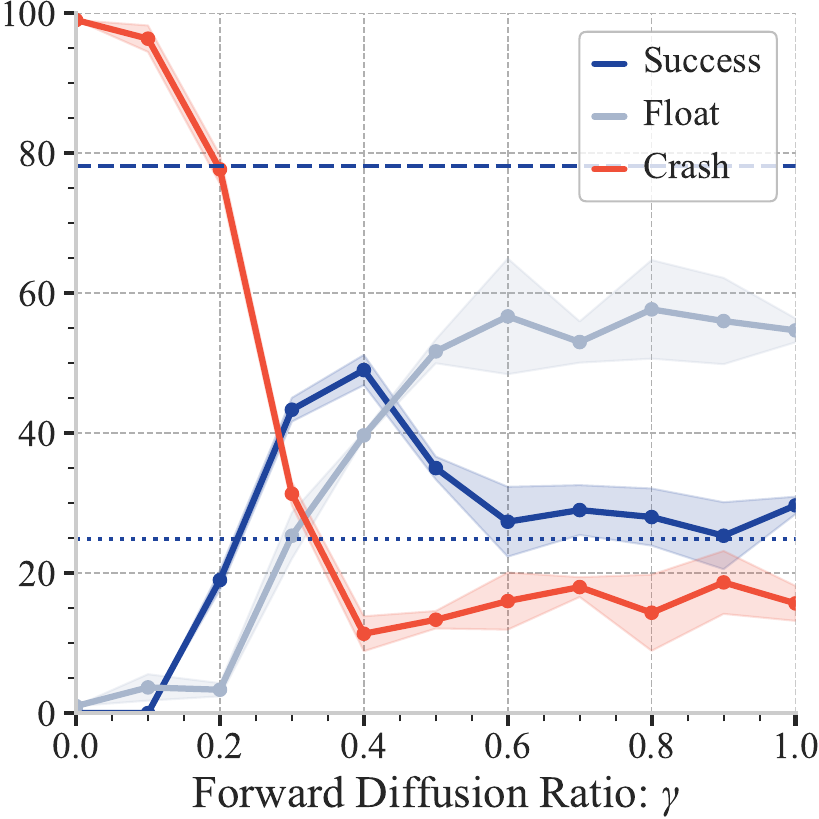}\label{fig:ll_noisy7}}
    \hfil
    \subfloat[$p_\text{noise}=0.8$]{\includegraphics[width=0.2\linewidth]{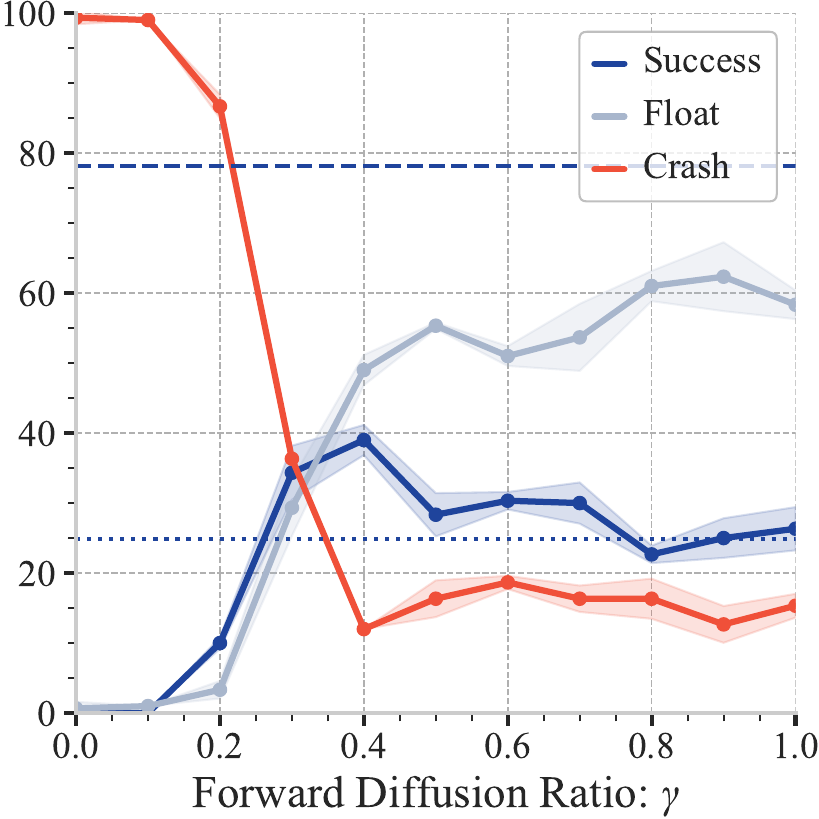}\label{fig:ll_noisy8}}
    \hfil
    \subfloat[$p_\text{noise}=0.9$]{\includegraphics[width=0.2\linewidth]{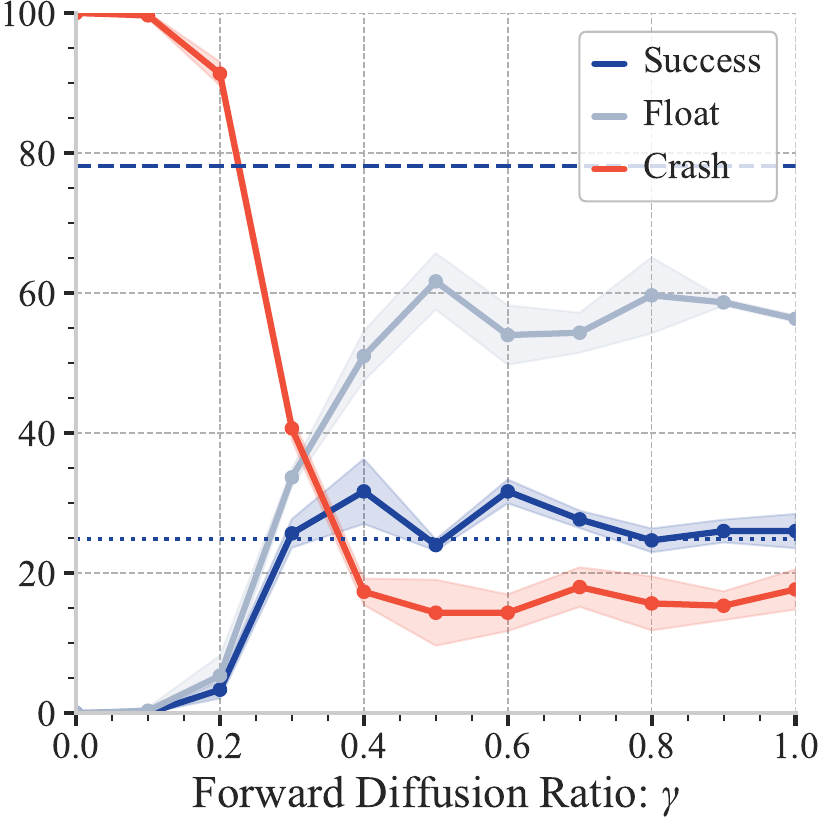}\label{fig:ll_noisy9}}
    
    \caption{Lunar Lander Performance as a function of the forward diffusion ratio $\fwr$ for a noisy pilot with noise values ranging from \subref{fig:ll_noisy2} $p_\text{noisy} = 0.2$ to  \subref{fig:ll_noisy9} $p_\text{noisy} = 0.9$. In all plots, the dashed blue line denotes the success rate of the expert pilot, while the dotted blue line is the success rate of our model when performing ``full'' diffusion (i.e., $\gamma = 1.0$) on an action sampled from a zero-mean isotropic Gaussian distribution, which we refer to as a ``Random'' pilot in the paper.}\label{fig:noisy_pilot_ll_landing_cmp}
\end{figure*}
Figure~\ref{fig:noisy_pilot_ll_landing_cmp} plots the effect of different settings for the forward diffusion ratio $\fwr$ in terms of success, crash, and float rates on Lunar Lander for different parameterizations of the Noisy pilot. At the smallest noise setting $p_\text{noisy} = 0.2$, the Noisy pilot is not significantly worse than the expert pilot without assistance ($\fwr = 0.0$). The Noisy pilot successfully lands at the goal approximately $66\%$ of the time compared to approximately $78\%$ for the expert pilot, and crashes (or goes out of bounds) approximately $19\%$ of the time compared to approximately $12\%$ for the expert pilot. As the noise level increases, the performance of the noisy pilot significantly deteriorates without assistance---the pilot rarely lands and instead either crashes (or goes out-of-bounds) or floats around until the episode times out. However, for each parameterization of the Noisy pilot, we see that the assistance of the copilot gives rise to a noticeable increase in task success rate and decrease in crash rate as we increase the forward diffusion ration to $\fwr = 0.4$. As the forward diffusion ratio increases further, the success rate decreases as expected given that the copilot does not have knowledge of the goal (i.e., the landing location). In this case, the copilot chooses actions that are consistent with the different goals that one or more previous experts reached. At the same time, we see only a slight or no increase in the crash rate as the forward diffusion ratio increases, particularly for pilots with higher noise settings.

\begin{figure*}
    \centering
    \subfloat[$p_\text{laggy}=0.2$]{\includegraphics[width=0.2\linewidth]{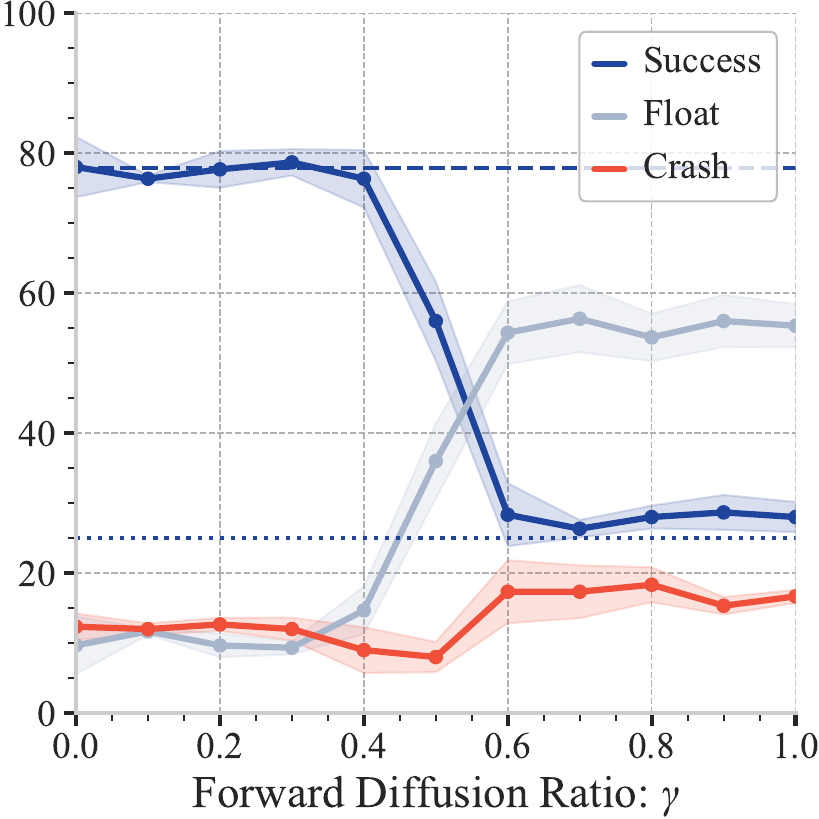}\label{fig:ll_laggy2}}
    \hfil
    \subfloat[$p_\text{laggy}=0.3$]{\includegraphics[width=0.2\linewidth]{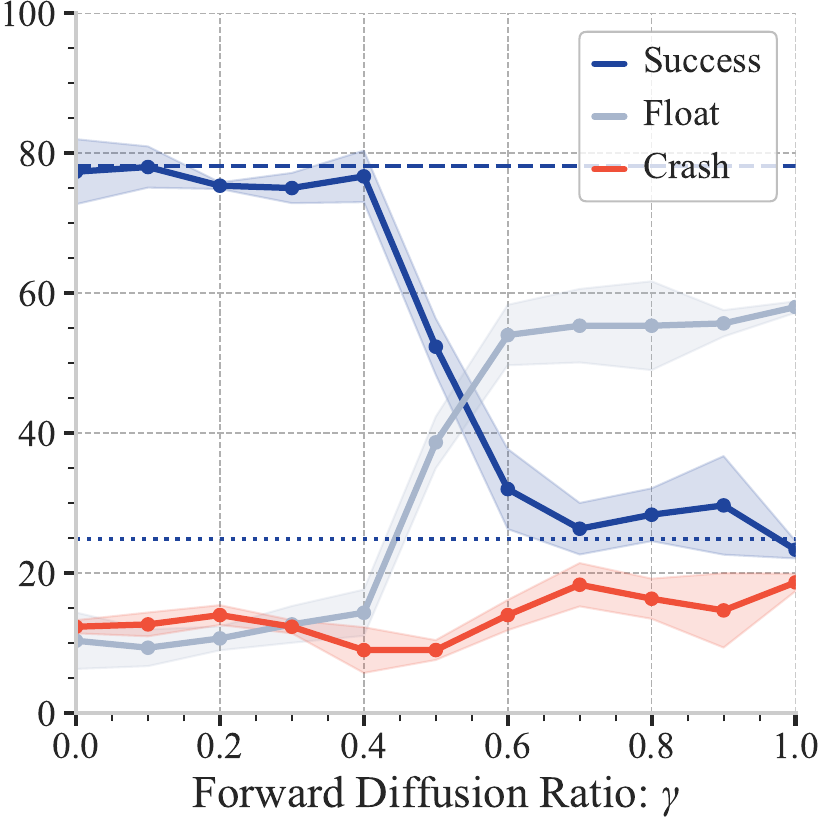}\label{fig:ll_laggy3}}
    \hfil
   \subfloat[$p_\text{laggy}=0.4$]{\includegraphics[width=0.2\linewidth]{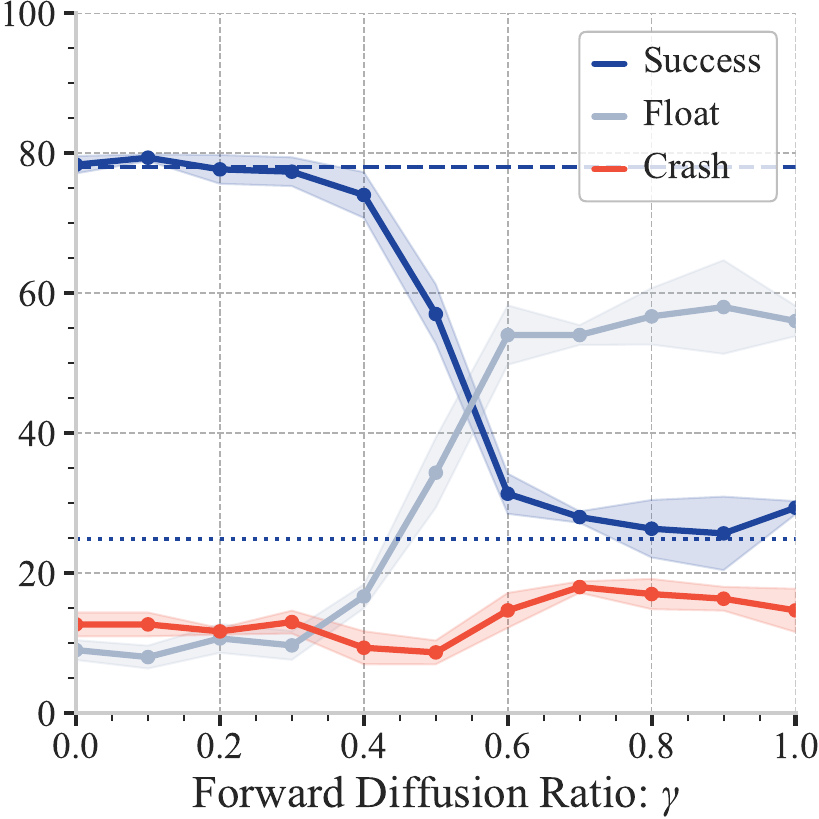}\label{fig:ll_laggy4}}
    \hfil
    \subfloat[$p_\text{laggy}=0.5$]{\includegraphics[width=0.2\linewidth]{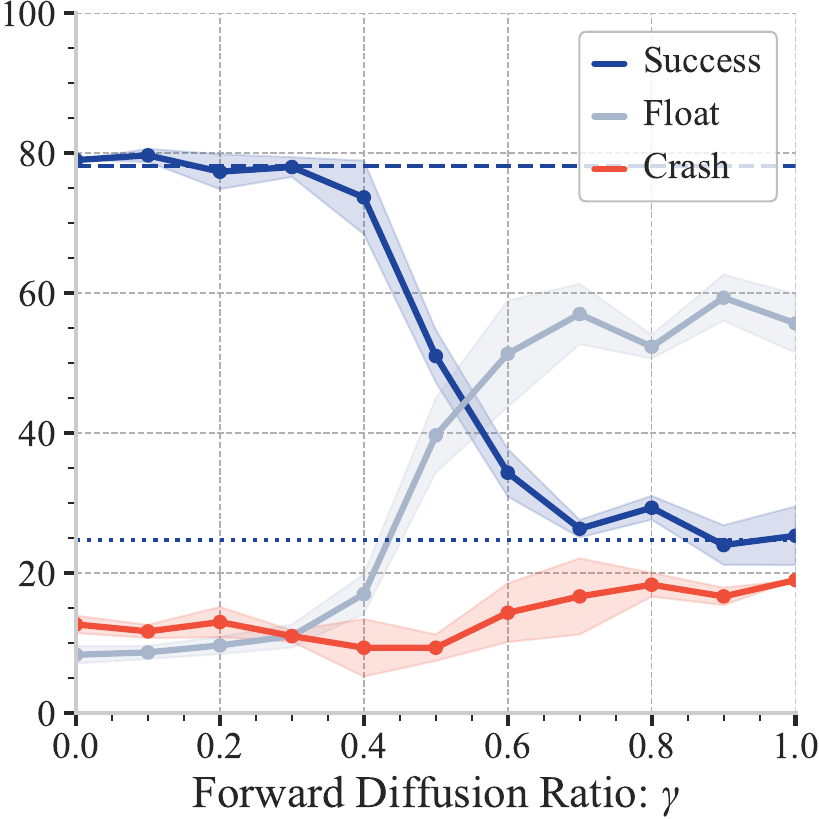}\label{fig:ll_laggy5}}
    \
    \subfloat[$p_\text{laggy}=0.6$]{\includegraphics[width=0.2\linewidth]{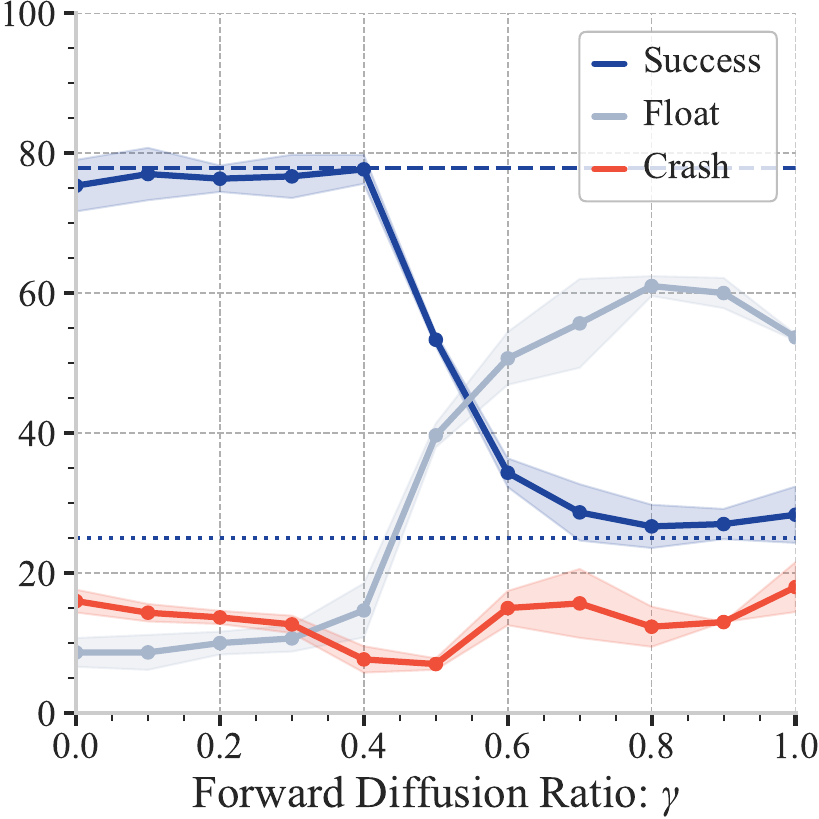}\label{fig:ll_laggy6}}
    \hfil
    \subfloat[$p_\text{laggy}=0.7$]{\includegraphics[width=0.2\linewidth]{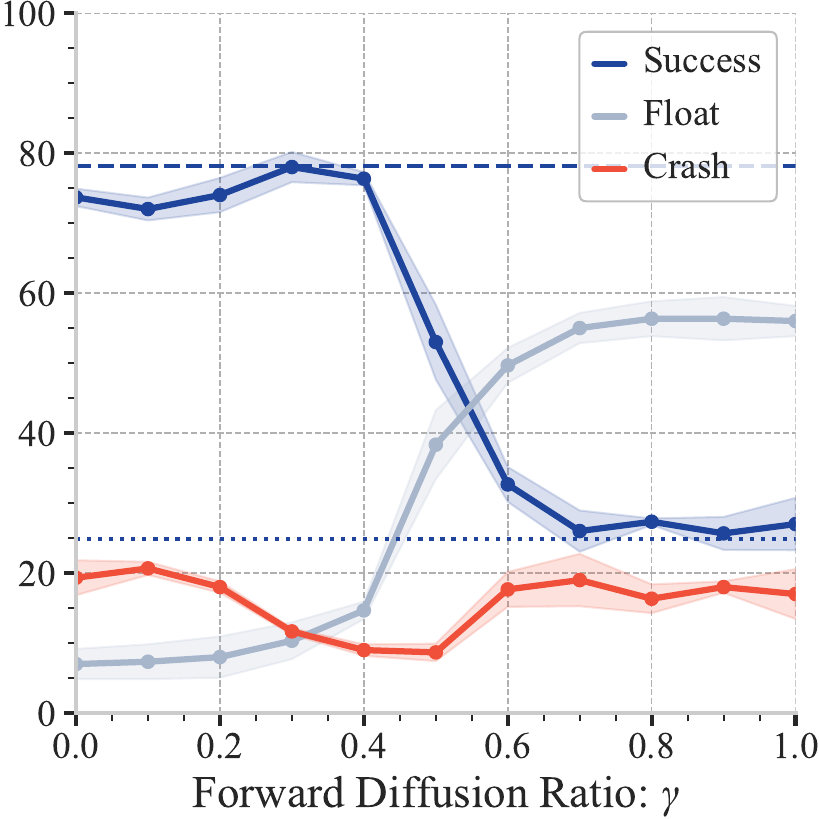}\label{fig:ll_laggy7}}
    \hfil
   \subfloat[$p_\text{laggy}=0.8$]{\includegraphics[width=0.2\linewidth]{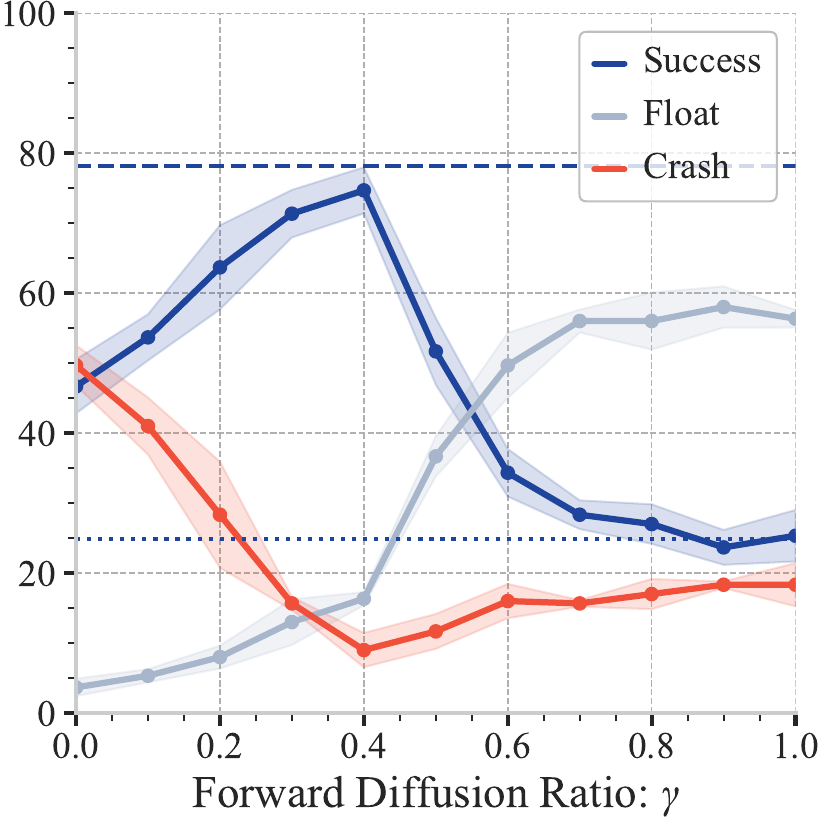}\label{fig:ll_laggy8}}
   \hfil
   \subfloat[$p_\text{laggy}=0.9$]{\includegraphics[width=0.2\linewidth]{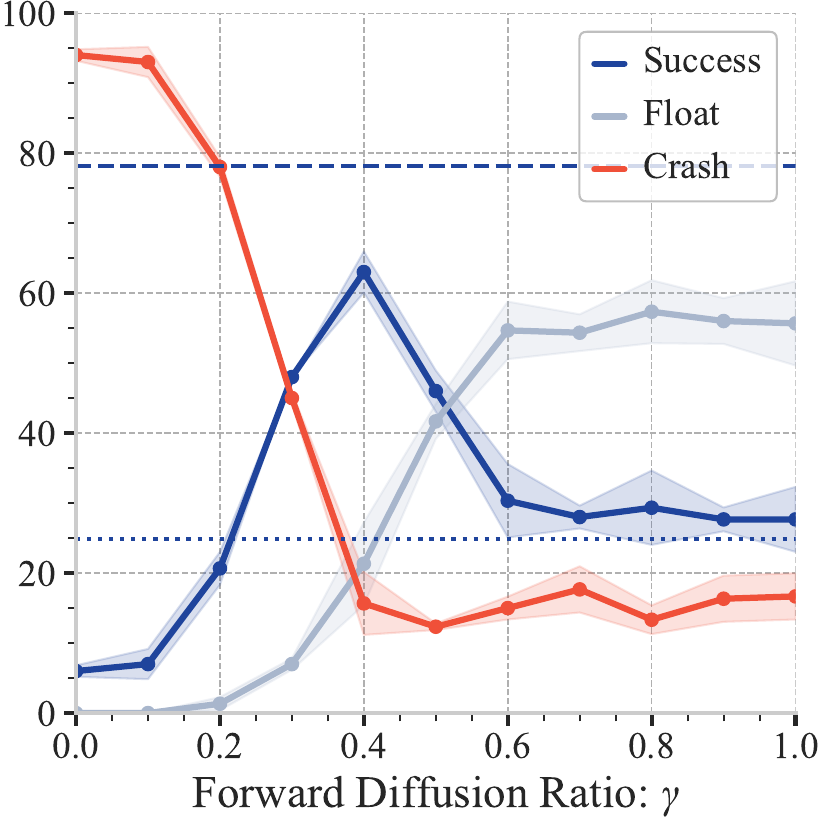}\label{fig:ll_laggy9}}
    
    \caption{Lunar Lander performance as a function of the forward diffusion ratio $\fwr$ for a laggy pilot with values ranging from \subref{fig:ll_laggy2} $p_\text{laggy} = 0.2$ to  \subref{fig:ll_laggy9} $p_\text{laggy} = 0.9$. In all plots, the dashed blue line denotes the success rate of the expert pilot, while the dotted blue line is the success rate of our model when performing ``full'' diffusion (i.e., $\gamma = 1.0$) on an action sampled from a zero-mean isotropic Gaussian distribution, which we refer to as a ``Random'' pilot in the paper.}\label{fig:laggy_pilot_ll_landing_cmp}
\end{figure*}
Figure~\ref{fig:laggy_pilot_ll_landing_cmp} presents the results for the same
set of experiments on Lunar Lander with different parameterizations of the Laggy
pilot. Consistent with previous work~\citep{sha-via-deeprl}, Lunar Lander and,
as we will see, Lunar Reacher are more tolerant of repeated actions than they
are of noisy actions. Consequently, the performance of Laggy Pilots is only
slightly worse than that of an expert pilot until $p_\text{laggy}$ exceeds
$0.8$. As with the Noisy pilot, we see that providing a Laggy pilot with the
assistance of a copilot increases the task success rate, while decreasing the
crash rate up to a forward diffusion ration of $\fwr = 0.4$ for all
parameterizations of the Laggy pilot. Increasing the forward diffusion
ratio beyond $\fwr = 0.4$ decreases the success rate since the copilot does not
have access to the pilot's goal, while the crash rate increases only slightly.

\begin{figure*}
    \centering
    \subfloat[$p_\text{noisy} = 0.2$]{\includegraphics[width=0.2\linewidth]{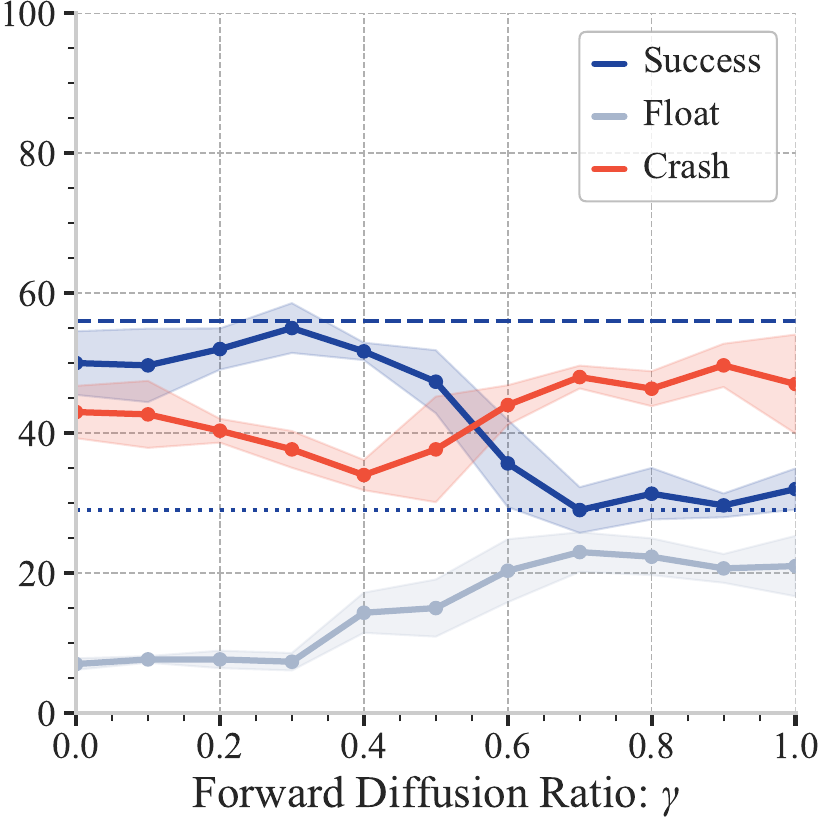}\label{fig:lr_noisy2}}
    \hfil
    \subfloat[$p_\text{noisy} = 0.3$]{\includegraphics[width=0.2\linewidth]{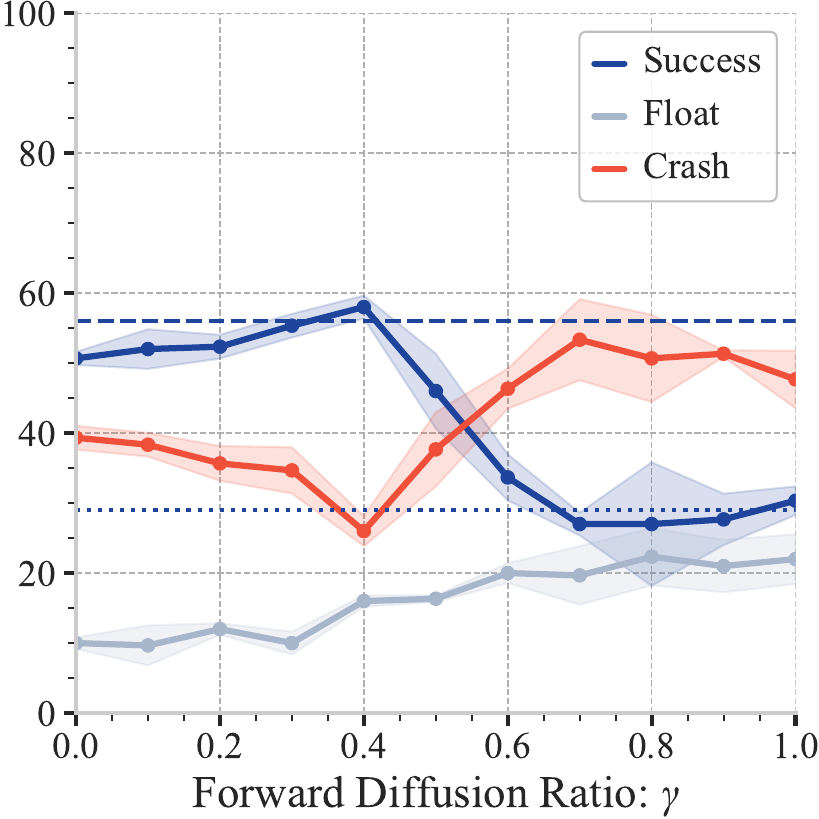}\label{fig:lr_noisy3}}
    \hfil
    \subfloat[$p_\text{noisy} = 0.4$]{\includegraphics[width=0.2\linewidth]{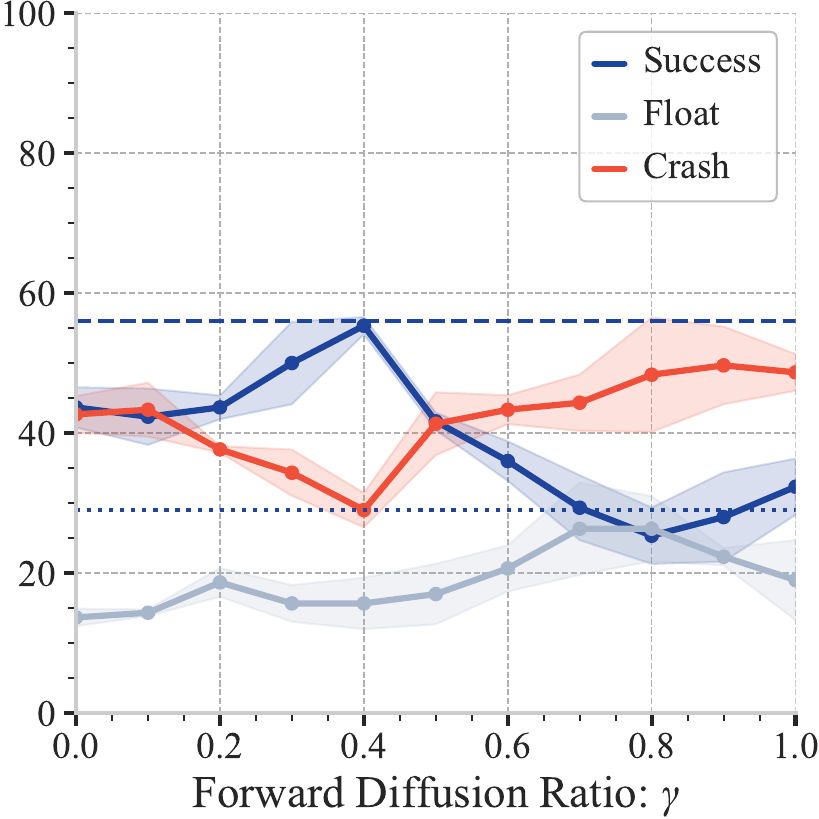}\label{fig:lr_noisy4}}
    \hfil
    \subfloat[$p_\text{noisy} = 0.5$]{\includegraphics[width=0.2\linewidth]{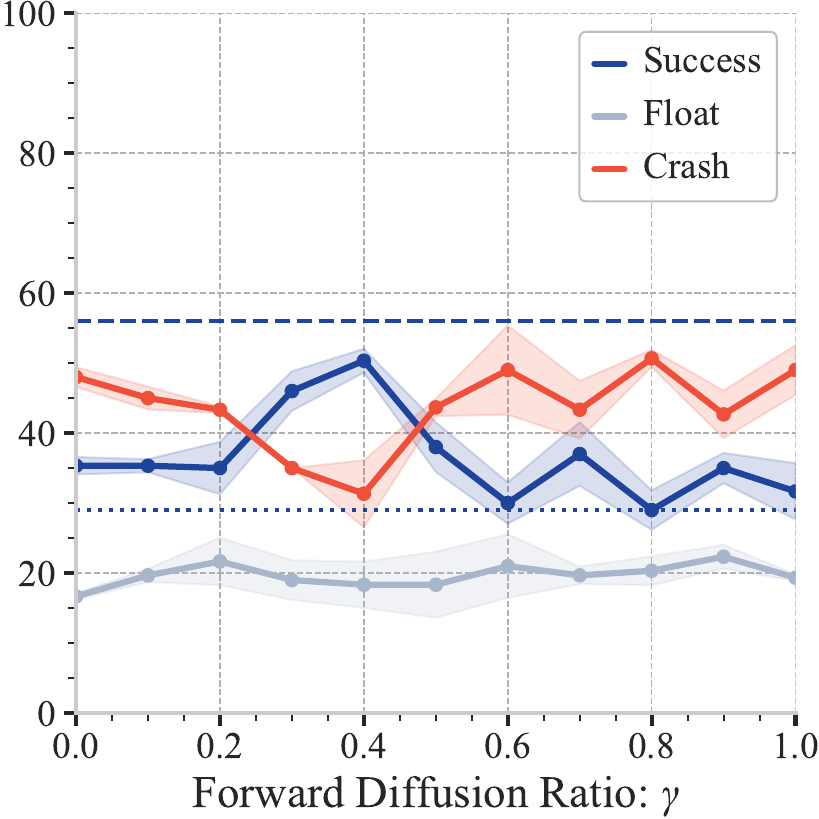}\label{fig:lr_noisy5}}
    \
    \subfloat[$p_\text{noisy} = 0.6$]{\includegraphics[width=0.2\linewidth]{figures/noisy_laggy_prob_cmp/Reaching/reaching_rates_noisy_6.pdf}\label{fig:lr_noisy6}}
    \hfil
    \subfloat[$p_\text{noisy} = 0.7$]{\includegraphics[width=0.2\linewidth]{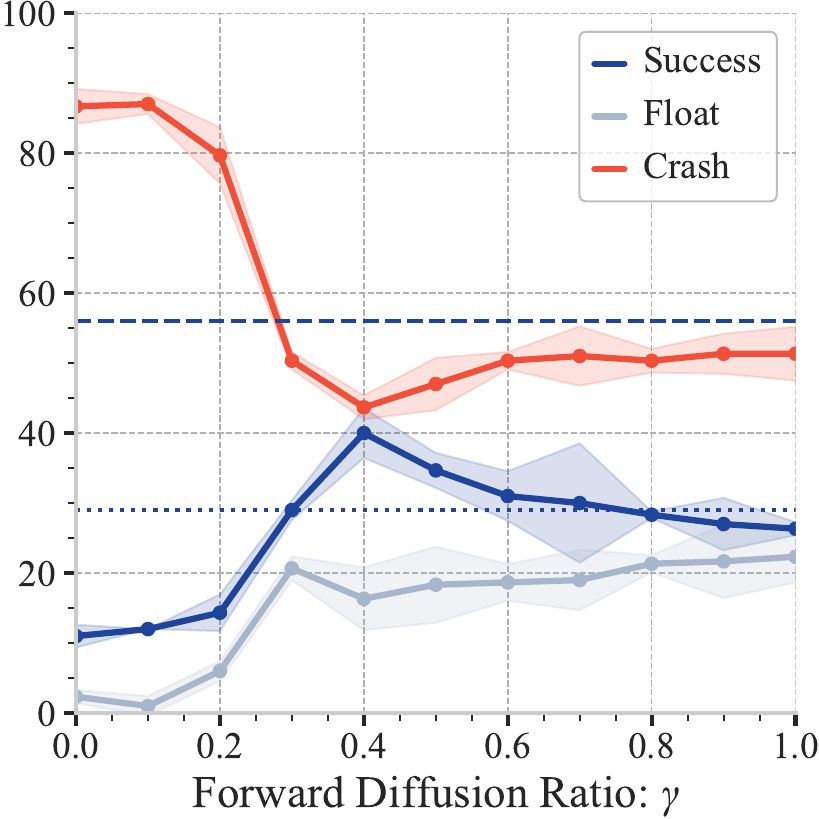}\label{fig:lr_noisy7}}
    \hfil
    \subfloat[$p_\text{noisy} = 0.8$]{\includegraphics[width=0.2\linewidth]{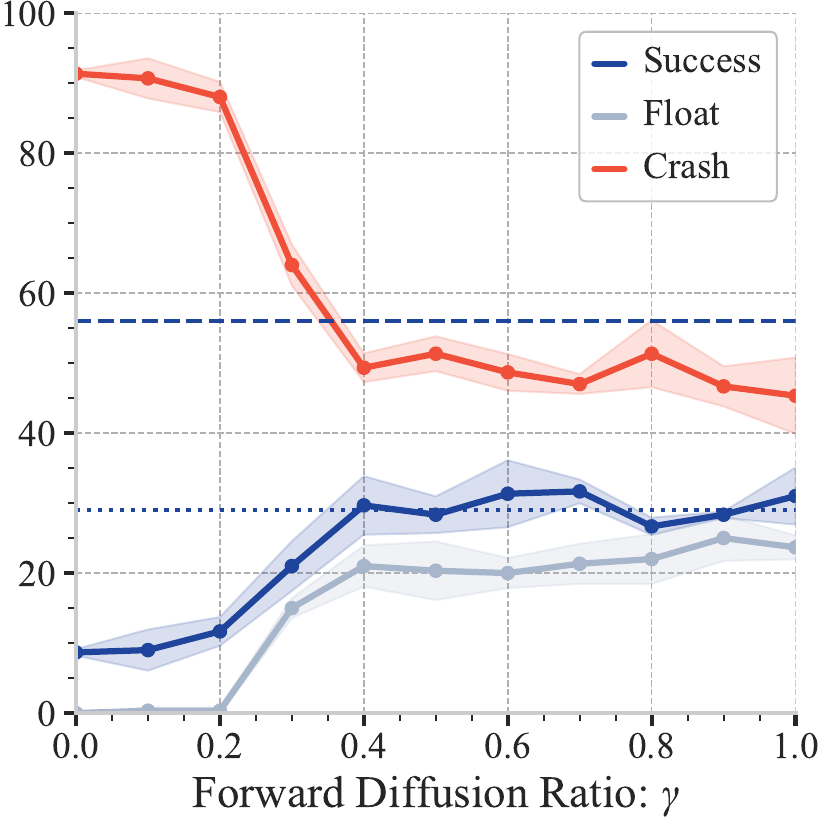}\label{fig:lr_noisy8}}
    \hfil
    \subfloat[$p_\text{noisy} = 0.9$]{\includegraphics[width=0.2\linewidth]{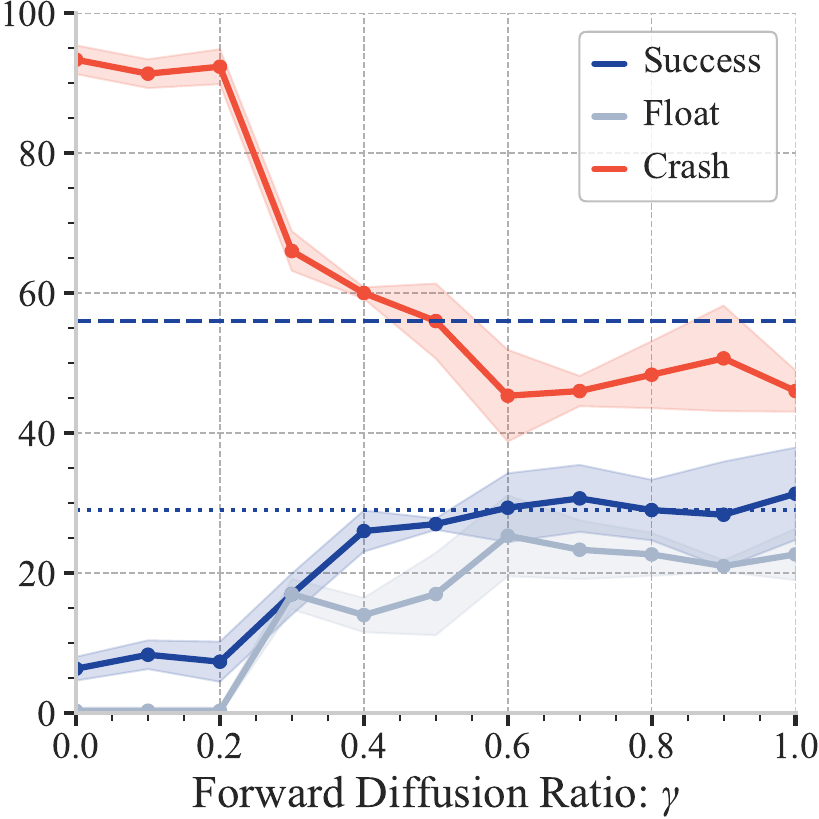}\label{fig:lr_noisy9}}
    \caption{Lunar Reacher performance as a function of the forward diffusion ratio $\fwr$ for a Noisy pilot with noise values ranging from \subref{fig:lr_noisy2} $p_\text{noisy} = 0.2$ to \subref{fig:lr_noisy9} $p_\text{noisy} = 0.9$. In all plots, the dashed blue line denotes the success rate of the expert pilot, while the dotted blue line is the success rate of our model when performing ``full'' diffusion (i.e., $\gamma = 1.0$) on an action sampled from a zero-mean isotropic Gaussian distribution, which we refer to as a ``Random'' pilot in the paper.}\label{fig:noisy_pilot_ll_reaching_cmp}
\end{figure*}
Figure~\ref{fig:noisy_pilot_ll_reaching_cmp} plots the performance of a Noisy pilot on Lunar Reacher for different noise settings. As with Lunar Lander, the success rate of the unassisted Noisy pilot is similar to that of the expert pilot when $p_\text{noisy}$ is low, yet for $p_\text{noisy} \geq 0.6$ the success rate is below $20\%$ and the pilot often goes out-of-bounds as indicated by the high crash rate. Again, we see that the assistance of a copilot increases the task success rate while decreasing the crash rate with a clear peak at $\fwr = 0.4$ for all but the highest noise settings, for which performance only changes slightly as $\fwr$ approaches $1.0$. Again, this decrease in success rate is expected since without access to the region that the pilot is attempting to reach, the copilot chooses actions that are consistent with the set of expert demonstrations on which the diffusion model was trained.

\begin{figure*}
    \centering
    \subfloat[$p_\text{laggy} = 0.2$]{\includegraphics[width=0.2\linewidth]{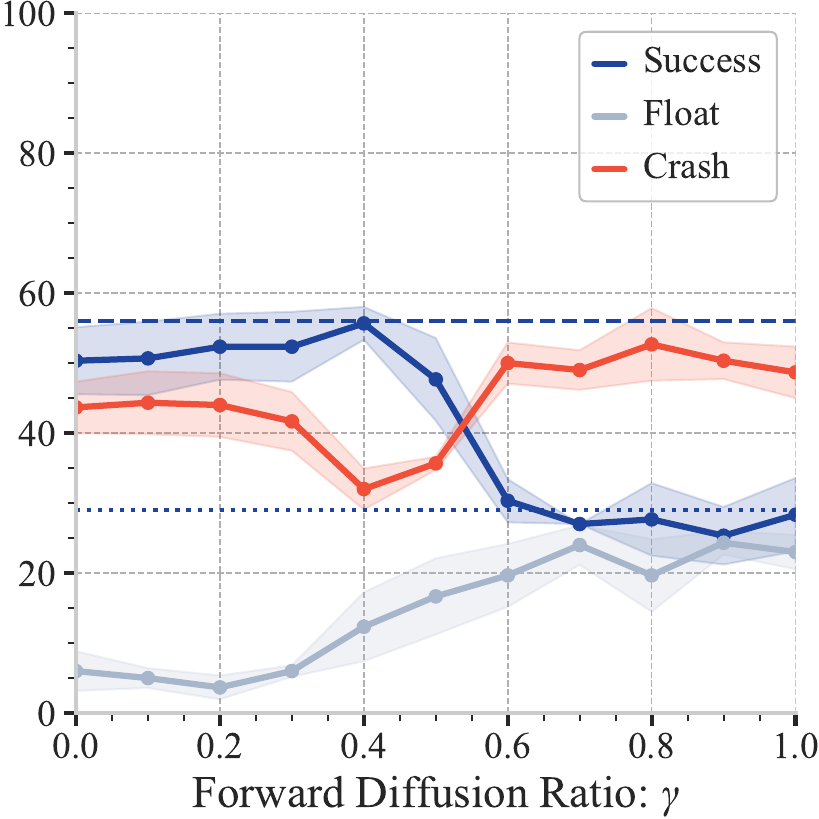}\label{fig:lr_laggy2}}
    \hfil
    \subfloat[$p_\text{laggy} = 0.3$]{\includegraphics[width=0.2\linewidth]{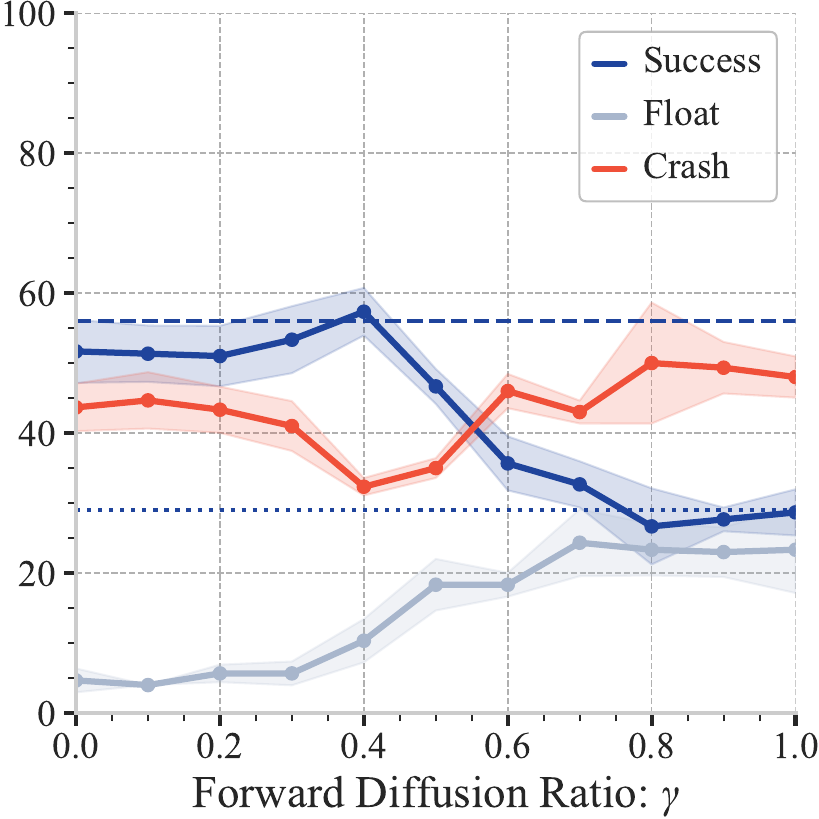}\label{fig:lr_laggy3}}
    \hfil
    \subfloat[$p_\text{laggy} = 0.4$]{\includegraphics[width=0.2\linewidth]{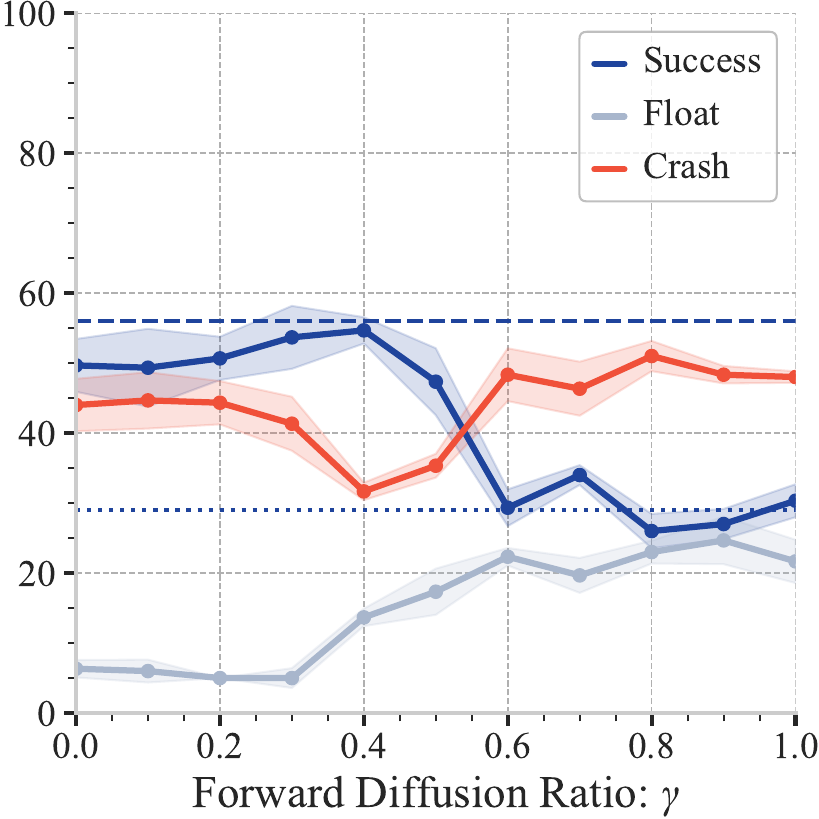}\label{fig:lr_laggy4}}
    \hfil
    \subfloat[$p_\text{laggy} = 0.5$]{\includegraphics[width=0.2\linewidth]{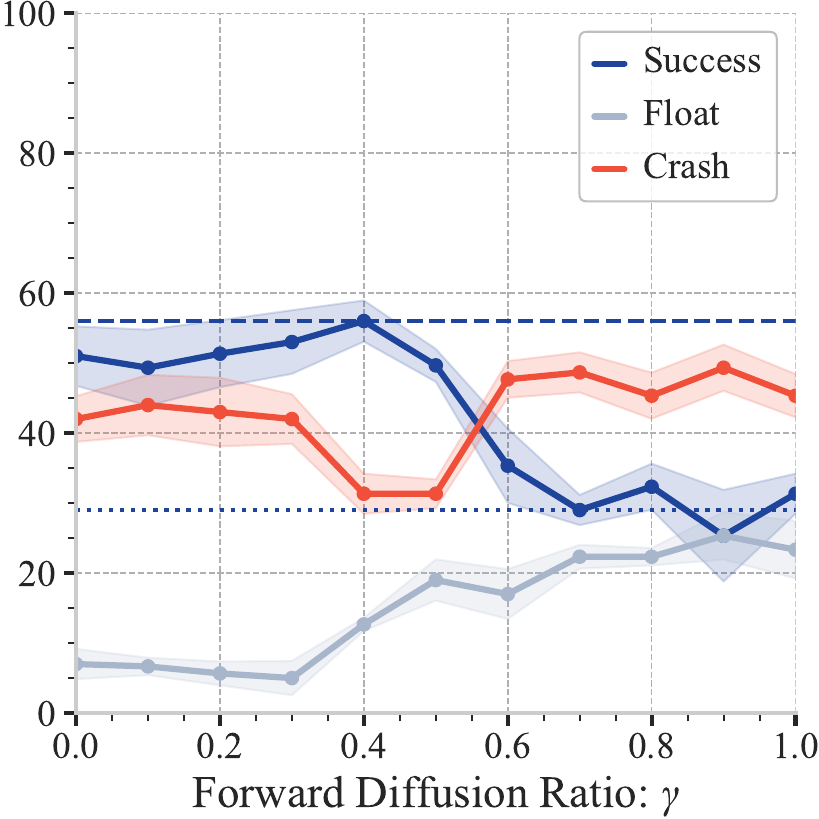}\label{fig:lr_laggy5}}
    \
    \subfloat[$p_\text{laggy} = 0.6$]{\includegraphics[width=0.2\linewidth]{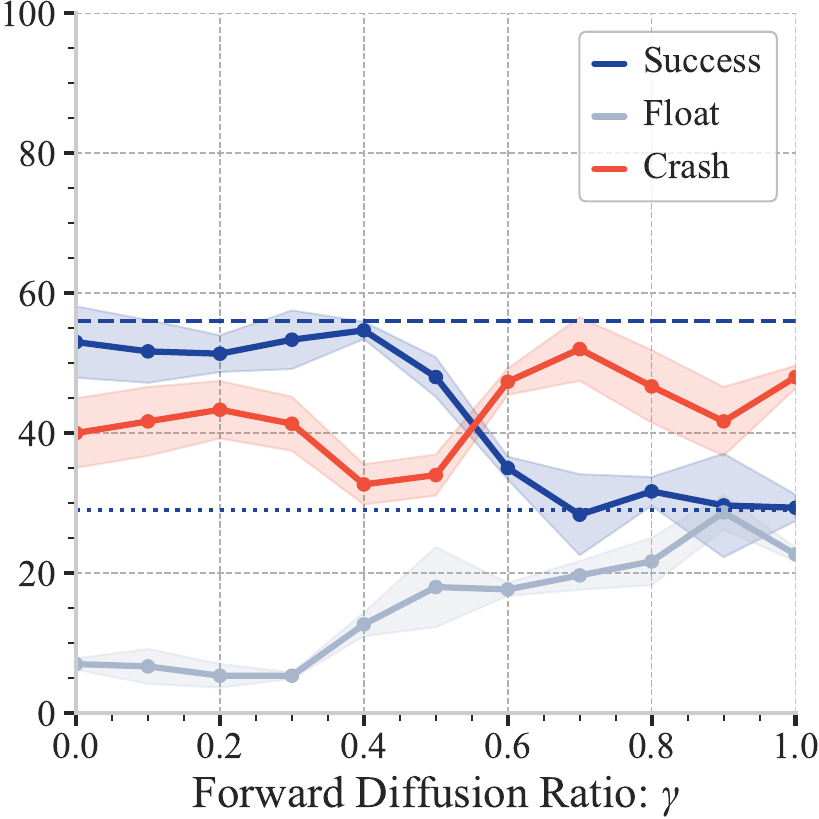}\label{fig:lr_laggy6}}
    \hfil
    \subfloat[$p_\text{laggy} = 0.7$]{\includegraphics[width=0.2\linewidth]{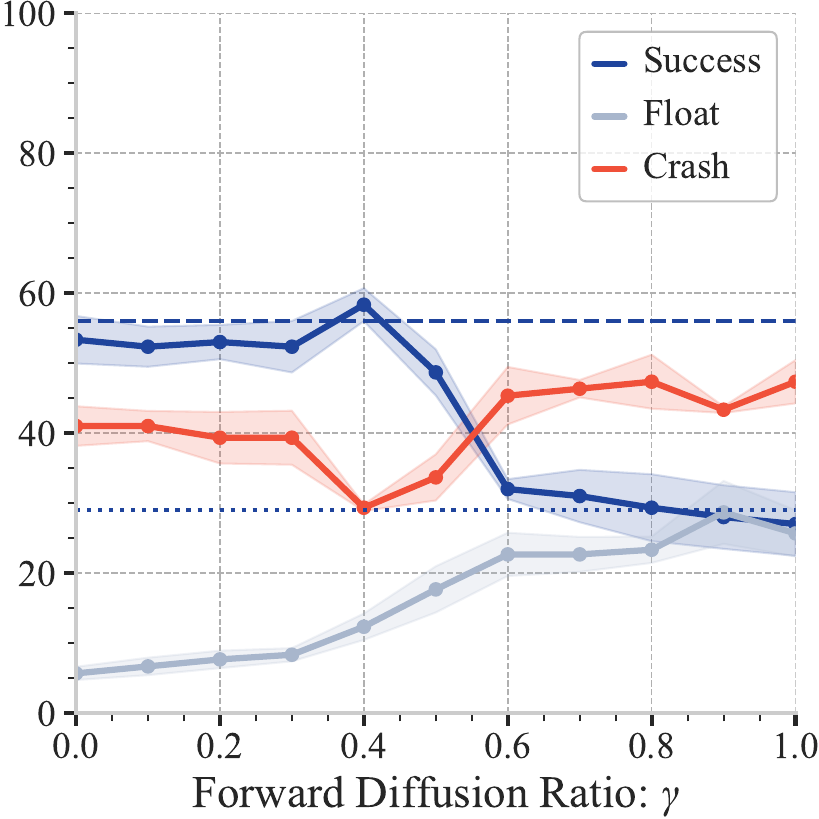}\label{fig:lr_laggy7}}
    \hfil
    \subfloat[$p_\text{laggy} = 0.8$]{\includegraphics[width=0.2\linewidth]{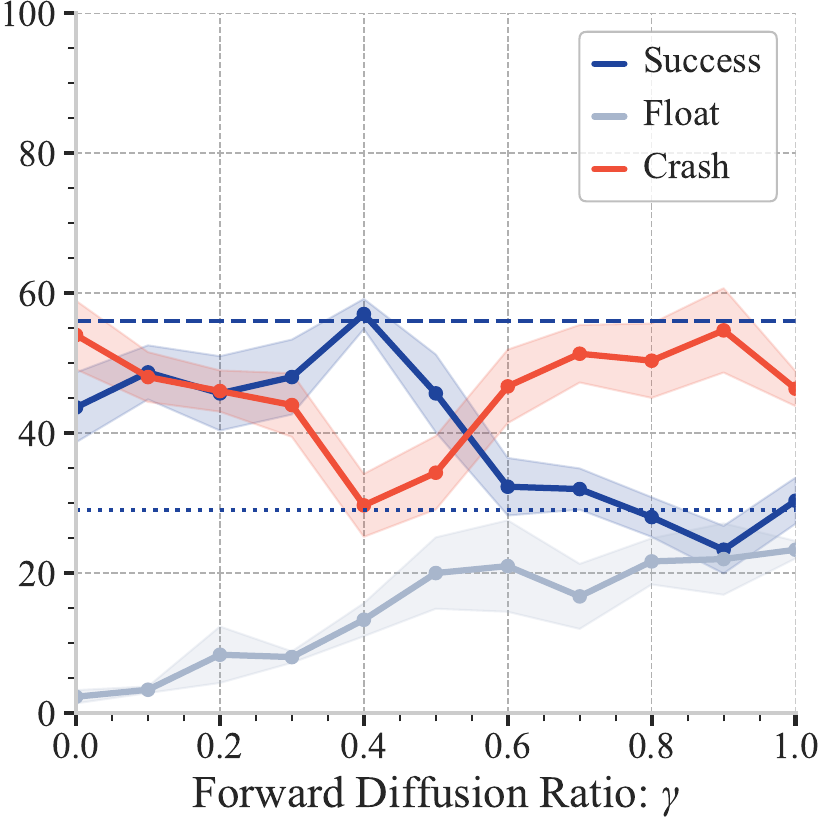}\label{fig:lr_laggy8}}
    \hfil
     \subfloat[$p_\text{laggy} = 0.9$]{\includegraphics[width=0.2\linewidth]{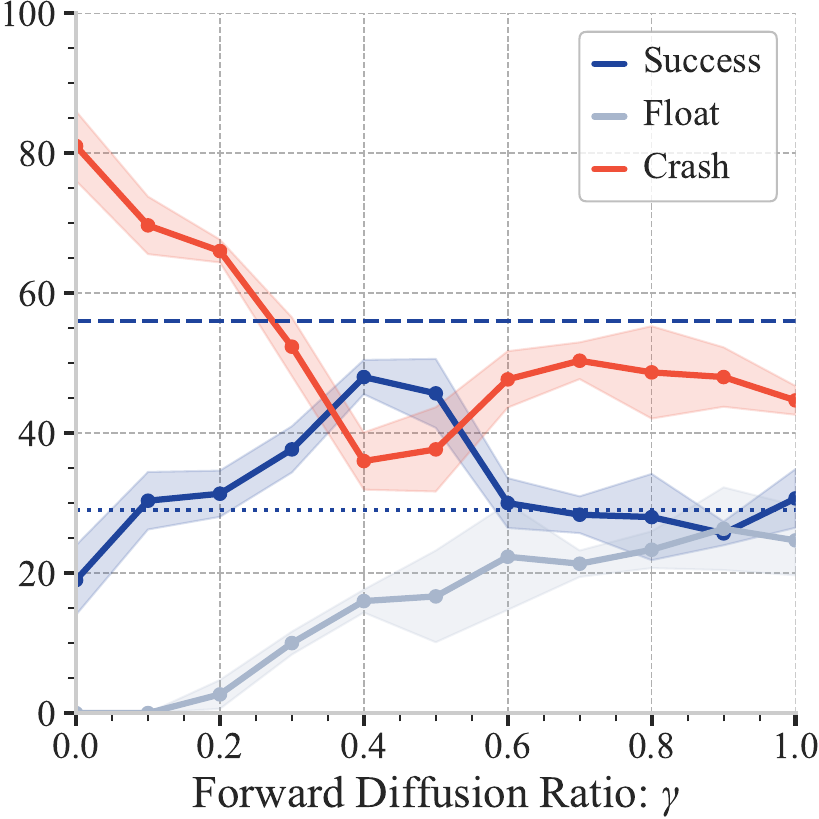}\label{fig:lr_laggy9}}
    
    \caption{Lunar Reacher performance as a function of the forward diffusion ratio $\fwr$ for a Laggy pilot with values ranging from \subref{fig:lr_laggy2} $p_{\text{laggy}} = 0.2$ to  \subref{fig:lr_laggy9} $p_{\text{laggy}} = 0.9$. In all plots, the dashed blue line denotes the success rate of the expert pilot, while the dotted blue line is the success rate of our model when performing ``full'' diffusion (i.e., $\gamma = 1.0$) on an action sampled from a zero-mean isotropic Gaussian distribution, which we refer to as a ``Random'' pilot in the paper.}\label{fig:laggy_pilot_ll_reaching_cmp}
\end{figure*}
Figure~\ref{fig:laggy_pilot_ll_reaching_cmp} plots the performance of the Laggy pilot on the Lunar Reacher task for different settings of $p_\text{laggy}$. Similar to Lunar Lander, we find that Lunar Reacher is tolerant of an otherwise (near-)expert policy that randomly repeats its previous actions. Due to the nature of the task, the success rate of the Laggy Pilot is only slightly worse than that of the expert until $p_\text{laggy}$ exceeds $0.8$. For all parameterizations of the Laggy pilot, we see that the assistance of a copilot with $\fwr = 0.4$ decreases the crash rate, while simultaneously maintaining or, in the case of $p_\text{laggy} \geq 0.8$, increasing the success rate.

\begin{table*}[ht]
    \centering
    \caption{Success and crash/out-of-bounds (OOB) rates on Lunar Lander and Lunar Reacher for different pilots with and without assistance, where we show the results for our chosen value for $\gamma = 0.4$  as well as the result of using full diffusion ($\gamma = 1.0$), which has the effect of removing information about the goal provided by the pilot. The plots corresponding to the data in the table is figure \ref{fig:ll-land-assisted}. 
    Each entry corresponds to $10$ episodes across $30$ random seeds. Note that the Zero and Random pilots have no knowledge of the goal.}
    \label{tb:lunar-lander-reacher-appendix}
    \begin{tabularx}{1.0\linewidth}{lYYYYYY}%
        \toprule
        & \multicolumn{3}{c}{Success Rate} & \multicolumn{3}{c}{Crash/OOB Rate}\\
        Pilot & w/o Copilot ($\gamma=0.0$) & w/ Copilot (ours) ($\gamma=0.4$) & w/ Copilot ($\gamma=1.0$) & w/o Copilot ($\gamma=0.0$) & w/ Copilot (ours) ($\gamma=0.4$) & w/ Copilot ($\gamma=1.0$)\\
        \midrule[0.75pt]
        \multicolumn{7}{c}{\bf Lunar Lander}\\
        \midrule
        Noisy & $ 20.67 \pm 4.50 $ &  $\bm{ 68.00 \pm 5.35 }$ &  $ 29.00 \pm 2.94 $
        & $ \hphantom{0}28.33 \pm 2.62 $ &  $ \bm{7.67 \pm 2.87 }$ &  $ 18.33 \pm 4.92 $\\
        Laggy & $ 21.33 \pm 2.05 $ &  $\bm{ 75.00 \pm 3.56} $ &  $ 26.33 \pm 4.78 $
        & $ \hphantom{0}76.67 \pm 2.49 $ &  $\bm{ 9.67 \pm 3.86 }$ &  $ 15.67 \pm 0.94 $\\
        \midrule[0.1pt]
        Zero & $ \hphantom{0}0.00 \pm 0.00 $ &  $ 27.00 \pm 0.82 $ &  $ 28.00 \pm 2.16 $
        & $ 100.00 \pm 0.00 $ &  $ 19.00 \pm 2.94 $ &  $ 13.33 \pm 2.49 $\\
        Random & $ \hphantom{0}0.00 \pm 0.00 $ &  $ 25.00 \pm 4.32 $ &  $ 27.67 \pm 1.70 $ 
        & $ 100.00 \pm 0.00 $ &  $ 19.33 \pm 5.25 $ &  $ 15.67 \pm 1.25 $\\
        Expert & $ 77.67 \pm 2.62 $ &  $ 78.67 \pm 2.87 $ &  $ 29.33 \pm 6.18 $
        & $ \hphantom{0}12.33 \pm 0.94 $ &  $ \hphantom{0}8.00 \pm 1.63 $ &  $ 15.33 \pm 1.70 $\\                          
        \midrule[0.75pt]
        \multicolumn{7}{c}{\bf Lunar Reacher}\\
        \midrule
        Noisy & $ 14.33 \pm 2.49 $ &  $ \bm{45.33 \pm 3.30} $ &  $ 30.67 \pm 4.11 $
        & $ \hphantom{0}77.33 \pm 3.09 $ &  $ \bm{38.00 \pm 2.94} $ &  $ 47.00 \pm 3.56 $\\
        Laggy & $ 30.67 \pm 5.56 $ &  $ \bm{55.33 \pm 6.13} $ &  $ 30.67 \pm 2.62 $
        & $ \hphantom{0}69.33 \pm 5.56 $ &  $ \bm{31.33 \pm 2.87} $ &  $ 45.67 \pm 0.94 $\\
        \midrule[0.1pt]
        Zero & $ \hphantom{0}0.00 \pm 0.00 $ &  $ 19.67 \pm 2.62 $ &  $ 31.00 \pm 6.98 $ 
        & $ 100.00 \pm 0.00 $ &  $ 58.33 \pm 3.09 $ &  $ 45.33 \pm 5.56 $\\
        Random & $ \hphantom{0}4.33 \pm 1.89 $ &  $ 22.33 \pm 2.87 $ &  $ 30.00 \pm 4.08 $
        & $ \hphantom{0}95.33 \pm 1.70 $ &  $ 59.00 \pm 0.82 $ &  $ 45.00 \pm 2.16 $\\
        Expert & $ 49.33 \pm 4.78 $ &  $ 55.00 \pm 2.16 $ &  $ 28.67 \pm 4.50 $
        & $ \hphantom{0}44.00 \pm 3.56 $ &  $ 31.67 \pm 2.87 $ &  $ 50.00 \pm 1.41 $\\ 
        \bottomrule
    \end{tabularx}
\end{table*}
Table~\ref{tb:lunar-lander-reacher-appendix} expands upon the analysis
presented in Table~\ref{tb:lunar-lander-reacher} and makes explicit the effect
of different settings for the forward diffusion ratio between $\fwr = 0.0$ (no
assistance), our chosen setting of $\fwr = 0.4$, and to $\fwr = 1.0$ (full
assistance) when paired with different pilots. Consistent with the plots in Figures~\ref{fig:noisy_pilot_ll_landing_cmp}--\ref{fig:laggy_pilot_ll_reaching_cmp}, setting the forward diffusion ratio to $\fwr = 1.0$ results in similar performance across all pilots both in terms of success rate and crash rate, consistent with the copilot selecting actions that \emph{conform} to the distribution over expert actions.

\subsection{Block Pushing}
\begin{figure*}
    \centering
    \subfloat[$p_\text{noisy} = 0.2$]{\includegraphics[width=0.2\linewidth]{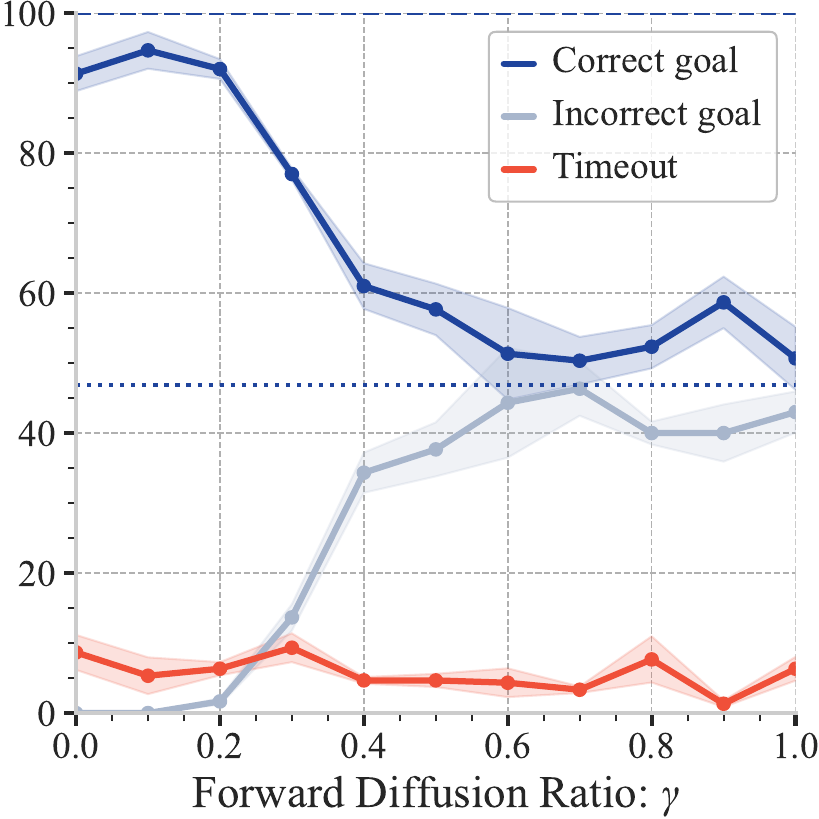}\label{fig:bp_noisy2}}
    \hfil
    \subfloat[$p_\text{noisy} = 0.3$]{\includegraphics[width=0.2\linewidth]{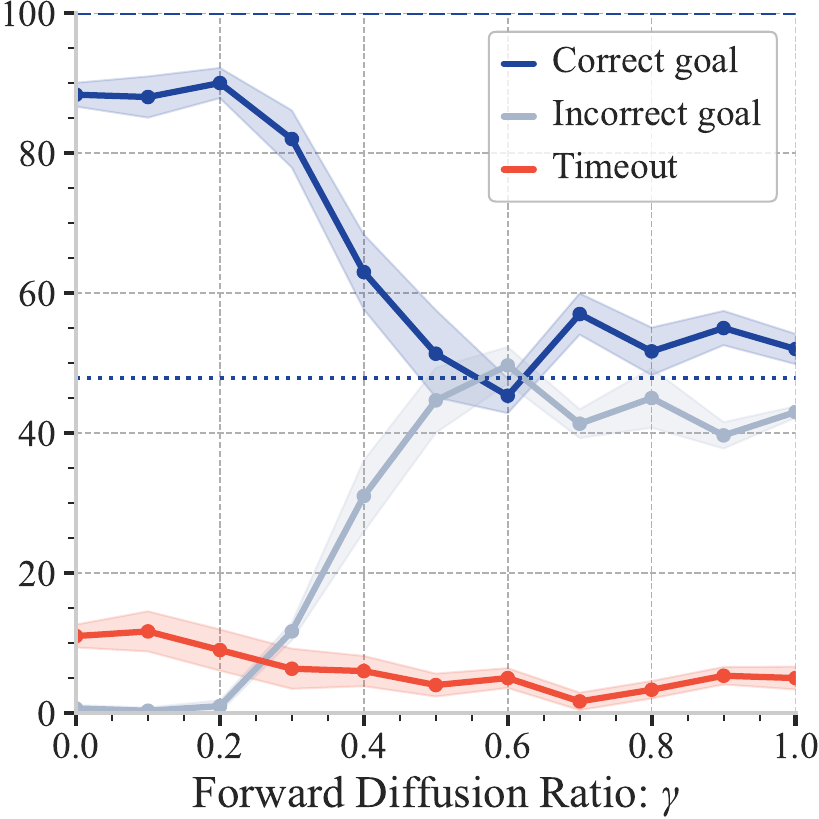}\label{fig:bp_noisy3}}
    \hfil
    \subfloat[$p_\text{noisy} = 0.4$]{\includegraphics[width=0.2\linewidth]{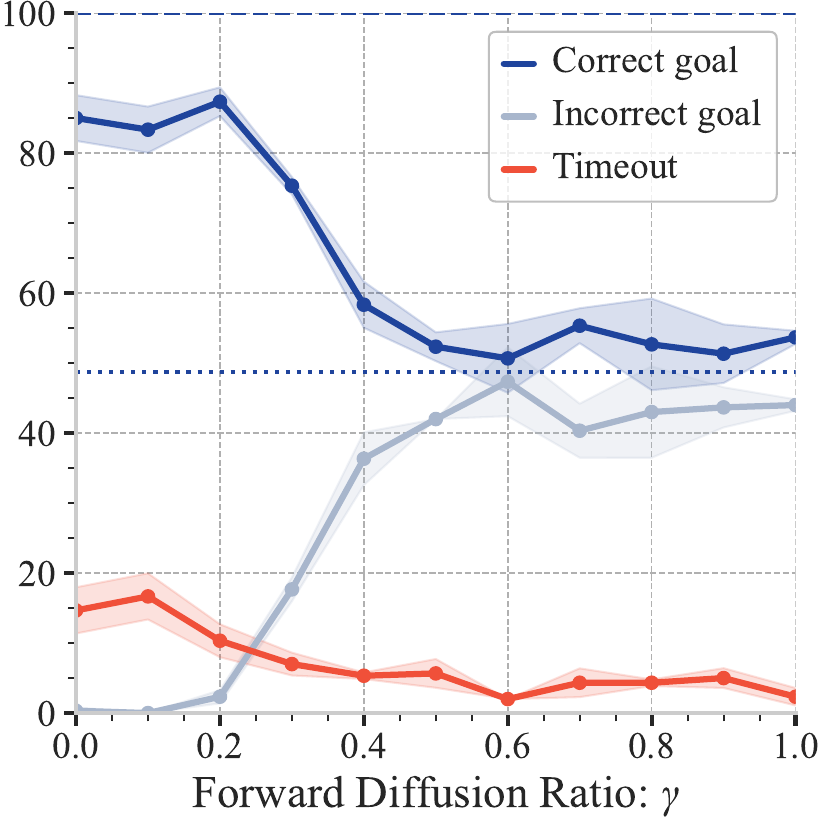}\label{fig:bp_noisy4}}
    \hfil
    \subfloat[$p_\text{noisy} = 0.5$]{\includegraphics[width=0.2\linewidth]{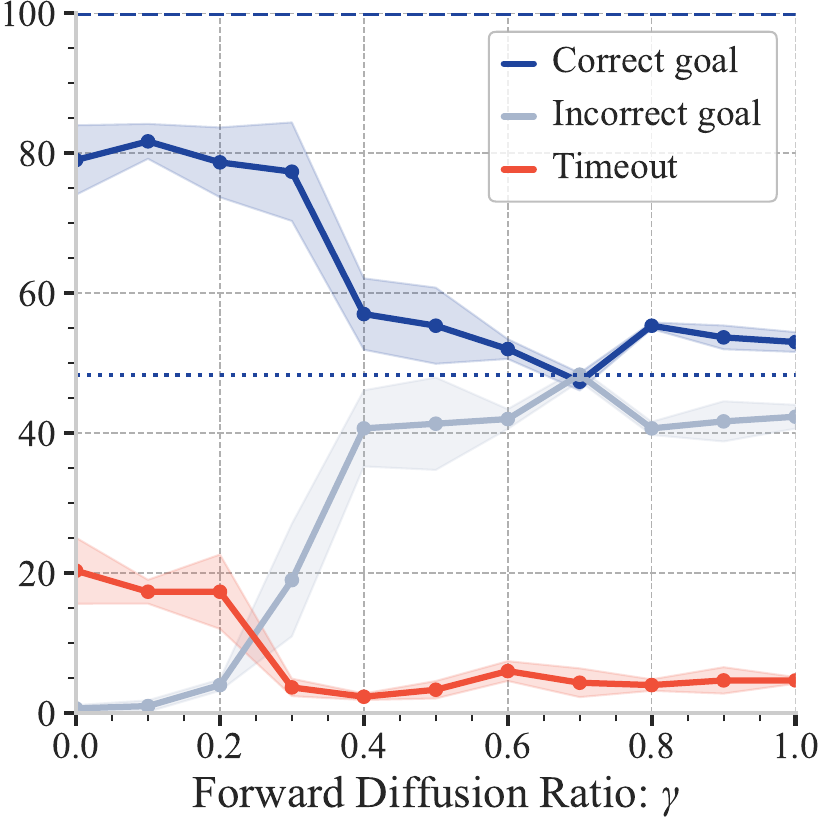}\label{fig:bp_noisy5}}
    \
    \subfloat[$p_\text{noisy} = 0.6$]{\includegraphics[width=0.2\linewidth]{figures/noisy_laggy_prob_cmp/BP/bp_rates_noisy_6.pdf}\label{fig:bp_noisy6}}
    \hfil
    \subfloat[$p_\text{noisy} = 0.7$]{\includegraphics[width=0.2\linewidth]{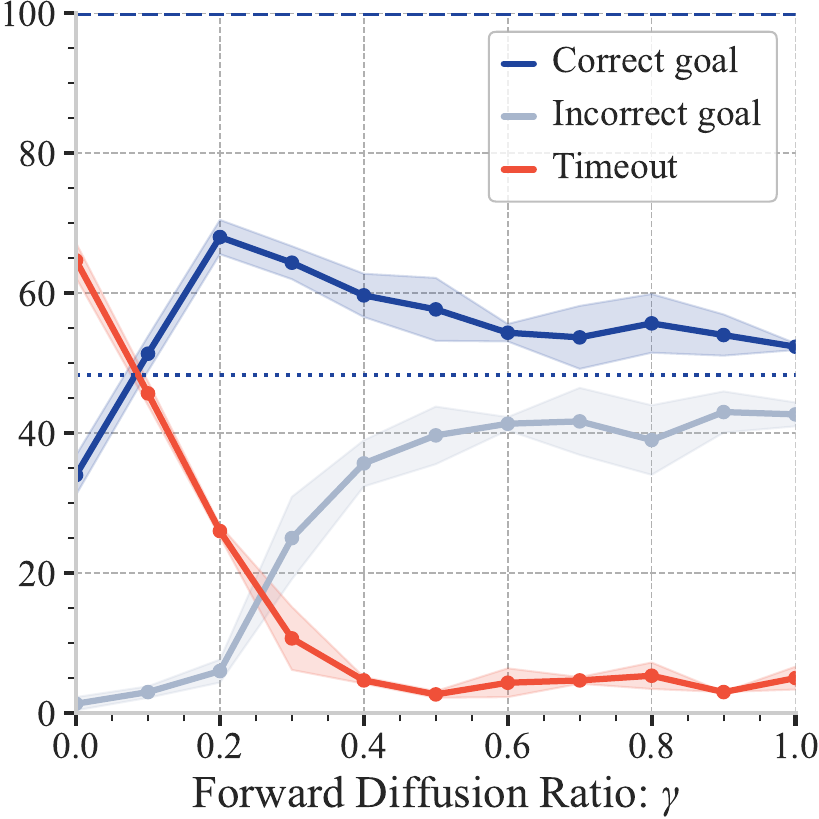}\label{fig:bp_noisy7}}
    \hfil
    \subfloat[$p_\text{noisy} = 0.8$]{\includegraphics[width=0.2\linewidth]{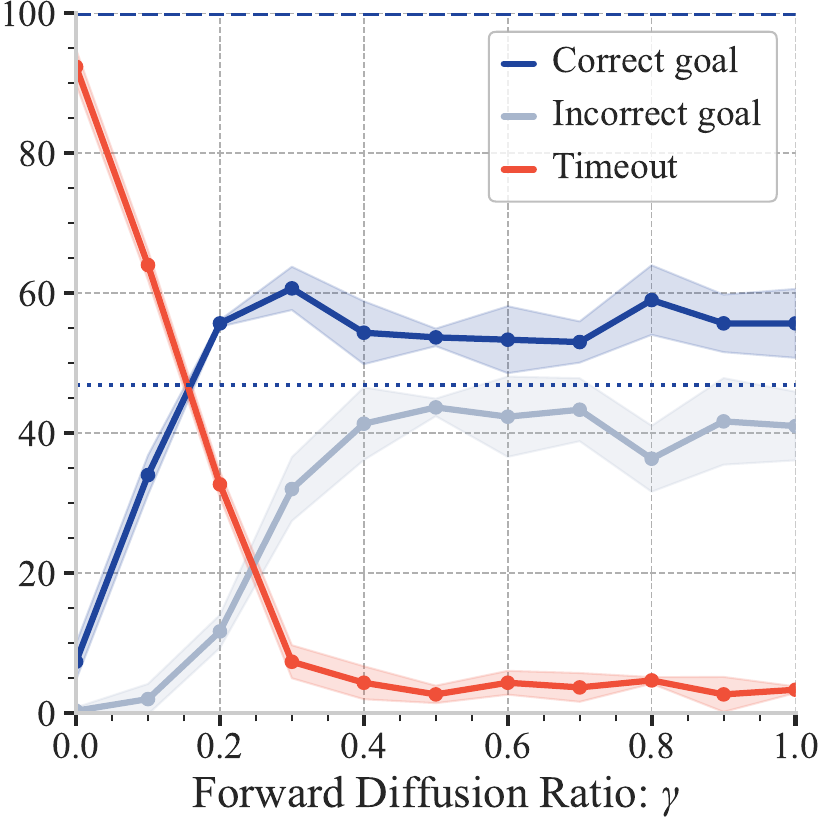}\label{fig:bp_noisy8}}
    \hfil
    \subfloat[$p_\text{noisy} = 0.9$]{\includegraphics[width=0.2\linewidth]{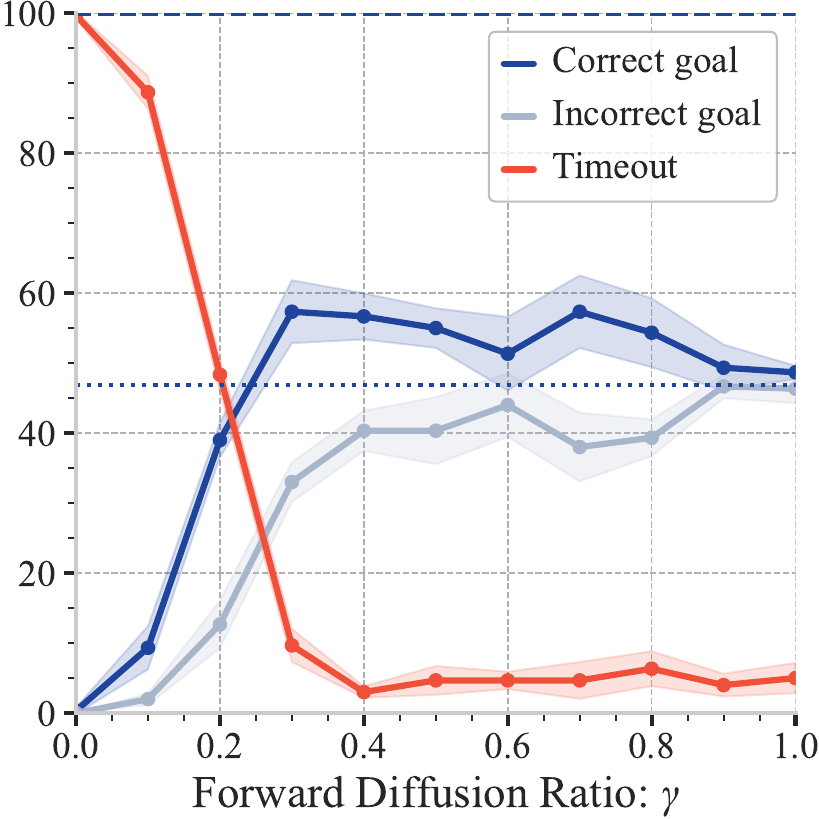}\label{fig:bp_noisy9}}
    \caption{Block Pushing performance in terms of the correct goal, wrong goal, and timeout rates as a function of the forward diffusion ratio $\fwr$ for a Noisy pilot with noise values ranging from \subref{fig:bp_noisy2} $p_{\text{noisy}} = 0.2$ \subref{fig:bp_noisy9} $p_{\text{noisy}} = 0.9$. In all plots, the dashed blue line denotes the success rate of the expert pilot, while the dotted blue line is the success rate of our model when performing ``full'' diffusion (i.e., $\gamma = 1.0$) on an action sampled from a zero-mean isotropic Gaussian distribution, which we refer to as a ``Random'' pilot in the paper.}\label{fig:noisy_pilot_bp_cmp}
\end{figure*}
Figure~\ref{fig:noisy_pilot_bp_cmp} provides plots of the performance of the Noisy pilot on the Block Pushing task in terms of the correct goal, incorrect goal, and timeout rates for different settings of $p_\text{noisy}$. When the noise level is at or below $0.5$, the Noisy pilot pushes the block to the correct goal more than $75\%$ of the time without assistance (vs.\ nearly $100\%$ for the expert) and never pushes the block to the incorrect goal. As the noise level increases from $p_\text{noise} = 0.6$, the rate at which the unassisted Noisy pilot times out significantly increases, while the rate with which it reaches the correct goal precipitously drops for $p_\text{noisy} \geq 0.8$. With the assistance of a copilot, the rate at which the pilot pushes the block to the correct goal increases and the timeout rate decreases, both significantly, as the forward diffusion ratio increases to $\fwr = 0.2$ for Noisy pilots with $p_\text{noisy} \geq 0.6$. As the forward diffusion ratio increases further and information about the correct goal is diminished, the assisted policy reaches one of the two candidate goals with roughly equal likelihood, while the timeout rate remains low, consistent with the set of expert demonstrations. For Noisy pilots with $p_\text{noise} \leq 0.5$, assistance does not negatively affect performance until the forward diffusion ratio exceeds $\gamma = 0.2$.

\begin{figure*}
    \centering
    \subfloat[$p_\text{laggy} = 0.2$]{\includegraphics[width=0.2\linewidth]{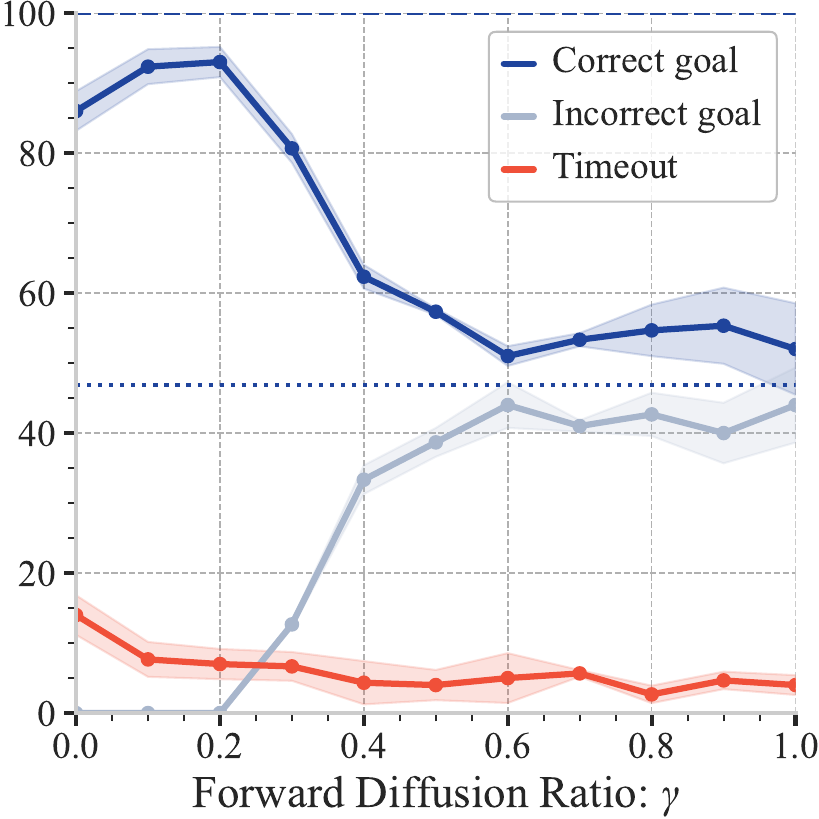}\label{fig:bp_laggy2}}
    \hfil
    \subfloat[$p_\text{laggy} = 0.3$]{\includegraphics[width=0.2\linewidth]{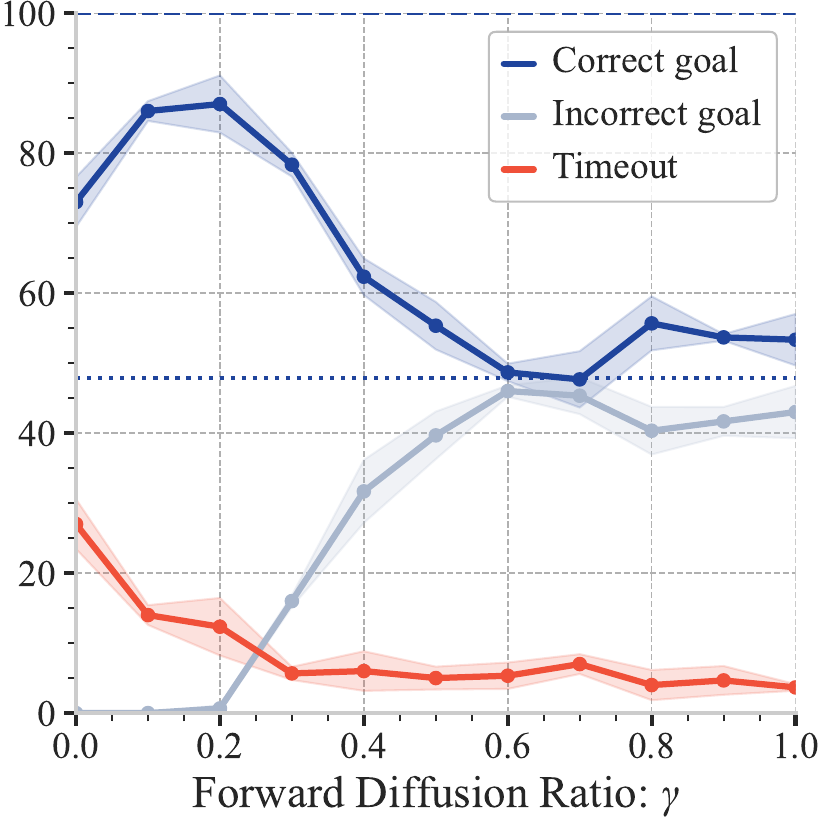}\label{fig:bp_laggy3}}
    \hfil
    \subfloat[$p_\text{laggy} = 0.4$]{\includegraphics[width=0.2\linewidth]{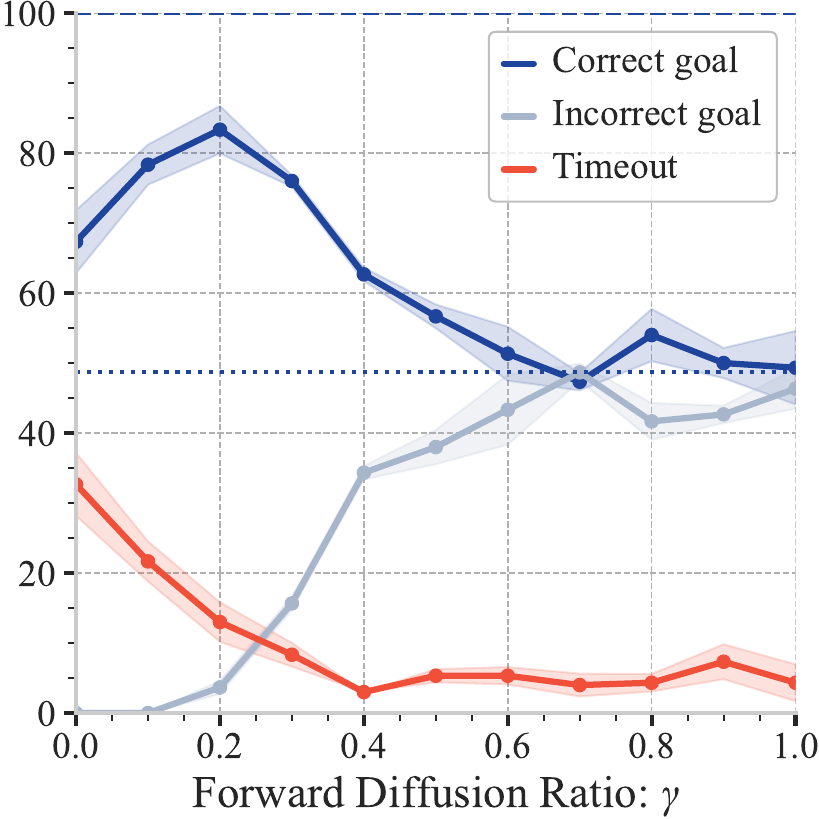}\label{fig:bp_laggy4}}
    \hfil
    \subfloat[$p_\text{laggy} = 0.5$]{\includegraphics[width=0.2\linewidth]{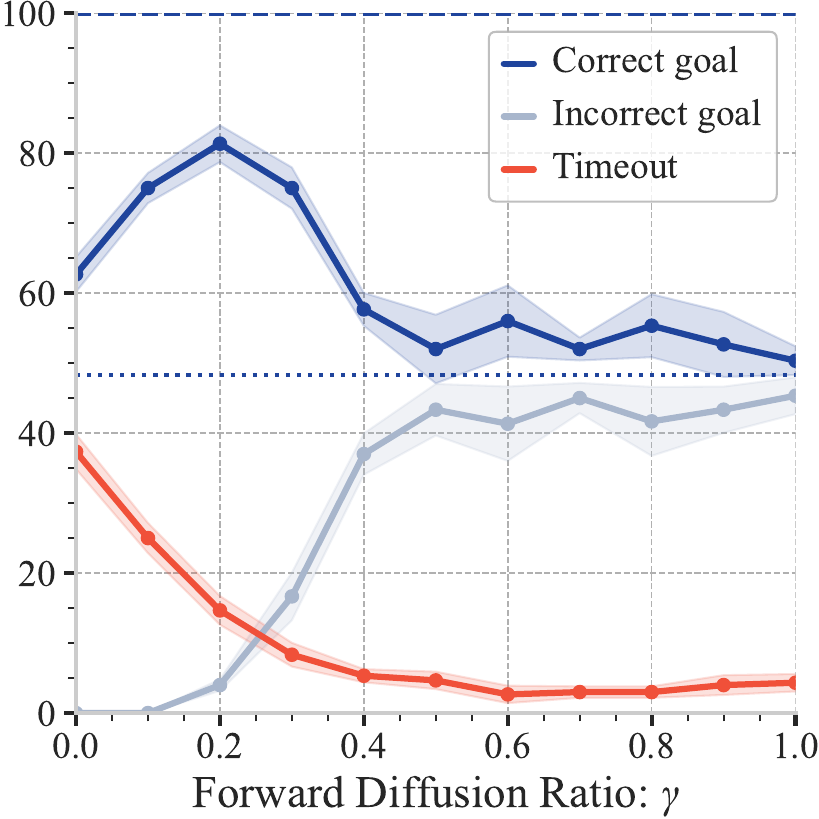}\label{fig:bp_laggy5}}
    \
    \subfloat[$p_\text{laggy} = 0.6$]{\includegraphics[width=0.2\linewidth]{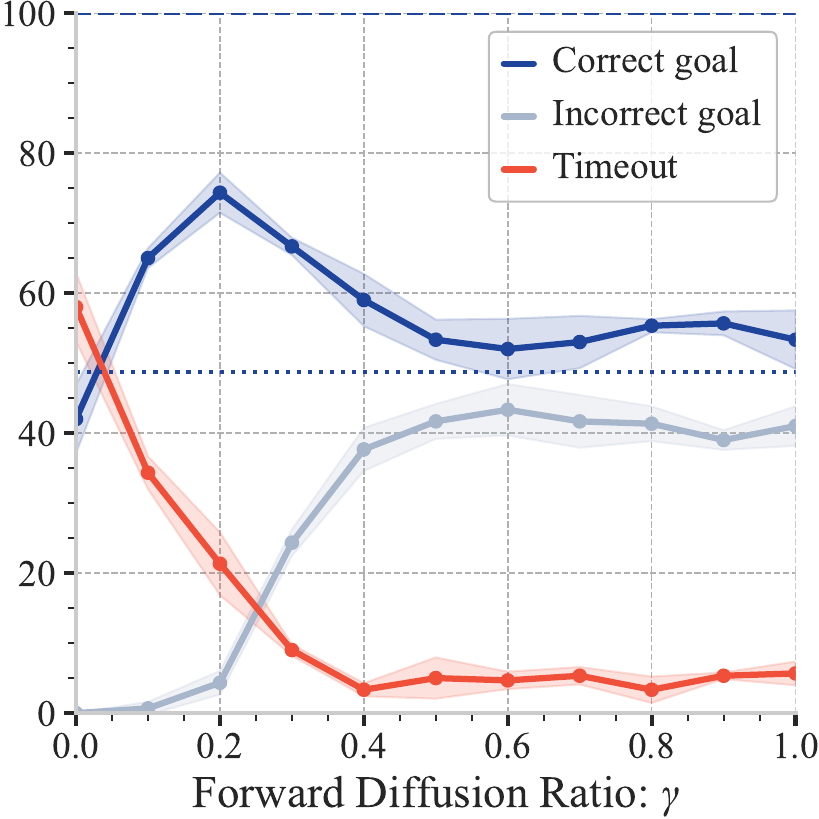}\label{fig:bp_laggy6}}
    \hfil
    \subfloat[$p_\text{laggy} = 0.7$]{\includegraphics[width=0.2\linewidth]{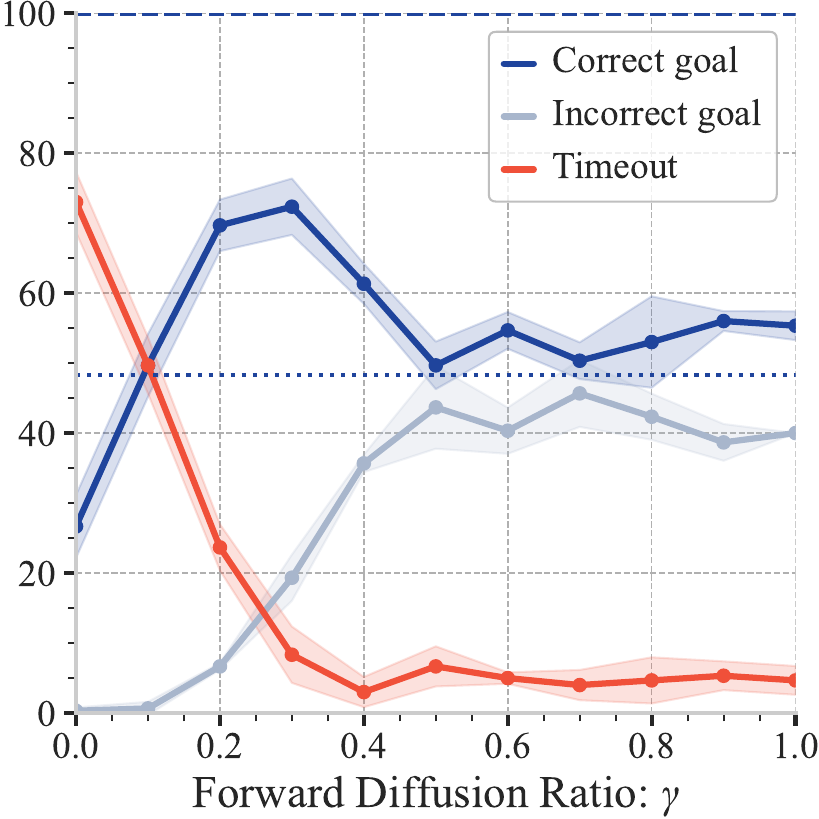}\label{fig:bp_laggy7}}
    \hfil
    \subfloat[$p_\text{laggy} = 0.8$]{\includegraphics[width=0.2\linewidth]{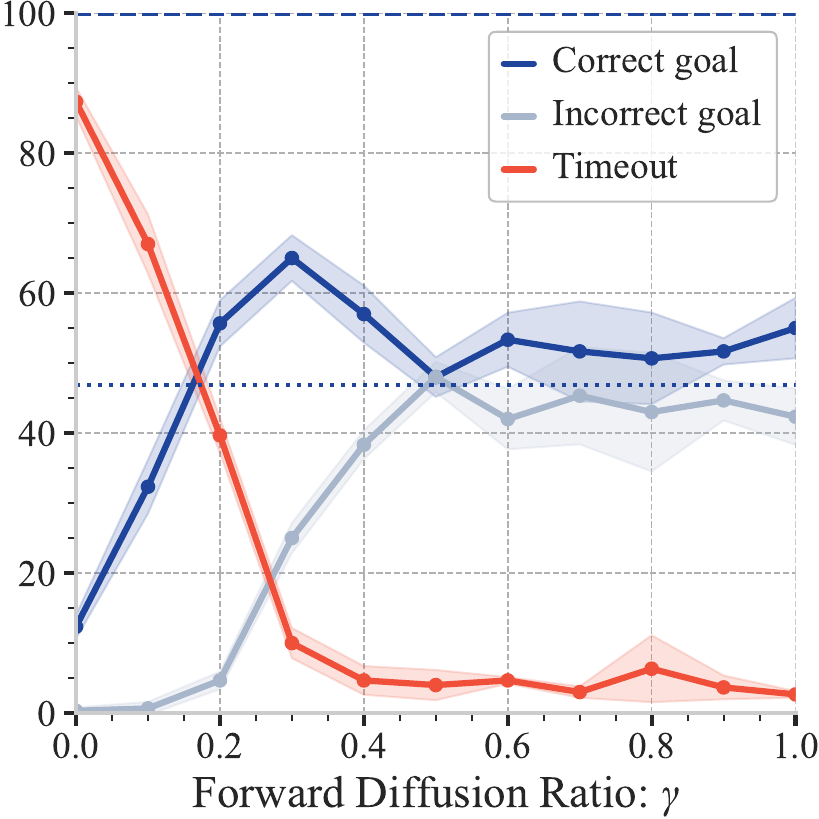}\label{fig:bp_laggy8}}
    \hfil
    \subfloat[$p_\text{laggy} = 0.9$]{\includegraphics[width=0.2\linewidth]{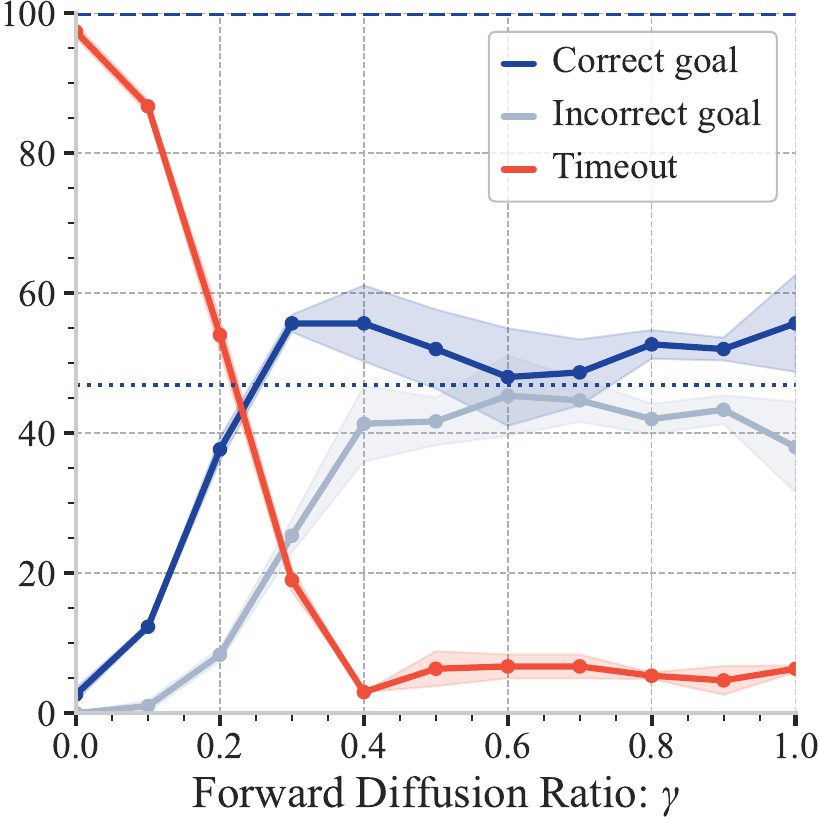}\label{fig:bp_laggy9}}
    
    \caption{Block Pushing performance in terms of the correct goal, wrong goal, and timeout rates as a function of the forward diffusion ratio $\fwr$ for a Laggy pilot with values ranging from \subref{fig:bp_laggy2} $p_\text{laggy} = 0.2$ to \subref{fig:bp_laggy9} $p_\text{laggy} = 0.9$. In all plots, the dashed blue line denotes the success rate of the expert pilot, while the dotted blue line is the success rate of our model when performing ``full'' diffusion (i.e., $\gamma = 1.0$) on an action sampled from a zero-mean isotropic Gaussian distribution, which we refer to as a ``Random'' pilot in the paper.}\label{fig:laggy_pilot_bp_cmp}
\end{figure*}
Figure~\ref{fig:laggy_pilot_bp_cmp} presents the performance of the Laggy pilot as a function of the forward diffusion ratio $\fwr$ for different settings of $p_\text{laggy}$. The behavior is similar to that of the Noisy pilot (Fig.~\ref{fig:noisy_pilot_bp_cmp}) except that increasing the level of copilot assitance to $\fwr = 0.2$ improves the correct goal rate for all parameterizations of the Laggy pilot. Again, as the forward diffusion ratio approaches $\fwr = 1.0$, the copilot effectively chooses randomly between the two goals, while maintaining a low timeout rate.

\begin{table*}[ht]
    \centering
    \caption{Correct goal and timeout rates on the Block Pushing task for different pilots with and without assistance. where we show the results for our chosen value for $\gamma = 0.2$  as well as the result of using full diffusion ($\gamma = 1.0$), which has the effect of removing information about the goal provided by the pilot. The plot corresponding to the data in the table is figure \ref{fig:bp-assisted}. Each entry corresponds to $10$ episodes across $30$ random seeds. Note that the Zero and Random pilots have no knowledge of the goal.}
    \label{tb:Block-Push-appendix}
    \begin{tabularx}{1.0\linewidth}{lYYYYYY}%
        \toprule
        & \multicolumn{3}{c}{Correct Goal Rate} & \multicolumn{3}{c}{Timeout Rate}\\
        Pilot & w/o Copilot ($\gamma=0.0$) & w/ Copilot (ours) ($\gamma=0.2$) & w/ Copilot ($\gamma=1.0$) & w/o Copilot ($\gamma=0.0$) & w/ Copilot (ours) ($\gamma=0.2$) & w/ Copilot ($\gamma=1.0$)\\
        \midrule
        Noisy & $ 62.33 \pm 2.49 $ &  $\bm{75.67 \pm 2.05} $ &  $ 52.67 \pm 7.41 $
        & $ \hphantom{0}37.33 \pm 2.05 $ &  $ \bm{20.00 \pm 3.27} $ &  $ 5.67 \pm 1.25 $\\
        Laggy & $ 42.00 \pm 4.90 $ &  $ \bm{74.33 \pm 2.87} $ &  $ 53.33 \pm 4.19 $ 
        & $ \hphantom{0}58.00 \pm 4.90 $ &  $ \bm{21.33 \pm 4.50} $ &  $ 5.67 \pm 1.70 $\\
        \midrule[0.1pt]
        Zero & $ \hphantom{0}0.00 \pm 0.00 $ &  $ 40.67 \pm 2.62 $ &  $ 52.33 \pm 0.47 $
        & $ 100.00 \pm 0.00 $ &  $ \hphantom{0}6.67 \pm 1.70 $ &  $ 3.67 \pm 2.36 $\\
        Random & $ \hphantom{0}0.00 \pm 0.00 $ &  $ 16.33 \pm 3.09 $ &  $ 52.00 \pm 3.74 $
        & $ 100.00 \pm 0.00 $ &  $ 68.00 \pm 1.63 $ &  $ 4.33 \pm 2.05 $\\
        Expert & $ 99.00 \pm 0.82 $ &  $ 94.67 \pm 2.05 $ &  $ 51.00 \pm 2.45 $
        & $ \hphantom{00}1.00 \pm 0.82 $ &  $ \hphantom{0}5.33 \pm 2.05 $ &  $ 5.67 \pm 2.49 $ \\                
        \bottomrule
    \end{tabularx}
\end{table*}

\section{Real Human User Experiments}\label{app:human-user-experiments}
\begin{figure*}[!h]
    \centering
    \subfloat[Lunar Lander (w/o Co-pilot)]{\includegraphics[width=0.245\linewidth, trim = 1.3cm 0.1cm 1.3cm 1.3cm, clip]{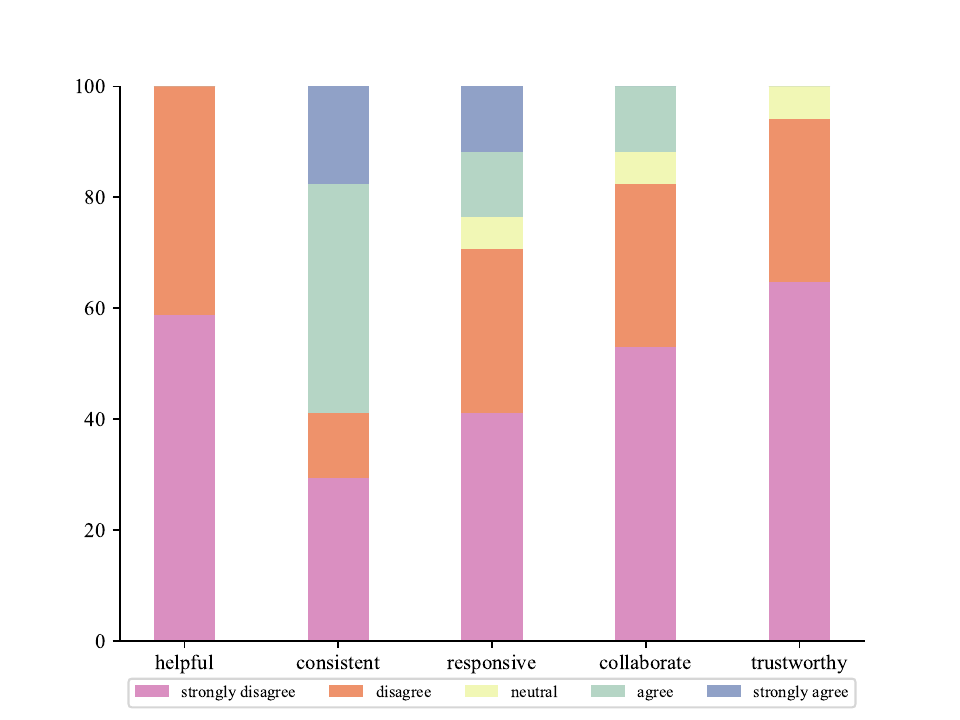}}\hfil
    \subfloat[Lunar Lander (w/ Co-pilot)]{\includegraphics[width=0.245\linewidth, trim = 1.3cm 0.1cm 1.3cm 1.3cm, clip]{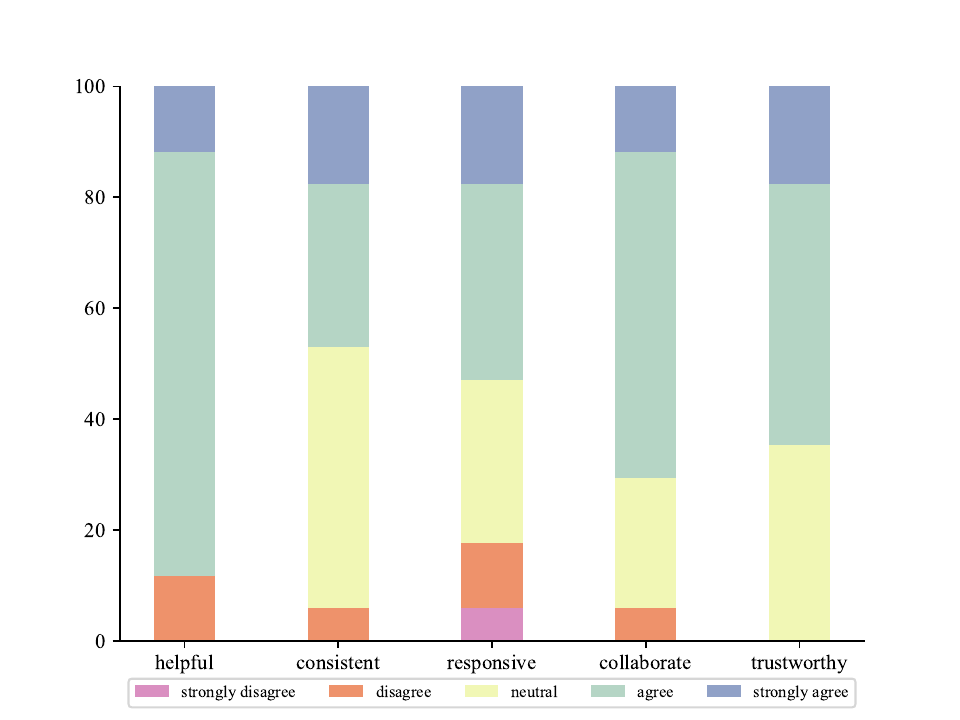}}
    \subfloat[Lunar Reacher (w/o Co-pilot)]{\includegraphics[width=0.245\linewidth, trim = 1.3cm 0.1cm 1.3cm 1.3cm, clip]{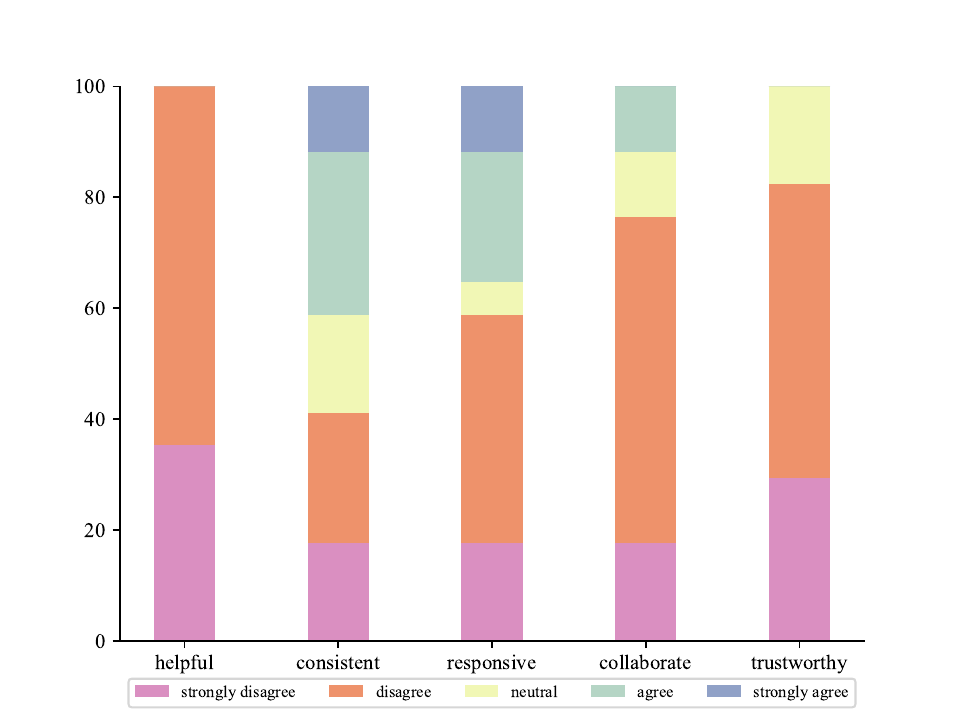}}\hfil
    \subfloat[Lunar Reacher (w/ Co-pilot)]{\includegraphics[width=0.245\linewidth, trim = 1.3cm 0.1cm 1.3cm 1.3cm, clip]{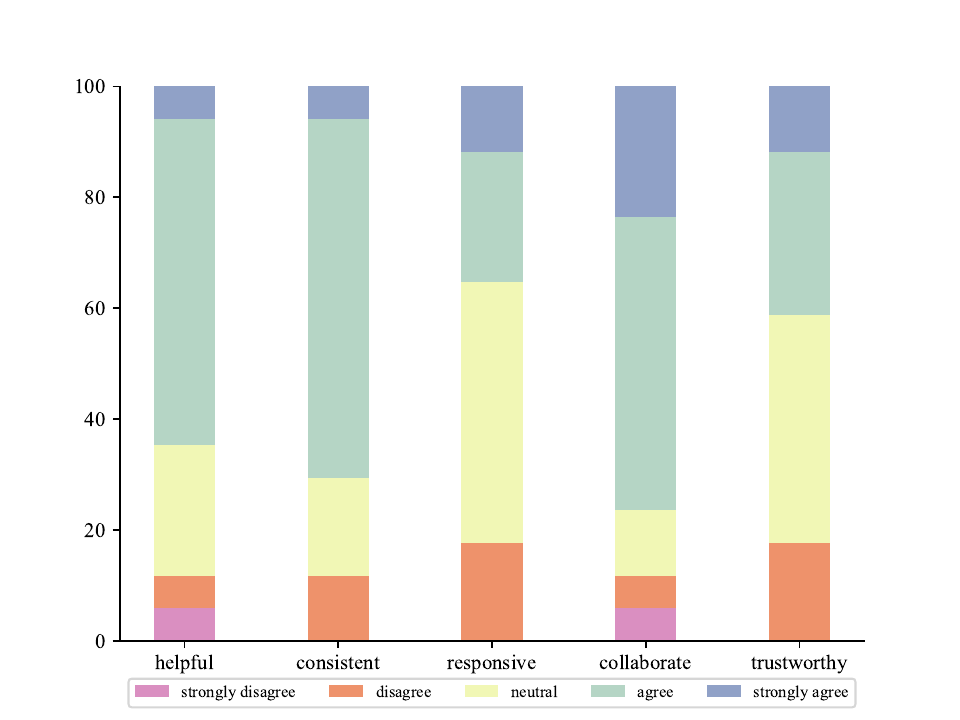}}
    \caption{Results of the human user qualitative surveys for Lunar Lander and Lunar Reacher without and with the assistance of our shared autonomy algorithm.}\label{fig:human-lunar-lander-reacher-survey}
\end{figure*}

In Lunar Lander, Lunar Reacher and real robot experiment, we asked the same set of quantitative questions using a five-point Likert scale.
Figure~\ref{fig:human-lunar-lander-reacher-survey} shows the results of Lunar Lander and Lunar Reacher. The users preferred our copilot over the placebo copilot in all of the metrics we asked in the question.

\begin{figure}[!t]
    \centering
    \subfloat[UR5 (w/o Co-pilot)]{\includegraphics[width=0.49\linewidth, trim = 1.3cm 0.1cm 1.3cm 1.3cm, clip]{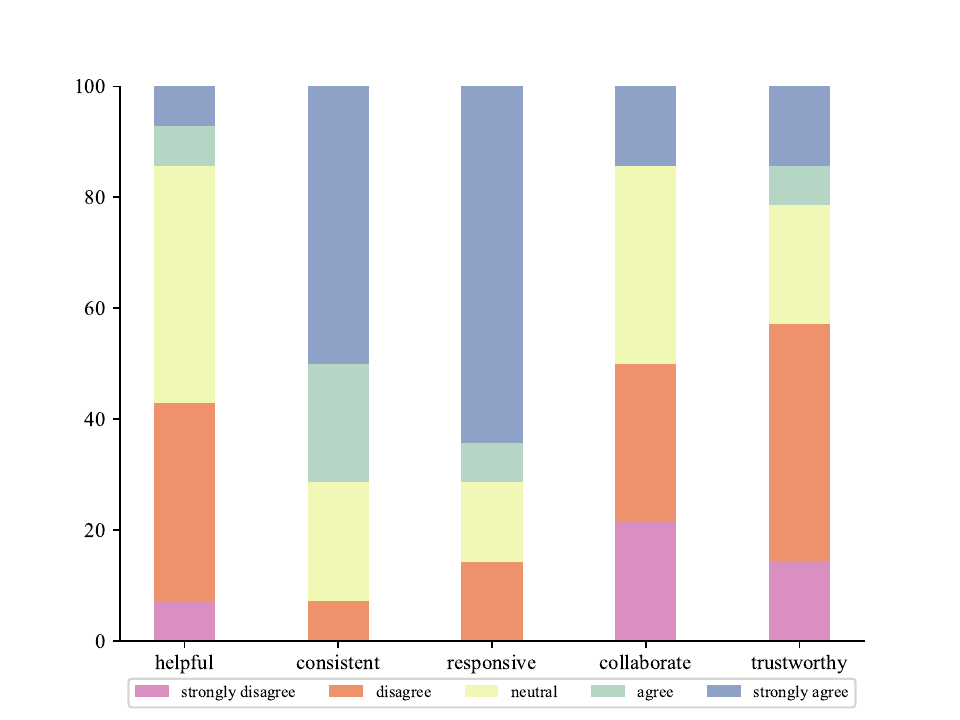}}\hfil
    \subfloat[UR5 (w/ Co-pilot)]{\includegraphics[width=0.49\linewidth, trim = 1.3cm 0.1cm 1.3cm 1.3cm, clip]{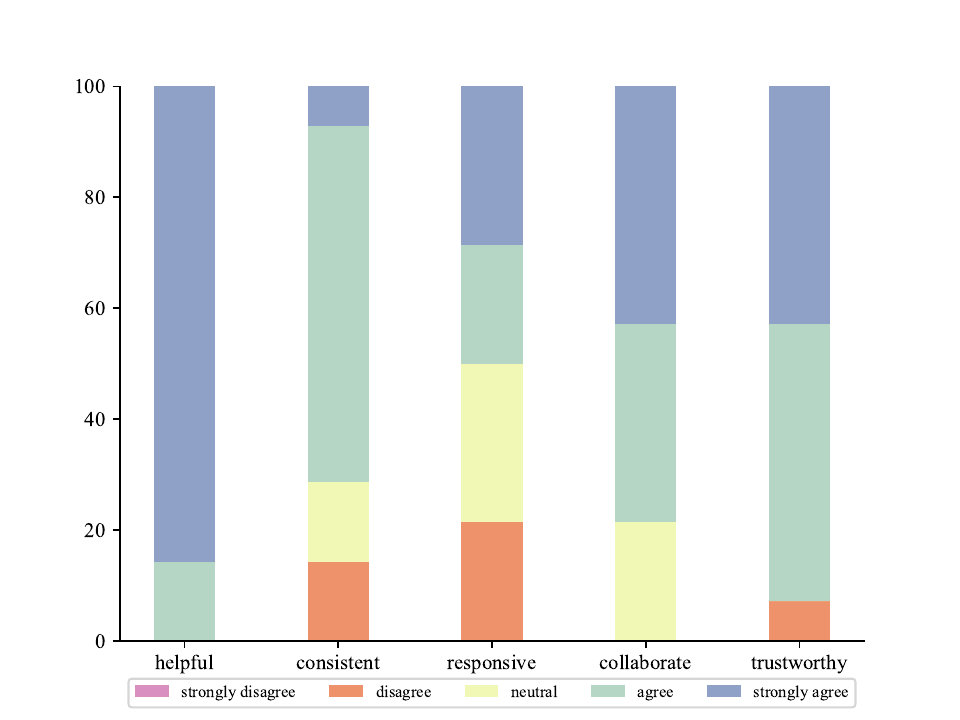}}
    \caption{Results of the human user qualitative surveys for the UR5 manipulation experiments (left) without and (right) with the assistance of our method.}\label{fig:human-UR5-survey}
    \vspace{7in} %
\end{figure}

The results of real robot experiment is summarized in Figure~\ref{fig:human-UR5-survey}. Our copilot got better scores in all metrics but for ``responsive''. We believe that this is because unreasonable user control inputs (e.g., going away from any of the goal locations) are often strongly corrected by our copilot, which in turn gives the user an impression that their action is disrespected. %

\end{document}